%% file: main_arxiv.tex
\documentclass{article}

\pdfoutput=1
\def\figpath{}
\input{\preamblepath packages}

\input{macros}

\usepackage[vmargin=1.00in,hmargin=1.00in,centering,letterpaper]{geometry}
  \usepackage{fancyhdr, lastpage}

\newlength\tindent
\newtoggle{arxiv}
\toggletrue{arxiv}
\newtoggle{icml}
\togglefalse{icml}

\title{Balancing Competing Objectives with Noisy Data \\ Score-Based Classifiers for Welfare-Aware Machine Learning}

\author{%
  Esther Rolf\thanks{esther\_rolf@berkeley.edu} \\
  UC Berkeley\\
  \and
  Max Simchowitz \\
  UC Berkeley \\
  \and
  Sarah Dean \\
  UC Berkeley \\
  \and
  Lydia T. Liu \\
  UC Berkeley \\
  \and
  Daniel Bj{\"o}rkegren \\
  Brown University \\
  \and
  Moritz Hardt \\
  UC Berkeley \\
  \and
  Joshua Blumenstock \\
  UC Berkeley \\
}
\date{}

\begin{document}

\maketitle

\setcitestyle{round}
\begin{abstract}
\input{\bodypath abstract}

\end{abstract}

\section{Introduction}
\label{sec:intro}
\input{\bodypath introduction}

\section{Related Work}
\label{sec:related_work}
\input{\bodypath related_work}

\section{Problem Setting: Pareto-optimal Policies}
\label{sec:setting}
\input{\bodypath problem_setting}

\section{Pareto Frontiers with Inexact Scores}
\label{sec:methods}
\input{\bodypath methods}

 \section{Experiments}
  \label{sec:experiments}
 \input{\bodypath experiments}

\section{Connections to Fairness Constraints}
  \label{sec:connections_to_fairness}
\input{\bodypath connections_to_fairness_shorter}

\section{Conclusions}
  \label{sec:conclusions}
\input{\bodypath conclusions}

\section*{Acknowledgements}
\input{\bodypath acks}

\bibliography{multi_obj.bib}
\bibliographystyle{plainnat}

\appendix

\clearpage

\onecolumn
\section{Proofs for General Characterization of Pareto Curves}
\input{\apppath technical_appendix}

\section{Experimental Details}
\label{sec:experiment_details}
\input{\apppath experiment_details}

\section{Appendix on Fairness}\label{app:fairness}
\input{\apppath fairness_appendix}
\end{document}

%% file: packages.tex

\usepackage{microtype}
\usepackage{graphicx}
\usepackage{subcaption}
\usepackage{booktabs} 
\usepackage{etoolbox}
\usepackage{amsmath,amssymb,enumitem}
\usepackage{amsthm}
\usepackage{hyperref}
\usepackage{cleveref}
\usepackage{comment}
\usepackage{xcolor}
\usepackage{natbib}
\usepackage{url}


%% file: macros.tex
\newcommand{\piopt}{\pi^{\mathrm{opt}}}

\newcommand{\Pareto}{\mathcal{P}}
\newcommand{\parfunc}{\mathbf{g}}

\newcommand{\emp}{\mathrm{emp}}
\newcommand{\exact}{\mathrm{exact}}

\newtheorem{thm}{Theorem}[section]

\newtheorem{lem}[thm]{Lemma}

\newtheorem{cor}[thm]{Corollary}
\newtheorem{prop}[thm]{Proposition}
\newtheorem{asm}{Assumption}

\newtheorem{example}{Example}

\newtheorem{defn}{Definition}[section]

\newcommand{\calD}{\mathcal{D}}

\newcommand{\calN}{\mathcal{N}}

\newcommand{\calC}{\mathcal{C}}

\newcommand{\R}{\mathbb{R}}
\newcommand{\I}{\mathbb{I}}
\newcommand{\Exp}{\mathbb{E}}

\renewcommand{\P}{\mathbf{P}}

\renewcommand{\Pr}{\mathbb{P}}

\renewcommand\part[1]{\vspace{.05in}\textbf{(#1)}}

\renewcommand{\Pr}{\mathbb{P}}
\renewcommand{\P}{\mathcal{P}}

\newcommand{\argmax}{\operatornamewithlimits{argmax}}

\newcommand{\maxprof}{\mathtt{MaxUtil}}

\newcommand{\supa}{_{\popa}}
\newcommand{\supb}{_{\popb}}
\newcommand{\supj}{_{\popj}}

\newcommand{\popa}{\mathsf{A}}
\newcommand{\popb}{\mathsf{B}}

\newcommand{\popj}{\mathsf{j}}
\newcommand{\popi}{\mathsf{i}}

\newcommand{\pol}{\boldsymbol{\tau}}
\newcommand{\polj}{\pol\supj}
\newcommand{\pola}{\pol\supa}
\newcommand{\polb}{\pol\supb}

\newcommand{\util}{\boldsymbol{u}}
\newcommand{\Util}{\mathcal{U}}

\newcommand{\prof}{\mathsf{P}}
\newcommand{\welf}{\mathsf{W}}

\newcommand{\mubar}{\overline{\mu}}
	\newcommand{\Piemp}{\Pi_{\mathrm{emp}}}

\newcommand{\fw}{f_{\welf}}

\newcommand{\fp}{f_{\prof}}
\newcommand{\fphat}{\widehat{f}_{\prof}}
\newcommand{\fwhat}{\widehat{f}_{\welf}}

\renewcommand{\util}{\mathcal{U}}
\newcommand{\utilw}{\util_{\welf}}
\newcommand{\utilp}{\util_{\prof}}
\renewcommand{\pol}{\pi}
\newcommand{\polhat}{\pi^{\mathrm{plug}}}
\newcommand{\polclass}{\Pi}
\newcommand{\polclasst}{\polclass_{\star}}

\newcommand{\welfscore}{w}
\newcommand{\profscore}{p}

\newcommand{\polst}{\pol^{\star}}

\newcommand{\indep}{\rotatebox[origin=c]{90}{$\models$}}

%% file: abstract.tex

While real-world decisions involve many competing objectives, algorithmic decisions are often evaluated with a single objective function. 
In this paper, we study algorithmic policies which explicitly trade off between a private objective (such as profit) and a public objective (such as social welfare).
We analyze a natural class of policies which trace an empirical Pareto frontier based on learned scores, and focus on how such decisions can be made in noisy or data-limited regimes.
Our theoretical results characterize the optimal strategies in this class, bound the Pareto errors due to inaccuracies in the scores, and show an equivalence between optimal strategies and a rich class of fairness-constrained profit-maximizing policies. 
We then present empirical results in two different contexts --- online content recommendation and sustainable abalone fisheries --- to underscore the applicability of our approach to a wide range of practical decisions. 
Taken together, these results shed light on inherent trade-offs in using machine learning for decisions that impact social welfare.

%% file: introduction.tex
From medical diagnosis 
and criminal justice 
to financial loans 
and humanitarian aid, 
consequential decisions increasingly rely on data-driven algorithms.
Machine learning algorithms used in these contexts are mostly trained to optimize a single metric of performance.
As a result, the decisions made by such algorithms can have unintended adverse side effects: profit-maximizing loans can have detrimental effects on borrowers \citep{skiba_payday_2009}
and fake news can undermine democratic institutions \citep{persily_2016_2017}. 

The field of fair machine learning proposes algorithmic approaches that mitigate the adverse effects of single objective maximization. Thus far it has predominantly done so by defining various fairness criteria that an algorithm ought to satisfy \citep[see e.g.,][and references therein]{barocas-hardt-narayanan}.
However, a growing literature highlights the inability of any one fairness definition to solve more general concerns of social equity~\citep{corbett2018measure}. The
 impossibility of satisfying all desirable criteria~\citep{kleinberg16inherent} and the unintended consequences of enforcing parity constraints based on sensitive attributes~\citep{kearns2017preventing} indicate that existing fairness solutions are not a panacea for these adverse effects. 
Recent work \citep{liu18delayed, hu19socialwelfare} contend that while social welfare is of primary concern in many applications, common fairness constraints may be at odds with the relevant notion of welfare.

In this paper, we consider \emph{welfare-aware machine learning} as an inherently multi-objective problem that requires explicitly balancing multiple objectives and outcomes. 
A central challenge is that certain objectives, like welfare, may be harder to measure than others. 
Building on the traditional notion of Pareto optimality, which provides a characterization of optimal policies under complete information, we develop methods to balance multiple objectives when those objectives are measured or predicted with error. 
%

%

We study a natural class of selection policies that balance multiple objectives (e.g., 
private profit and public welfare) when each individual has predicted \emph{scores} for each objective (e.g., their predicted contribution to total welfare and profit).
We show that this class of score-based policies has a natural connection to statistical parity constrained classifiers and their $\epsilon$-fair analogs. 
In the likely case where scores are imperfect predictors, we bound the sub-optimality of the multi-objective utility as a function of the estimator errors. 
Simulation experiments 
highlight characteristics of problem settings (e.g. correlation of the true scores) that affect the extent to which we can jointly maximize multiple objectives.  

We apply the  multi-objective framework to data from two diverse decision-making settings. 
We first consider an ecological setting of sustainable fishing, where we study score degradation to mimic certain dimensions being costly or impossible to measure. 
Our second empirical study uses existing data on the popularity and `social health' of roughly 40,000 videos promoted by YouTube's recommendation algorithm, 
and shows that multi-objective optimization could produce substantial increases in average video quality for almost negligible reductions in user engagement. 

In summary, we provide a characterization, theoretical analysis, and empirical study of a score-based multi-objective optimization framework for learning welfare-aware policies. 
{
We hope that our framework may help decouple the complex problem of defining and measuring welfare, which  has been studied at length in the  social sciences, \citep[e.g][]{deaton2016measuring}, from a machine toolkit geared towards optimizing it.
}

%% file: related_work.tex
\subsection{Fair and Welfare-Aware Machine Learning}
The growing subfield of \emph{fairness in machine learning} has investigated the implementation and implications of machine learning algorithms that satisfy definitions of fairness \citep{Dwork:2012,barocas2016big,barocas-hardt-narayanan}. Machine learning systems in general cannot satisfy multiple definitions of group fairness \citep{chould16fair,kleinberg16inherent}, and there are inherent limitations to using observational criteria~\citep{Kilbertus17}. Alternative notions of fairness more directly encode specific trade-offs between separate objectives, such as per-group accuracies~\citep{kim2019multiaccuracy} and overall accuracy versus a continuous fairness score \citep{zliobaite2015relation}. 
These fairness strategies represent trade-offs with domain specific implications, for example in tax policy~\citep{fleurbaey2018optimal} or targeted 
poverty prediction~\citep{noriegaalgorithmic}.

An emerging line of work is concerned with the long-term impact of algorithmic decisions on societal welfare and fairness \citep{ensign2018runaway,hu_welfare_2018,mouzannar_fair_2018,liu2019disparate}.  \citet{liu18delayed} investigated the potentially harmful delayed impact that a fairness-satisfying decision policy has on the well-being of different subpopulations.  In a similar spirit, \citet{hu19socialwelfare} showed that always preferring ``more fair'' classifiers does not abide by the Pareto Principle (the principle that a policy must be preferable for at least one of multiple groups) in terms of welfare. Motivated by these findings, our work acknowledges that algorithmic policies affect individuals and institutions in many dimensions, and explicitly encodes these dimensions in policy optimization.

We will show that fairness constrained policies that result in per-group score thresholds and their $\epsilon$-fair equivalent soft-constrained analogs~\citep{elzayn2019fair} can be cast as specific instances of the Pareto framework that we study. 
Analyzing the limitations of this optimization regime with imperfect scores therefore connects to a recent literature on achieving group fairness with noisy or missing group class labels~\citep{lamy19noisetolerant,awasthi2019equalized}, including using proxies of group status~\citep{gupta18proxy,chen2019fairness}.
The explicit welfare effects of selection in our model also complement the notion of utilization in fair allocation problems~\citep{elzayn2019fair,donahue2020fairness}.

\subsection{Multi-objective Machine Learning}
We consider two simultaneous goals of a learned classifier: achieving high profit value of the classification policy, while improving a measure of social welfare.  This relates to an existing literature on multi-objective optimization in machine learning~\citep{jin2008pareto,jin2006multi}, where many algorithms exist for finding or approximating global optima ~\citep{Deb:2001:MOU,knowles2006parego, desideri2012multiple} under different problem formulations.  

Our work studies the Pareto solutions that arise from learned score functions, and is therefore related to, but distinct from a large literature on learning Pareto frontiers directly. Evolutionary strategies are a popular class of approaches to estimating a Pareto frontier from empirical data, as they refine a class of several policies at once~\citep{Deb:2001:MOU,kim2005adaptive}. 
Many of these strategies use surrogate convex loss functions to afford better convergence to solutions. Surrogate functions can be defined over each dimension independently~\citep{knowles2006parego}, or as a single function over both objective dimensions~\citep{loshchilov2010mono}. While surrogate loss functions play an important role in a direct optimization of non-convex utility functions, our framework provides an alternative approach, so long as scores functions can be reliably estimated. 

Another class of methods explicitly incorporates models of uncertainty in dual-objective optimization~\citep{peitz2018gradient,paria2018flexible}.
For sequential decision-making, there has been recent work on finding Pareto-optimal policies for reinforcement learning settings~\citep{van2014multi, liu2014multiobjective,roijers2017multi}. 
To promote applicability of our work to a variety of real-world domains where noise sources are diverse, and the effects of single policy enactments complex, we first develop a methodology under a noise-free setting, then extend to reasonable forms of error in provided estimates.

\subsection{Measures of Social Welfare}
\label{sec:welfare_literature}
The definition and measurement of welfare is an important and complex problem that has received considerable attention in the social science literature \citep[cf.][]{deaton_measurement_1980,deaton2016measuring,stiglitz_measurement_2009}.
There, a standard approach is to sum up individual measures of welfare, to obtain an aggregate measure of societal welfare. 
The separability assumption (independent individual scores) is a standard simplifying assumption~\citep[e.g.][]{florio2014applied} that appears in the foundational work of~\cite{pigou1920}, as well as 
\cite{burk_reformulation_1938}, \cite{samuelson_foundations_1947}, ~\cite{arrow1963social} and ~\cite{sen1973behaviour}. 
Future work may explore alternative social welfare function \citep[e.g.][]{clark_satisfaction_1996}. Our focus is on bringing machine learning to the most common notion of welfare.


%% file: problem_setting.tex



We consider a setting in which a centralized policymaker has two simultaneous objectives: to maximize some private return (such as revenue or user engagement), which we generically refer to as \emph{profit}; and to improve a public objective (such as social welfare or user health), which we refer to as \emph{welfare}. 
The policymaker makes decisions about \emph{individuals}, who are specified by feature vectors $x \in \R^d$. 
Decision policies are functions that output a randomized decision $\pol(x) \in [0,1]$  corresponding to the \emph{probability} that an individual with features $x$ is selected. 
To each individual we associate a value $\profscore$ representing the expected profit to be garnered from approving this individual and $\welfscore$ encoding the change in welfare.
The profit and welfare objectives are thus expectations over the joint distribution of $(w,p,x)$:
\begin{align}\label{eq:utility}
\utilw(\pol)=\Exp[\welfscore \cdot \pol(x) ] \quad \text{and} \quad \utilp(\pol)=\Exp[\profscore \cdot \pol(x) ]\:.
\end{align}
{
Notice that this aggregate measure of societal welfare is defined as a sum of individual measures of welfare; this is a standard approach in the social science literature (see Section~\ref{sec:welfare_literature}).
While this induces limitations on the form of the welfare function, it affords flexibility when focusing instead on the resulting binary decision, a point we expand on in Section~\ref{sec:connections_to_fairness}. 
}

\iftoggle{arxiv}{
\begin{figure}[ht!]
  \centering
    \includegraphics[width=0.96\textwidth]{\figpath 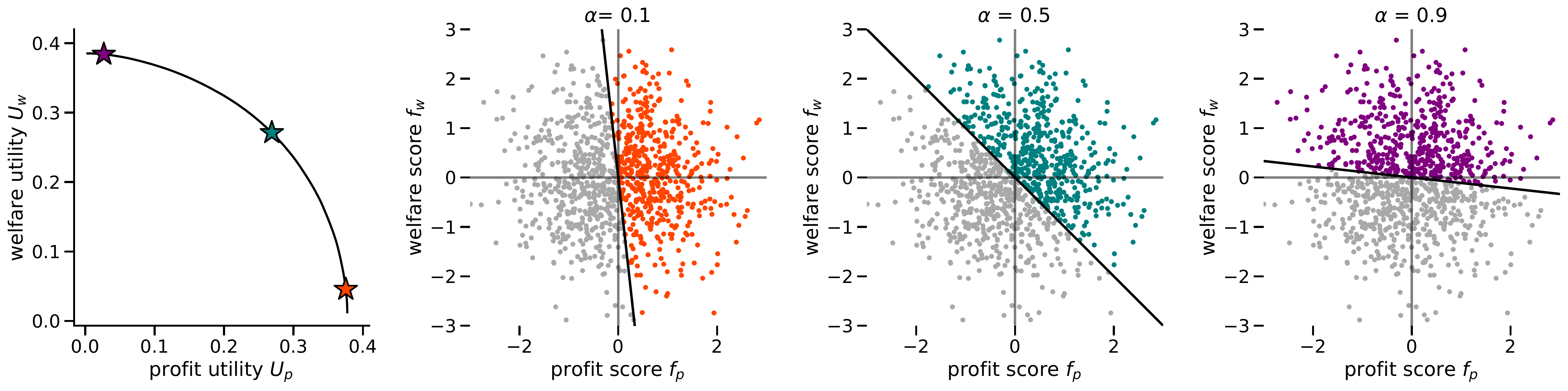}
    \caption{Illustration of a Pareto curve (bottom left) and the decision boundaries induced by three different trade-off parameters $\alpha$. Colored (darker in gray scale) points indicate selected individuals. 
    \label{fig:pareto_sim_first}}
\end{figure}
}
{ 
\begin{figure}[t]
  \centering
    \includegraphics[width=0.48\textwidth]{\figpath vary_alpha_exact_scores_square.pdf}
    \caption{Illustration of a Pareto curve (bottom left) and the decision boundaries induced by three different trade-off parameters $\alpha$. Colored (darker in gray scale) points indicate selected individuals. 
    \label{fig:pareto_sim_first}}
\end{figure}
}

Given two objectives, one can no longer define a unique optimal policy $\pol$. Instead, we focus on policies $\pol$ which are \emph{Pareto-optimal} \citep{pareto1906manuale}, in the sense that they are not strictly dominated by any alternative policy, i.e. there is no $\pol'$ such that both $\utilp$ and $\utilw$ are strictly larger under $\pol'$.

For a general set of policy classes (defined in \Cref{prop:composite}\footnote{All references starting with letters appear in the appendices.}), it is equivalent to consider policies that maximize a weighted combination of both objectives. 
We can thus parametrize the Pareto-optimal policies by $\alpha\in[0,1]$:
\begin{defn}[Pareto-optimal policies]
An $\alpha$-Pareto-optimal policy (for $\alpha \in [0,1]$) satisfies: 
\iftoggle{arxiv}{
	\begin{align*}
\polst_\alpha &\in \argmax \ \util_{\alpha}(\pol),
\\
 \util_{\alpha}(\pol) &:= (1-\alpha)\utilp(\pol) + \alpha \utilw(\pol).
\end{align*}}{
    \begin{align*}
	\polst_\alpha \in \argmax \ \util_{\alpha}(\pol),
	\\
	 \util_{\alpha}(\pol) := (1-\alpha)\utilp(\pol) + \alpha \utilw(\pol).
	\end{align*}
}

\end{defn}

In the definition above, the maximization of $\pi$ is taken over the class of randomized policies $\pi(x) \rightarrow [0,1]$. In Section~\ref{sec:exact_scores} we show that when features $x$ can exactly encode scores the optimal policy is a threshold of the scores. 

\subsection{Optimal Policies with Exact Scores}
\label{sec:exact_scores}
We briefly consider an idealized setting, where the welfare and profit contributions $\welfscore$ and $p$ can be directly determined from the features $x$ via exact \emph{score functions},
$\fw(x) = \welfscore$, $\fp(x) = \profscore$.
These exact score functions can be thought of as sufficient statistics for the decision: 
the expected weighted contribution from accepted individuals is described by $((1-\alpha)\profscore + \alpha \welfscore)$.
Therefore, one can show (\Cref{prop:pareto-thresholds}) that the optimal policy is given by thresholding this composite:
\begin{align} \label{eq:alpha_pareto_score_pol}
	\polst_{\alpha}(\profscore,\welfscore) = \I((1-\alpha) \profscore + \alpha \welfscore  \ge 0).
	\end{align}
Though they are all Pareto-optimal, the policies $\polst_{\alpha}$ induce different trade-offs between the two objectives.
The parameter $\alpha$ determines this trade-off, tracing the \emph{Pareto frontier}:
\begin{align*}
	 \Pareto_{\exact} := \{(\util_{\prof}(\polst_{\alpha}), \util_\welf(\polst_\alpha)): \alpha \in [0,1]\}
	 \end{align*}
Figure~\ref{fig:pareto_sim_first} plots an example of this curve (bottom-left panel) and the corresponding decision rules for three points along it.
We note the concave shape of this curve, a manifestation of \emph{diminishing marginal returns}: as a decision policy forgoes profit to increase total welfare, less welfare is gained for the same amount of profit forgone. The notion of diminishing return is formalized in \Cref{thm:exact_pareto}.

%% file: methods.tex
In many settings, we typically do not know the profit score $\profscore$ or welfare score $\welfscore$ --- or the score functions $\fp$ and $\fw$ --- for all individuals a priori. 
Instead, we might estimate score functions $\fphat(x)$ and $\fwhat(x)$ from data in the hope that these models can provide good predictions on future examples.

We study the class of \emph{score-based policies} that act on the predicted scores:
\begin{defn}[Score-based policy class]
\begin{align*}
 \Piemp := \{\,\pi : \,(\fphat(X),\fwhat(X))\mapsto [0,1]\,\}
\end{align*}
\end{defn}
Focusing on this class of policies allows us to characterize optimal policies within this class (Section~\ref{sec:opt_scores}), derive diagnosable bounds the utility of suboptimal policies (Section~\ref{sec:bounds}), and relate our results to common fairness criteria (Section.~\ref{sec:connections_to_fairness}).
We summarize additional benefits as well as potential limitations of restricting our study to this policy class in Section~\ref{sec:conclusions}.

\subsection{Pareto-optimality for Learned Scores}
\label{sec:opt_scores}

To characterize Pareto-optimal policies over $\Piemp$, we define the following conditional expectations over the  distribution $\calD$ of $(x,\profscore, \welfscore)$:
\begin{align*}
\mubar_{\prof}(  \fphat(x),\fwhat(x))  &:= \Exp_{\calD}[\profscore \mid \fphat(x),\fwhat(x)], \\
\mubar_{\welf}( \fphat(x),\fwhat(x))  &:= 
\Exp_{\calD}[\welfscore \mid \fphat(x),\fwhat(x)]\:.
\end{align*}
Intuitively, these values represent our best guesses of $p$ and $w$, given the predicted scores.
We define $\piopt_{\alpha}$ as the threshold policy on the composite of these predictions:
\begin{align*}
\piopt_{\alpha} :=
\I( (1-\alpha) \cdot \mubar_{\prof}+\alpha \cdot \mubar_{\welf} \ge 0).
\end{align*}

\begin{thm}[Pareto frontier in inexact knowledge case]\label{thm:pareto_inexact}
Given any population distribution $\calD$ over $(x,\profscore, \welfscore)$ and empirical score functions $\fwhat$ and $\fphat$,
\begin{enumerate}[label=(\roman*),leftmargin=*]
 	\item {The policies $\piopt_{\alpha}$ are Pareto optimal over the class $\Piemp$, with $\piopt_{\alpha} \in \argmax_{\pi \in \Piemp}\Util_{\alpha}(\pi)$.
	}
	\item {The Pareto frontier $\Pareto(\Pi_{\emp})$ is given by $\{(\util_{\prof}(\piopt_{\alpha}), \util_\welf(\piopt_\alpha): \alpha \in [0,1]\}$. The associated function mapping $\sup_{\pol \in \Piemp}\{\utilw(\pol): \utilp(\pol) = p\}$ is concave and non-increasing in $p$. }
	\item {The empirical frontier $\Pareto_{\emp}$ is dominated by the exact frontier $\Pareto_{\exact}$. That is, if 
		$(p,\welfscore_{\exact}) \in \Pareto_{\exact}$ and $(p,\welfscore_{\emp}) \in \Pareto_{\emp}$, then  $\welfscore_{\emp} \le \welfscore_{\exact}$.
	\label{item:estimated_curve_below_true_curve}}
 \end{enumerate}
\end{thm}
Thus an optimal empirical-score based policy can also be realized as a threshold policy (this time of the conditional expectations), and it obeys the same diminishing-returns phenomenon as in the exact score case. 
One example of score predictors that achieves this optimality is the 
\emph{Bayes optimal estimators} i.e., $\fphat(x) = \Exp[\profscore\mid x]$ and $\fwhat(x) = \Exp[\welfscore\mid x]$. 
We present a proof of~\Cref{thm:pareto_inexact} in Appendix~\ref{sec:prop_suff_stat}.
\subsection{Plug-in Policies}
In general, we may have access to score predictions or the ability to learn them from data, but not a guarantee that the predictions are Bayes' optimal. In the hopes that the predicted scores will suffice, we can define a natural selection rule based on $\alpha$-defined \emph{plug-in threshold policies}.
\begin{defn}[Plug-in policy]
For $\alpha \in [0,1]$ and score predictions $\fphat(x), \fwhat(x)$, the $\alpha$-plug-in policy is:
\begin{align}\label{eq:empirical_score_threshold}
\polhat_{\alpha}(x) = \I((1-\alpha)\fphat(x) + \alpha \fwhat(x)\ge 0)\:.
\end{align}
\end{defn}

Since $\piopt_{\alpha}$ requires computing conditional expectations over the distribution $\calD$, it will in general will differ from the plug-in policy~\eqref{eq:empirical_score_threshold}.  
The following corollary of~\Cref{thm:pareto_inexact} gives a condition in which $\piopt_{\alpha}$ and $\polhat_{\alpha}$ coincide. 
\begin{cor}
\label{cor:well_calibrated}
The plug-in policies $\polhat_{\alpha}$ are optimal in the class $\Piemp$ as long as the predicted score functions are \emph{well-calibrated}, in the sense that $\Exp[\profscore \mid \fphat(x),\fwhat(x)] = \fphat(x)$ and $\Exp[\welfscore \mid \fphat(x),\fwhat(x)] = \fwhat(x)$. 
\end{cor}
\begin{proof}
In this case, $\bar{\mu}_\profscore = \fphat(x)$ and $\bar{\mu}_\welfscore = \fwhat(x)$, so we may invoke~\Cref{thm:pareto_inexact}. 
\end{proof}
Under typical conditions~\citep{liu18implicit}, this form of calibration can be achieved by empirical risk minimization.

 
%
In Section~\ref{sec:bounds}, we bound the error in the plug-in policies by the error by the individual errors in each score.
Simulation experiments in Section~\ref{sec:experiments} detail the use of the plug in policy under controlled degradations of learned score accuracy. Real-data experiments provide further insight into using the plug-in policy for welfare-aware optimization in practice.

\iftoggle{arxiv}{
\begin{figure*}[ht!]
    \centering
    \begin{subfigure}[h]{.36\linewidth}
        \centering
        \includegraphics[height=1.6in]{\figpath 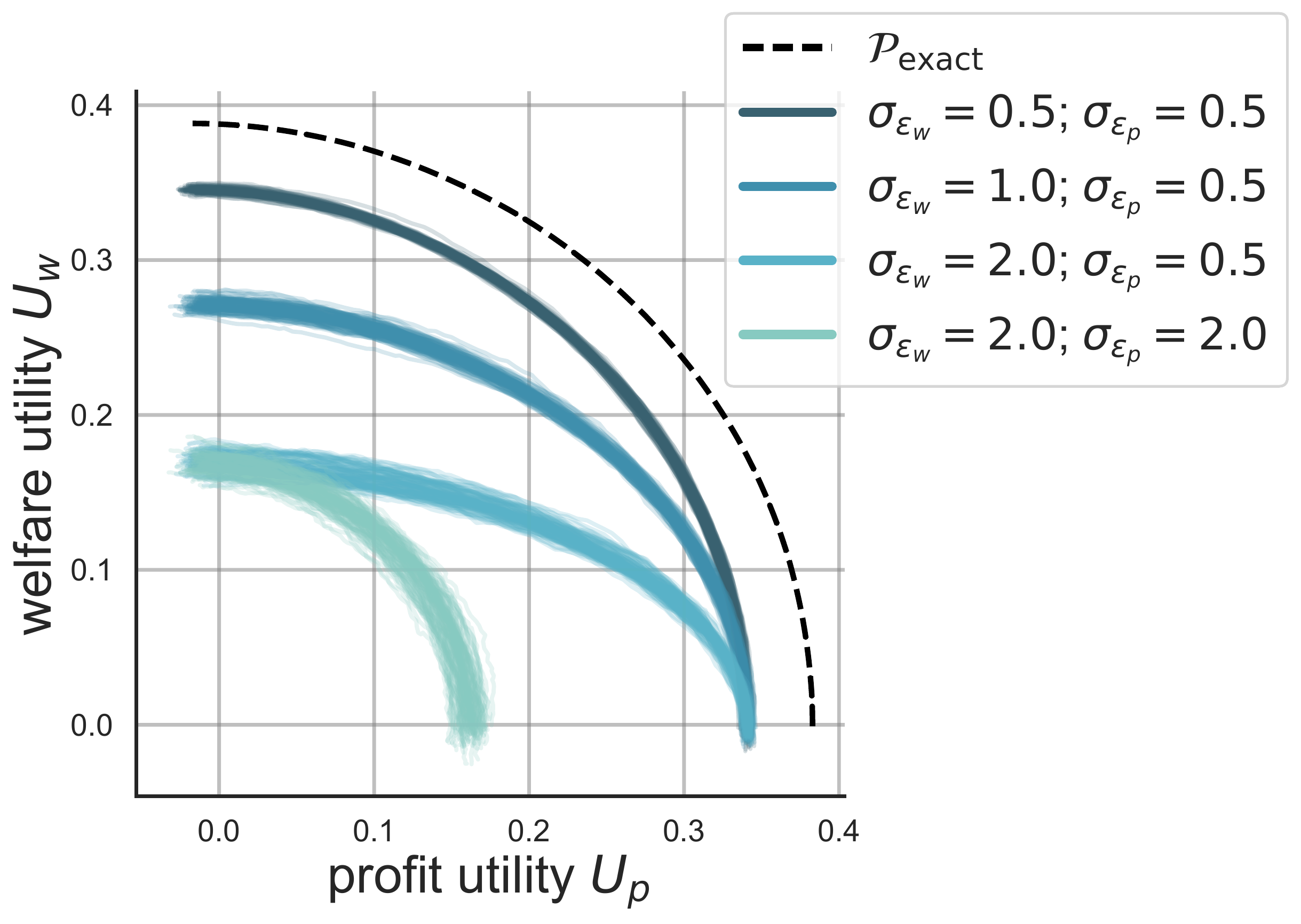}
           \caption{ No correlation, varying noise. 
            \label{fig:uncorrelated_plus_noise}}
    \end{subfigure}%
    ~ 
    \begin{subfigure}[h]{.3\linewidth}
        \includegraphics[height=1.6in]{\figpath 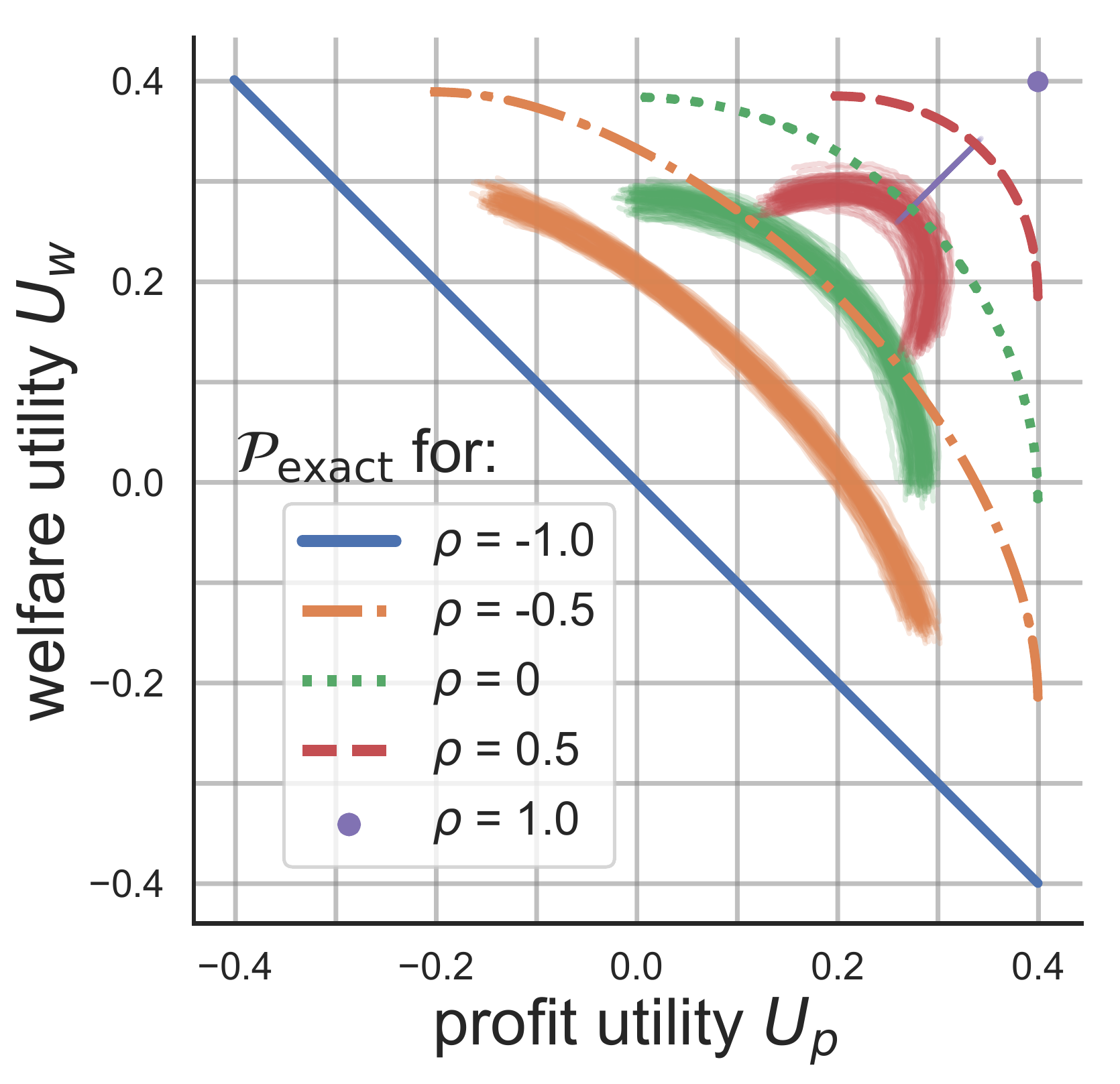} 
          \caption{ Varying correlation, fixed and equal noise $\sigma_{\varepsilon_p} = \sigma_{\varepsilon_w} = 1$.
            \label{fig:corr_scores_fixed_noise}}
    \end{subfigure}
    ~
    \begin{subfigure}[h]{.3\linewidth}
        \centering
       \includegraphics[height=1.6in]{\figpath 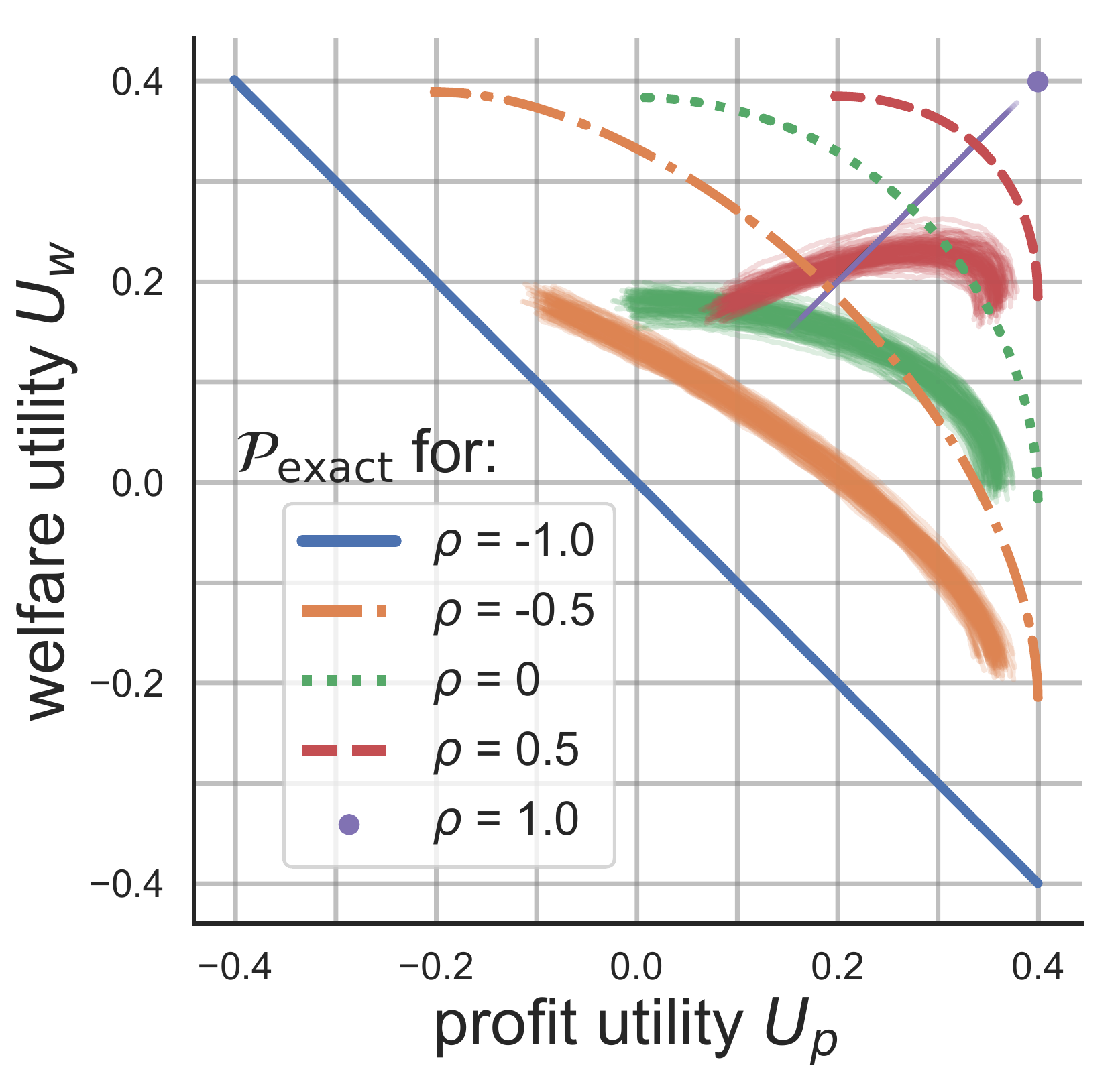}
        \caption{ Varying correlation, fixed and unequal noise $\sigma_{\varepsilon_p} = 0.5$,  $\sigma_{\varepsilon_w} = 2.0$.\label{fig:corr_scores_differing_noise}}
    \end{subfigure}
        \caption{Simulated experiments corresponding to the setting in \Cref{ex:gaussian_scores_subgaussian_error} (fixing $\sigma_w = \sigma_p = 1$).
        Empirical frontiers $\Pareto(\Pi_{\emp})$ 
        for 100 random trials with $n=5,000$ each are shown  as overlaid translucent  curves.
        Exact frontiers $\Pareto_\exact$ are shown as dashed curves.
        }
\end{figure*}
}
{   
\begin{figure*}[h!]
    \centering
    \begin{subfigure}[h]{.36\linewidth}
        \centering
        \iftoggle{arxiv}{
        \includegraphics[height=1.8in]{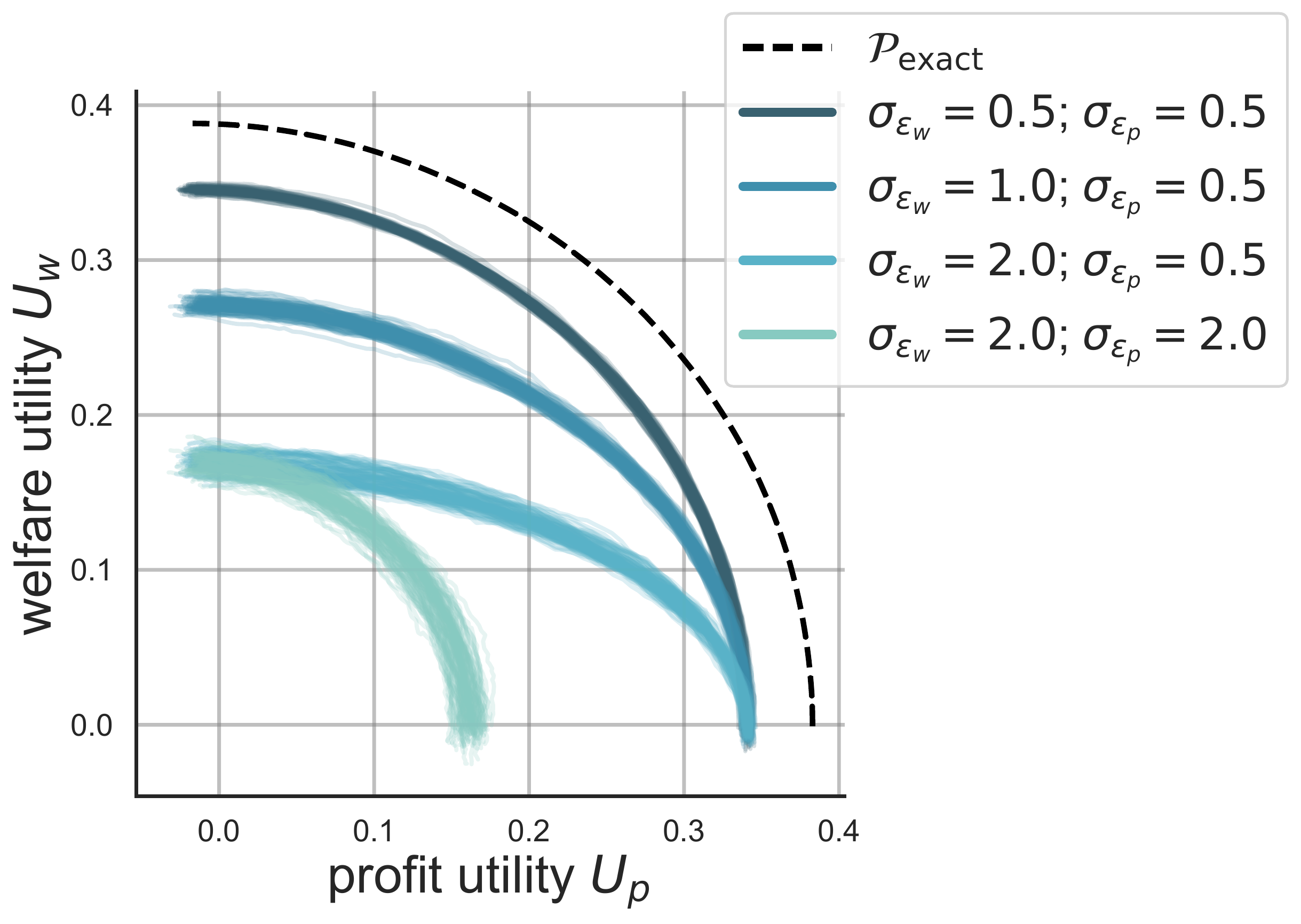}
        }
        {\includegraphics[height=1.8in]{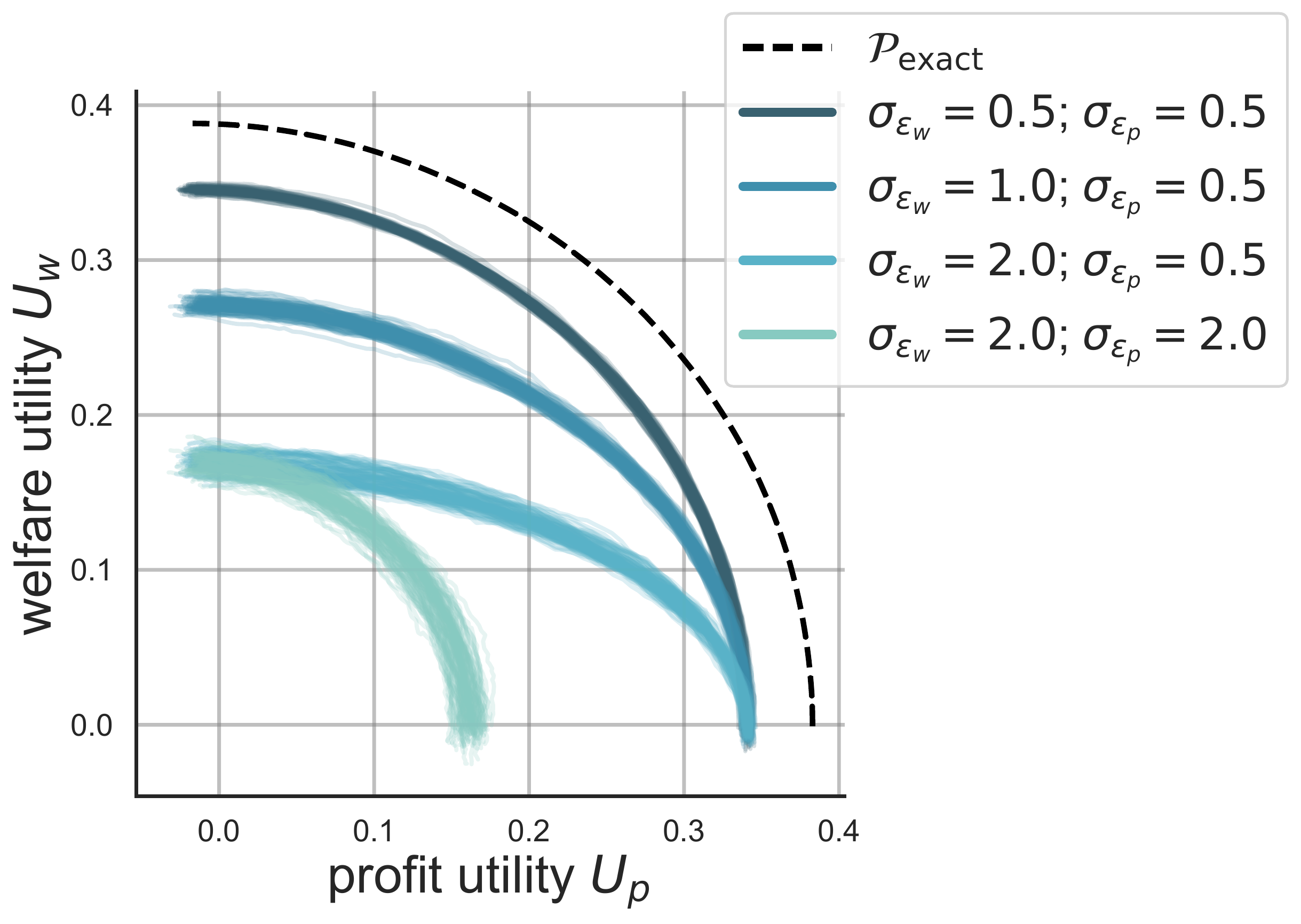}}
           \caption{ No correlation, varying noise. \\ \
            \label{fig:uncorrelated_plus_noise}}
    \end{subfigure}%
    ~ 
    \begin{subfigure}[h]{.3\linewidth}
        \centering
        \iftoggle{arxiv}{
        \includegraphics[height=1.8in]{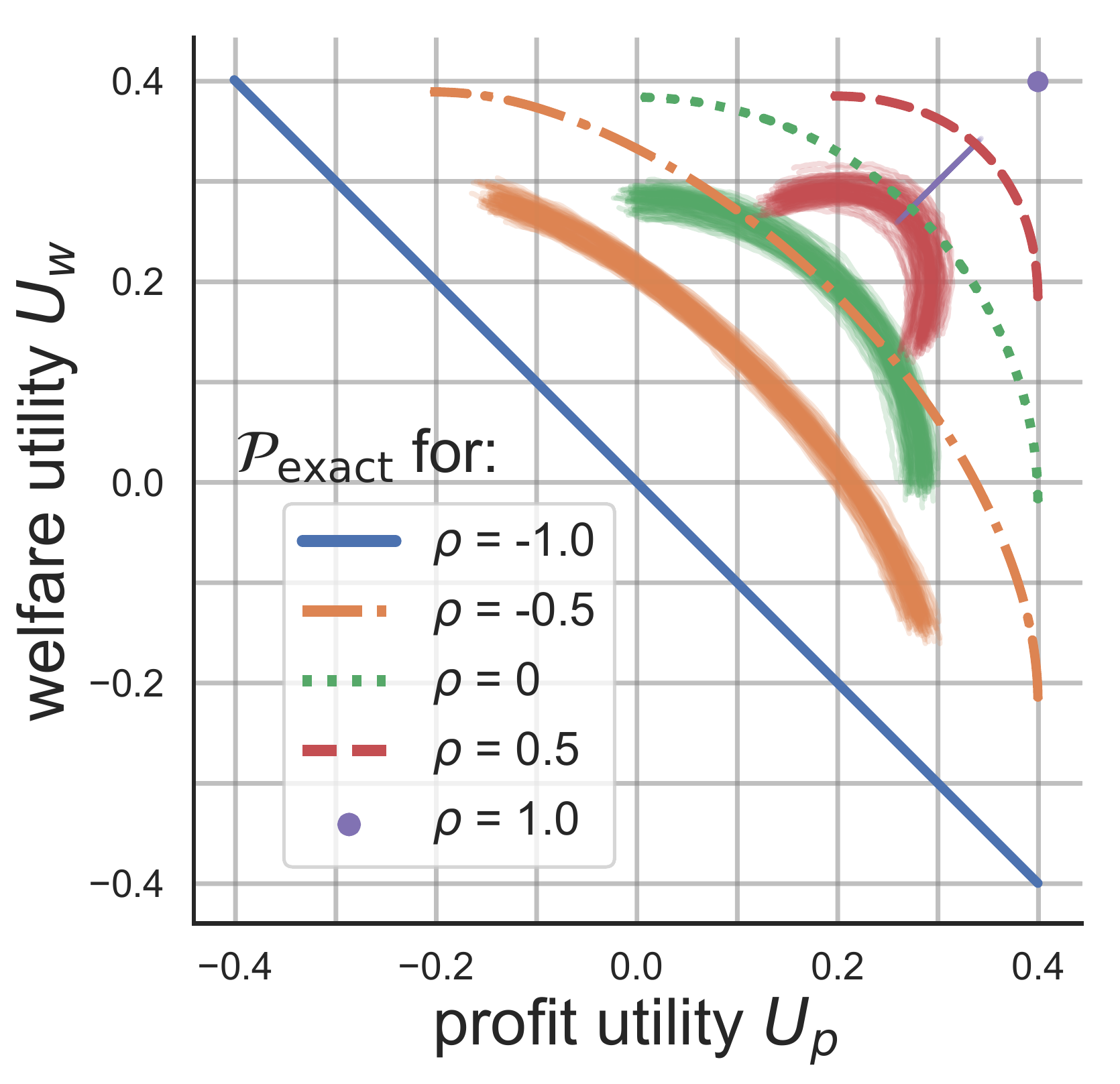}
        }
        {\includegraphics[height=1.8in]{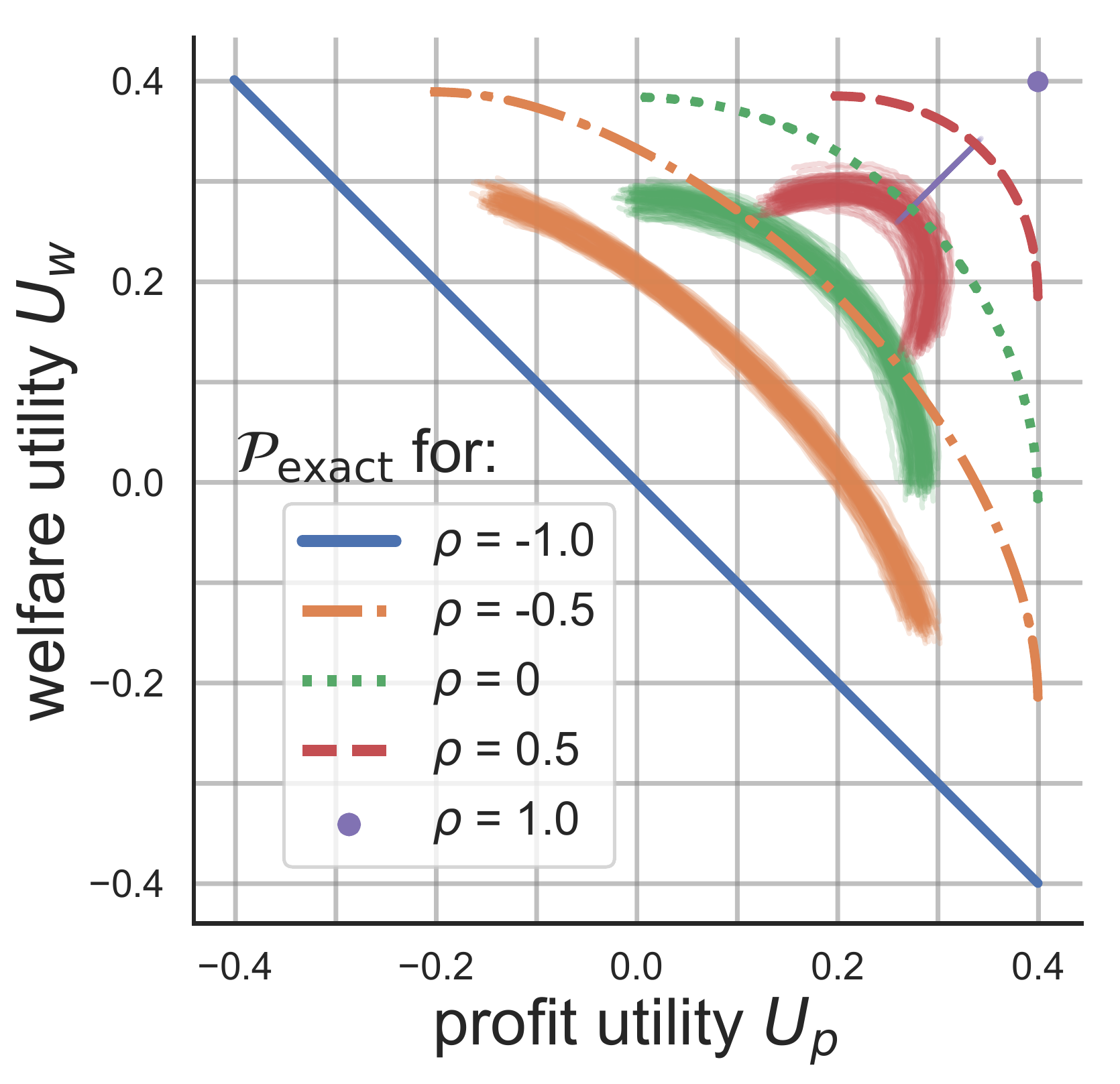}}
          \caption{ Varying correlation, fixed and equal noise $\sigma_{\varepsilon_p} = \sigma_{\varepsilon_w} = 1$.
            \label{fig:corr_scores_fixed_noise}}
    \end{subfigure}
    ~~
    \begin{subfigure}[h]{.3\linewidth}
        \centering
        \iftoggle{arxiv}{
        \includegraphics[height=1.8in]{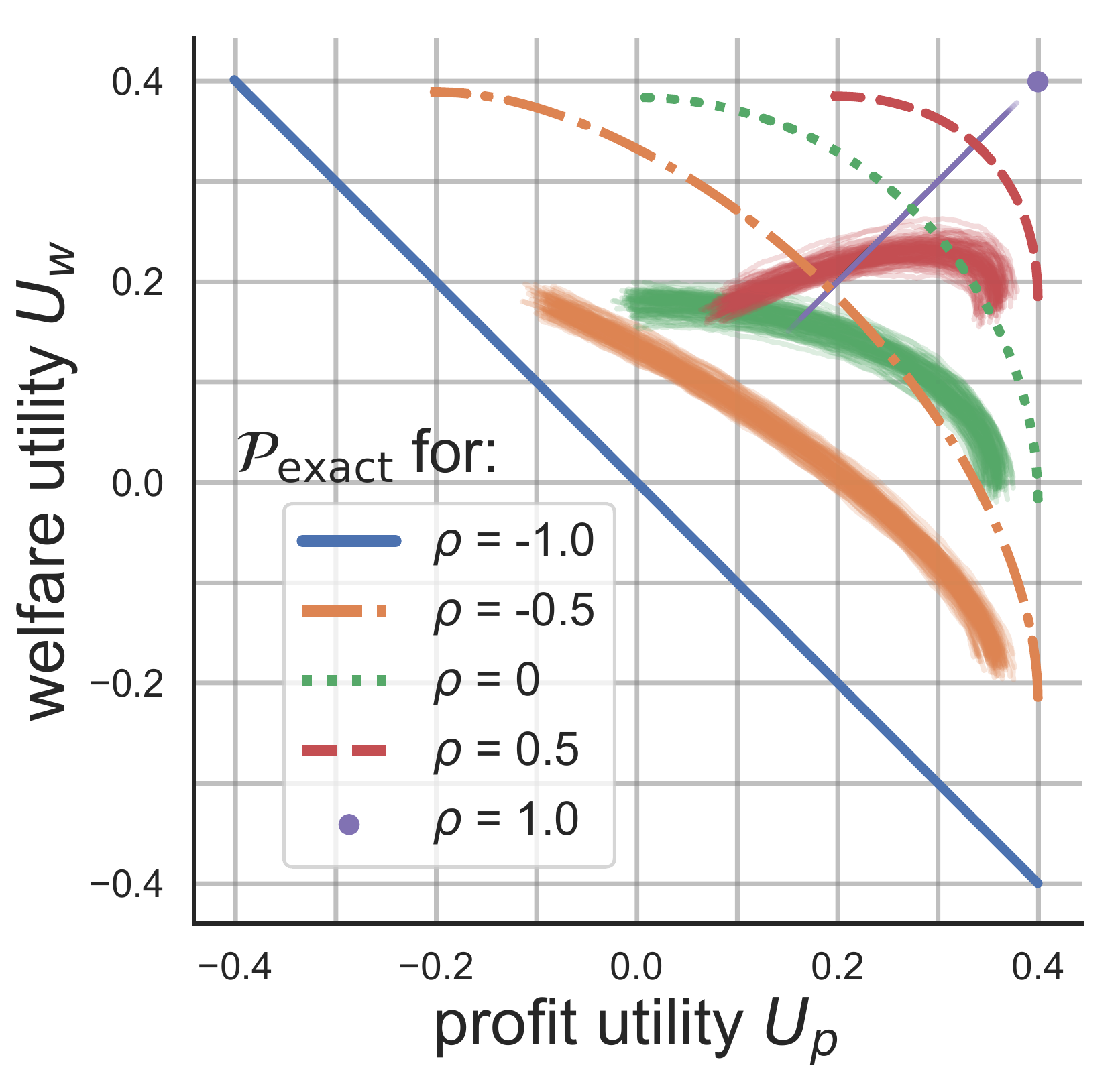}
        }
        {\includegraphics[height=1.8in]{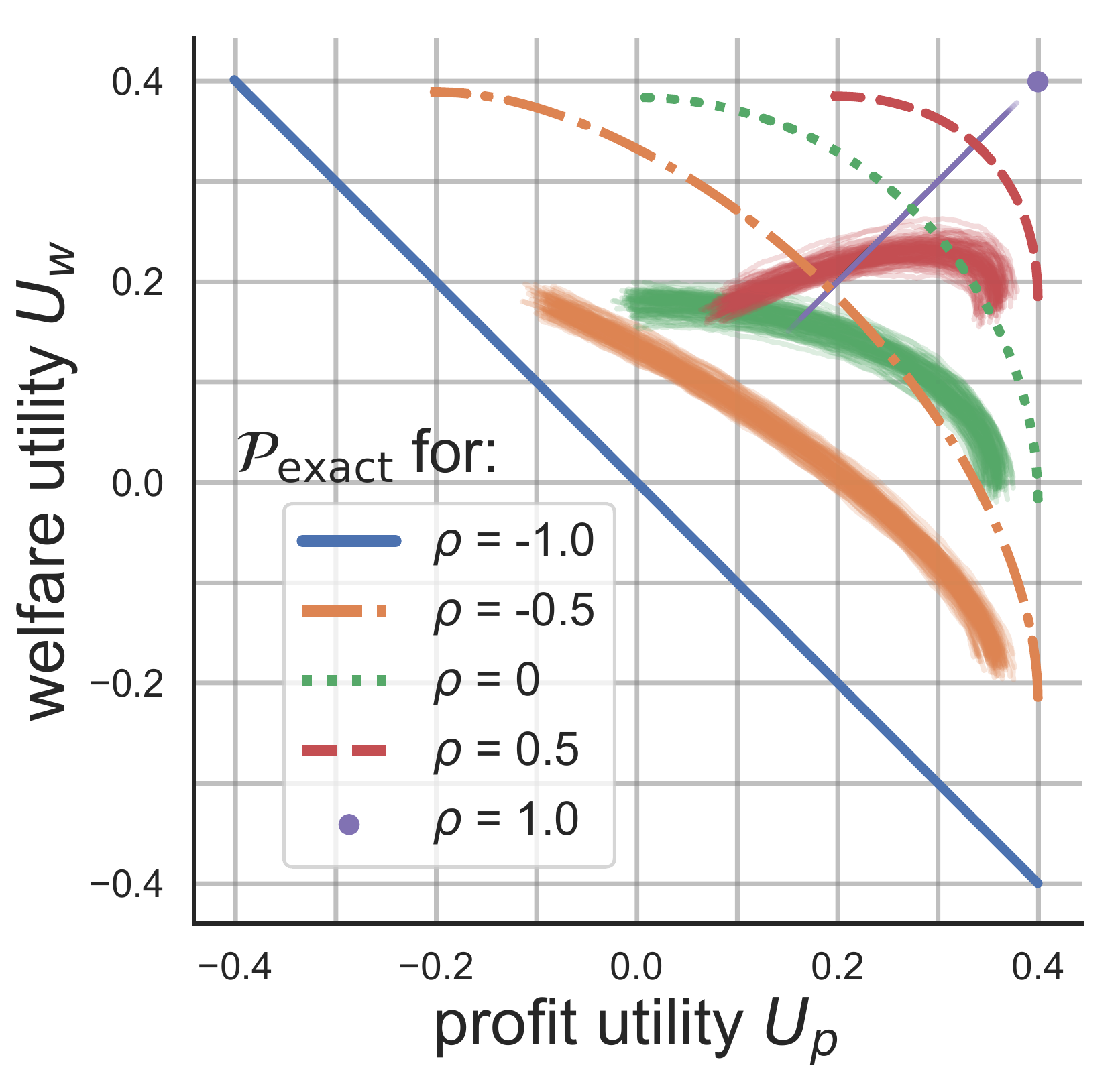}}
        \caption{ Varying correlation, fixed and unequal noise $\sigma_{\varepsilon_p} = 0.5$,  $\sigma_{\varepsilon_w} = 2.0$.
            \label{fig:corr_scores_differing_noise}}
    \end{subfigure}
        \caption{Simulated experiments corresponding to the setting in \Cref{ex:gaussian_scores_subgaussian_error} (fixing $\sigma_w = \sigma_p = 1$).
        Empirical frontiers $\Pareto(\Pi_{\emp})$ 
        for 100 random trials with $n=5,000$ each are shown  as overlaid translucent  curves.
        Exact frontiers $\Pareto_\exact$ are shown as dashed curves.
        }
\end{figure*}
}
\subsection{Bounding Pareto Inefficiencies}
\label{sec:bounds}
Even when plug-in policies are not optimal, 
the sub-optimality of the resulting classifier in terms of the utility function $\util_{\alpha}$ is bounded by the $\alpha$-weighted sum of $\ell_1$ errors in the profit and welfare scores. 

\begin{prop}[Sub-optimality Bound]
\label{prop:suboptimal} For any score prediction functions $\fphat(x),\fwhat(x)$  and $\alpha \in [0,1]$, the gap in $\alpha$-utility from applying the plug-in policy~\eqref{eq:empirical_score_threshold} with $\fphat(x), \fwhat(x)$ versus applying the optimal policy~\eqref{eq:alpha_pareto_score_pol} with true scores $\fp, \fw$, is bounded above as
\iftoggle{arxiv}{
	\begin{align}
	\util_\alpha(\polst_{\alpha}) - \util_\alpha(\polhat_{\alpha})  \le (1-\alpha) \Exp[|\fphat(x) - \fp(x)|] +\alpha\Exp[|\fwhat(x) - \fw(x)|]. \label{eq:suboptimality_bound} 
	\end{align}}{
    \begin{multline}
    \util_\alpha(\polst_{\alpha}) - \util_\alpha(\polhat_{\alpha})  \le \\(1-\alpha) \Exp[|\fphat(x) - \fp(x)|] +\alpha\Exp[|\fwhat(x) - \fw(x)|]. \label{eq:suboptimality_bound} 
    \end{multline}
}
\end{prop}

Note that by definition of $\polst_{\alpha}$, $ \util_\alpha(\polst_{\alpha})- \util_\alpha(\polhat_{\alpha}) \geq 0$. The proof of \Cref{prop:suboptimal} is given in Appendix~\ref{sec:subopt}.

\Cref{prop:suboptimal} provides a general bound on the $\alpha$-performance of the plug-in policy which holds for any distribution on scores and estimator errors. 
To provide further insight,
we consider a specific distributional setting.


\begin{example}
\label{ex:gaussian_scores_subgaussian_error}
Suppose that individuals' true scores are distributed as:  
\begin{align}
\label{eq:gaussian_scores}
(w_i,p_i ) \sim_{i.i.d.} \calN \left(\begin{bmatrix} 0 \\ 0 \end{bmatrix},
\begin{bmatrix} \sigma^2_w & \rho \sigma_w \sigma_p \\ 
\rho \sigma_w \sigma_p  & \sigma^2_p \end{bmatrix} \right)
\end{align}
Let the prediction errors  $\varepsilon_{p_i} := \hat{p}_i - p_i$ and  $\varepsilon_{w_i} : = \hat{w}_i - w_i$ be independent of the true scores $p_i,w_i$, zero-mean, and sub-Gaussian with  parameters $\sigma_{\varepsilon_p}$ and $\sigma_{\varepsilon_w}$, respectively.
\end{example}
This example elucidates how correlation between profit and welfare scores affects the empirical Pareto frontier.
\begin{prop}
\label{claim:gaussian_score_bound}
In the setting of Example~\ref{ex:gaussian_scores_subgaussian_error} with $-1 \leq \rho \leq 1$, 
$\Exp[\Util_\alpha(\polst_{\alpha})] = \frac{\sigma_y}{\sqrt{2\pi}}$ and the expected $\alpha$-utility of the plug in policy is at least:~\footnote{The constant on $\tilde{\sigma}^2$can be reduced to $1$ when $\varepsilon_{w} \indep \varepsilon_{p}$.}
\begin{align}
\label{eq:lower_bound_gaussian}
\Exp[\Util_\alpha(\polhat_{\alpha})] 
&\geq \Exp[\Util_\alpha(\polst_{\alpha})] \left( 1 - \frac{2 \cdot \tilde{\sigma}^2}{\tilde{\sigma}^2 + \sigma^2_y}\right) 
\end{align}
where 
{{$\sigma^2_y = \alpha^2\sigma^2_w + (1-\alpha)^2 \sigma^2_p + 2 \rho \alpha (1-\alpha) \sigma_w \sigma_p$}} and $\tilde{\sigma}^2 = 4(\alpha^2 \sigma^2_{\varepsilon_{w}} + (1-\alpha)^2 \sigma^2_{\varepsilon_{p}})$. 
\end{prop}

The proof of \Cref{claim:gaussian_score_bound} is given in Appendix~\ref{sec:subopt}.
This lower bound is in terms of both the optimal $\alpha$-utility and a discount factor. Because $\sigma^2_y$ is increasing in $\rho$ for any $\alpha \in (0,1)$, both of these terms are increasing in $\rho$. Thus, the expected $\alpha$-utility of the plug in policy is higher for correlated scores, not only because the optimal $\alpha$-utility is higher, but also because the discount factor is closer to 1.

\begin{figure}
    \centering
    \includegraphics[height=2.2in]{\figpath 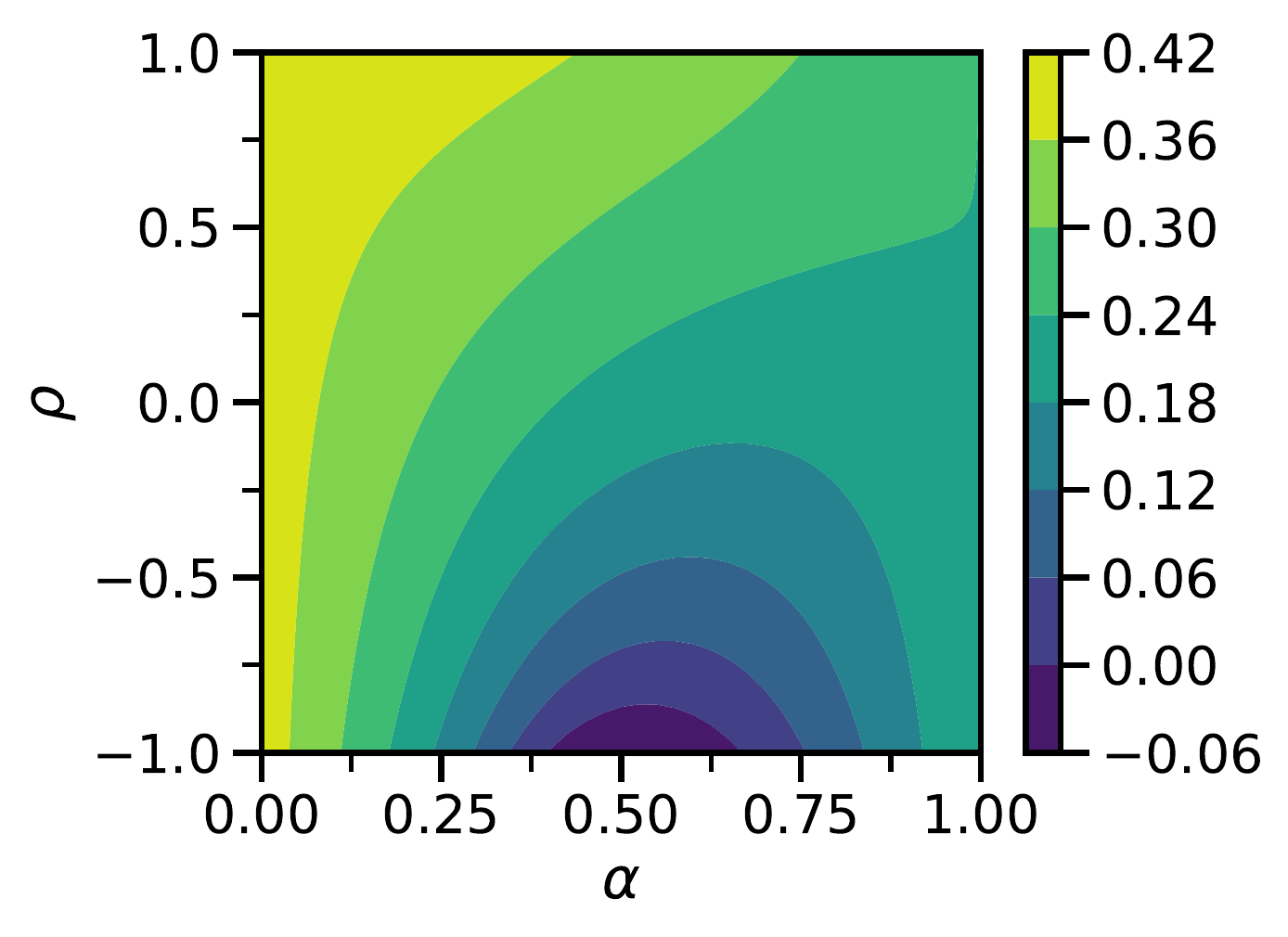}
        \caption{ Lower bound (right hand side of Eq.~\eqref{eq:lower_bound_gaussian}) on expected $\alpha$-utility as a function of $\alpha$ and correlation in the true scores, from~\Cref{claim:gaussian_score_bound}, with $\sigma_w = \sigma_p = 1$; $\sigma_{\varepsilon_w} = .5$; $\sigma_{\varepsilon_p}=0.1$. \label{fig:lower_bound} }
\end{figure}

Figure~\ref{fig:lower_bound} shows the lower bound on expected $\alpha$-utility with noisy scores as a function of possible score correlations $\rho$ and trade-off parameters $\alpha$, for a fixed setting of predictor noise in~\Cref{ex:gaussian_scores_subgaussian_error}.  For comparatively small error in profit scores and moderate welfare error, the lower bound on the $\alpha$-utility increases as the correlation ($\rho$) between the scores increases. This captures how the low-noise profit score indirectly improves decisions about the high-noise welfare. The lower bound is decreasing in $\alpha$ for positive $\rho$, which reflects the higher variance introduced by placing more weight on the noisier welfare score.


%% file: experiments.tex

 This section presents three sets of empirical results. 
 In Section~\ref{sec:sim_experiments} we corroborate our theoretical results 
 under different simulated distributions on scores and prediction errors. 
 Our second experiment studies empirical Pareto frontiers from learned scores with realistic degradation of training data, in the context of sustainable abalone collection in Section~\ref{sec:abalone}. 
 Our third experiment in Section~\ref{sec:youtube} shows how our methods facilitate trading off between user engagement with predicted quality of content in a corpus of YouTube videos, using pre-learned scores.

\subsection{Simulation Experiments}
\label{sec:sim_experiments}

Our first set of simulations shows the performance of the plug-in policy 
when scores are perturbed by additive noise of varying degrees in each dimension (Fig.~\ref{fig:uncorrelated_plus_noise}). 
We instantiate true scores $ \welfscore_i$ and $\profscore_i$  as in Eq.~\eqref{eq:gaussian_scores} with $\rho = 0$ and $\sigma^2_w = \sigma^2_p = 1$, and instantiate predicted scores as:
\begin{align}
\label{eq:sim_add_noise}
\fwhat(x_i) = \welfscore_i + \varepsilon_{w_i} \quad \varepsilon_{w_i}\sim \calN(0,\sigma^2_{\varepsilon_\welfscore}), \\
\fphat(x_i) = \profscore_i + \varepsilon_{p_i} \quad \varepsilon_{p_i} \sim \calN(0,\sigma^2_{\varepsilon_\profscore}) \nonumber
\end{align}
These score predictions satisfy the \emph{well-calibrated} condition of~\Cref{cor:well_calibrated}.
The results for different pairs ($\sigma_{\varepsilon_\welfscore},\sigma^2_{\varepsilon_\profscore}$ are shown in Figure~\ref{fig:uncorrelated_plus_noise}. As the noise in scores increases, the empirical Pareto frontiers recede from the exact frontier $\Pareto_\exact$. Additionally, higher noise in the predicted scores imposes a wider distribution of empirical Pareto frontiers. 

Next, we study the effect of noise in predictions when scores are correlated (Fig.~\ref{fig:corr_scores_fixed_noise}). We draw $\welfscore_i$ and $\profscore_i$ according to Eq.~\eqref{eq:gaussian_scores} with $\sigma_w = \sigma_p = 1$ and correlation parameter $\rho$. We then add random noise as in Eq.~\eqref{eq:sim_add_noise} with parameters $\sigma_{\varepsilon_\welfscore }= \sigma_{\varepsilon_\profscore} = 1.0$. Note that in this setting, scores are in general not calibrated due to the correlation between $\welfscore_i$ and $\profscore_i$.
For positive values of $\rho$, the exact and empirical utilities are greatest at $\alpha=0.5$, since the correlation in the scores allows us to overcome some of the noise in each individual parameter, as predicted by~\Cref{claim:gaussian_score_bound}.

Lastly, we study the space of empirical and exact frontiers with degraded noise when scores are correlated and prediction error is higher in the welfare the score, with $\sigma_{\varepsilon_p} = 0.5$ whereas $\sigma_{\varepsilon_w} = 2.0$ (Fig.~\ref{fig:corr_scores_differing_noise}). 
While the optimal Pareto frontiers are the same as in Fig.~\ref{fig:corr_scores_fixed_noise}, we see a stark change in the empirical Pareto frontiers. 
Compared to the case of no correlation, the empirical Pareto frontier is expanded when $\rho > 0$ and when $\rho < 0$ the frontier recedes. 
Additionally, we see evidence that due to the correlation, $\polhat_\alpha$ is no longer guaranteed to be optimal, as welfare utility decreases for large enough $\alpha$ when $\rho= 0.5$.

\subsection{Learned Scores with Imperfect Data: Abalone}
\label{sec:abalone}

Our next example is motivated by the domain of ecologically sustainable selection, where the goal is to select profitable mollusks to catch and keep, while having minimal impact on the natural development of the mollusks' ecosystem. 
We learn scores for the age and profitability of each abalone from data, and perform experiments to test the degradation of the empirical Pareto frontiers under realistic degradations of the data.  
While our characterization of the problem is highly simplified, the main focus of this experiment is to demonstrate the instantiation of Pareto curves for different predictor function classes and different regimes of data availability.

The welfare measure we use is an increasing function of age (see Appendix~\ref{sec:experiment_details} for full experimental details),
encoding that it is more sustainable to harvest older abalones.  We define the profit score of each abalone as a linear function of meat weight and shell area. We use the features (sex, total weight, height, width, and diameter) to train score predictors.  We derive these measures from physical data collected by
~\citet{nash1994population} (accessed via the UCI data repository~\citep{Dua:2019}). The correlation of the profit and welfare scores is $0.56$.

\iftoggle{arxiv}{
\begin{figure}[ht!]
    \centering
    \begin{subfigure}[h]{.45\textwidth}
        \centering
        \includegraphics[height=3in]{\figpath 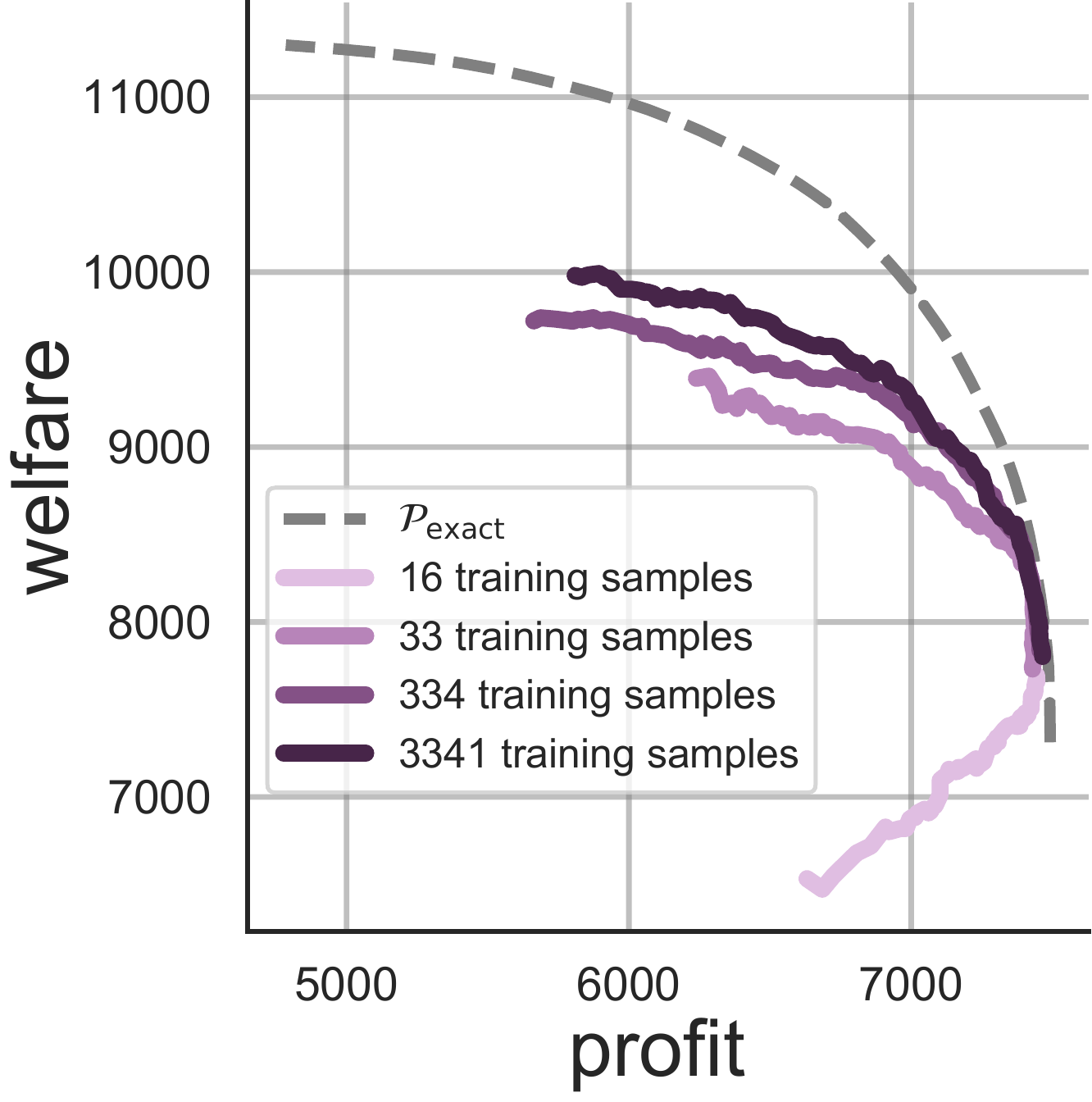}
        \caption{Ridge regression model. \label{fig:abalone_1_ridge}}
    \end{subfigure}%
    ~~
    \begin{subfigure}[h]{=.45\textwidth}
        \centering
        \includegraphics[height=3in]{\figpath 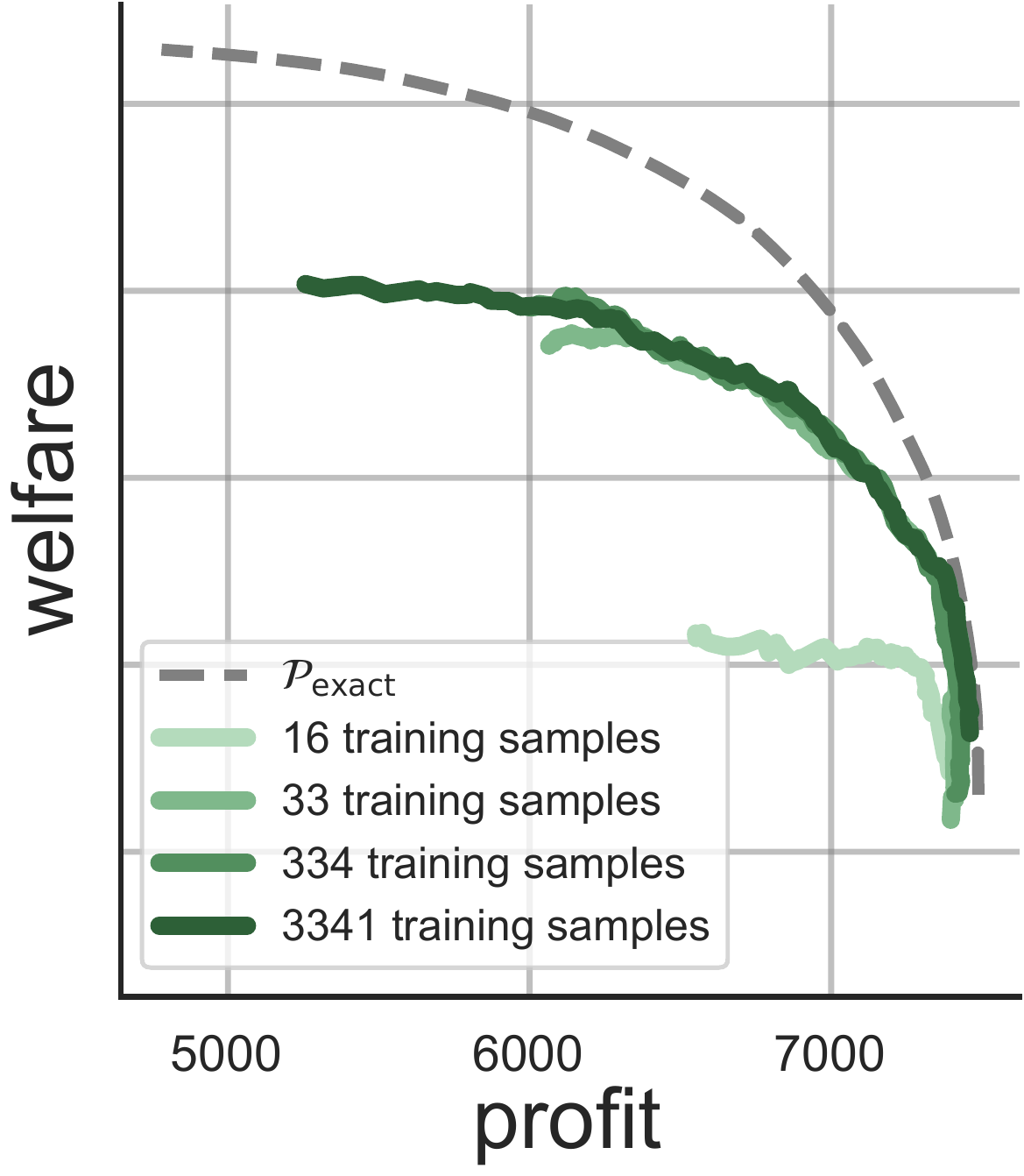}
        \caption{Random forest model. \label{fig:abalone_n_forest}}
    \end{subfigure}
        \caption{Abalone empirical frontiers as training set size increases. \label{fig:abalone_n}}
\end{figure}
\begin{figure}[ht!]
    \centering
    \begin{subfigure}[h]{.49\linewidth}
        \centering
        \includegraphics[height=3in]{\figpath 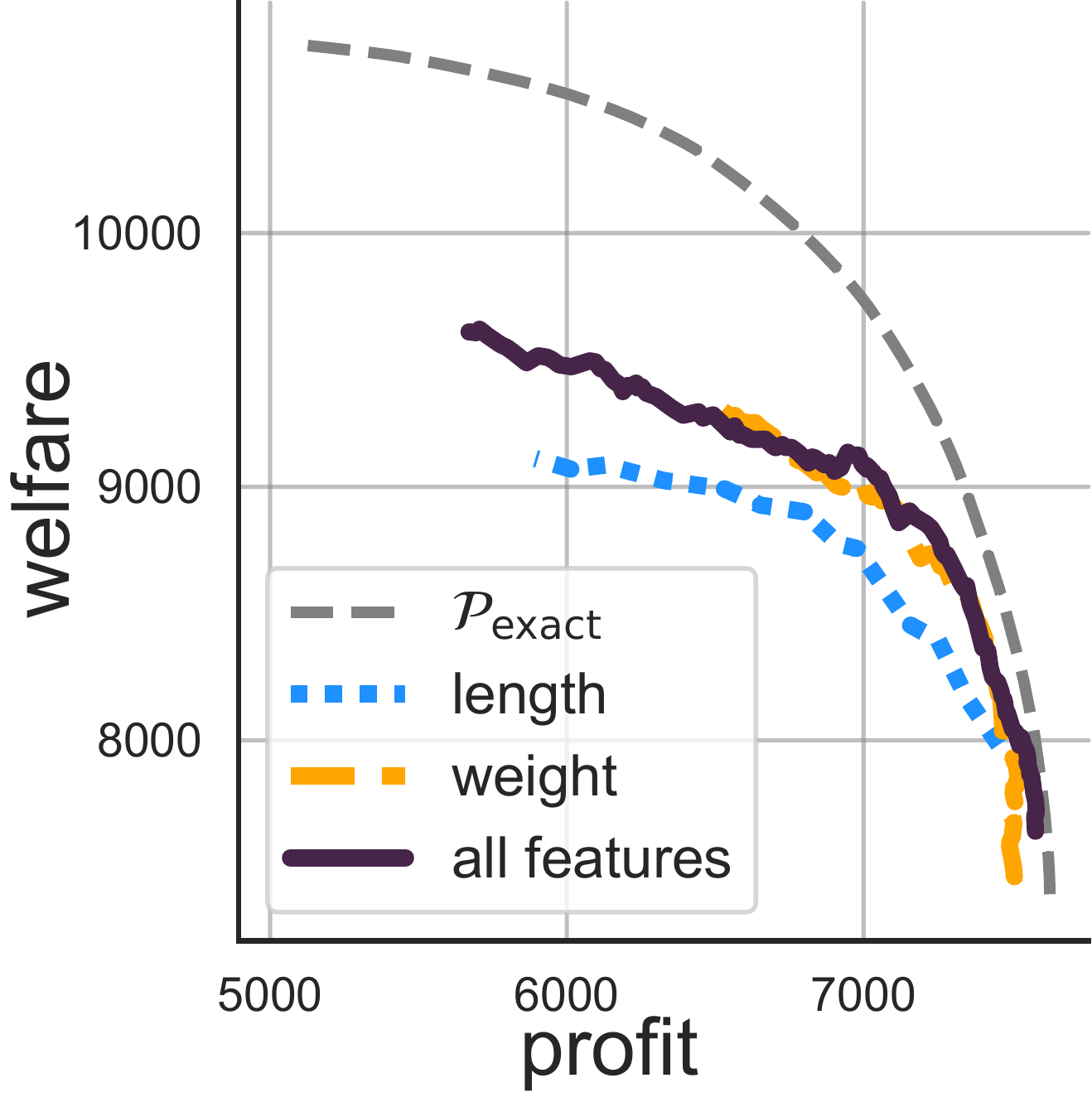}
        \caption{Ridge regression model. \label{fig:abalone_d_ridge}}
    \end{subfigure}%
    ~ 
    \begin{subfigure}[h]{.49\linewidth}
        \centering
        \includegraphics[height=3in]{\figpath 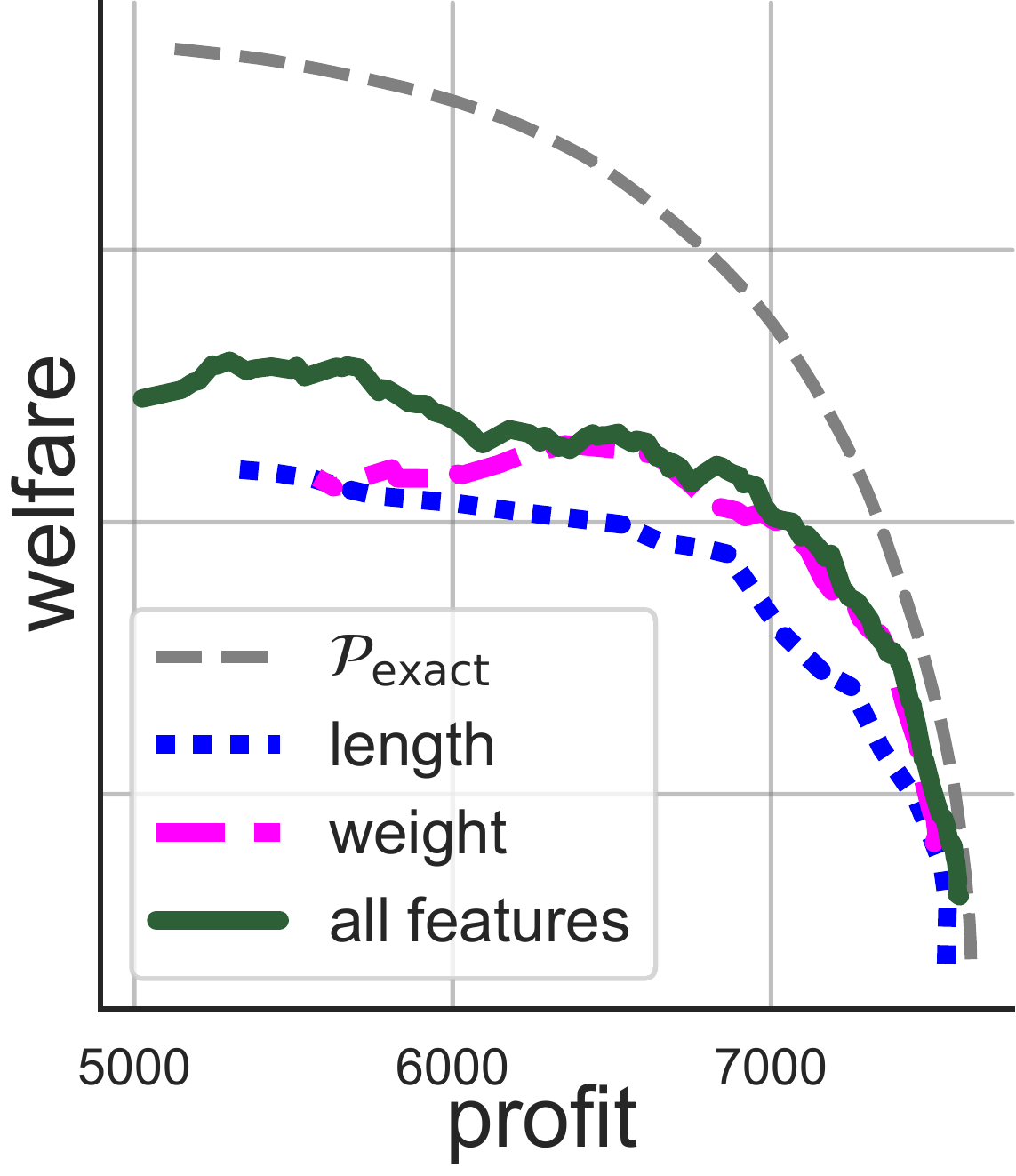}
        \caption{Random forest model. \label{fig:abalone_d_forest}}
    \end{subfigure}
        \caption{Abalone empirical frontiers for different feature sets.  \label{fig:abalone_d}}
\end{figure}
}
{   
\begin{figure}[ht!]
    \centering
    \begin{subfigure}[h]{.49\linewidth}
        \centering
        \includegraphics[height=1.6in]{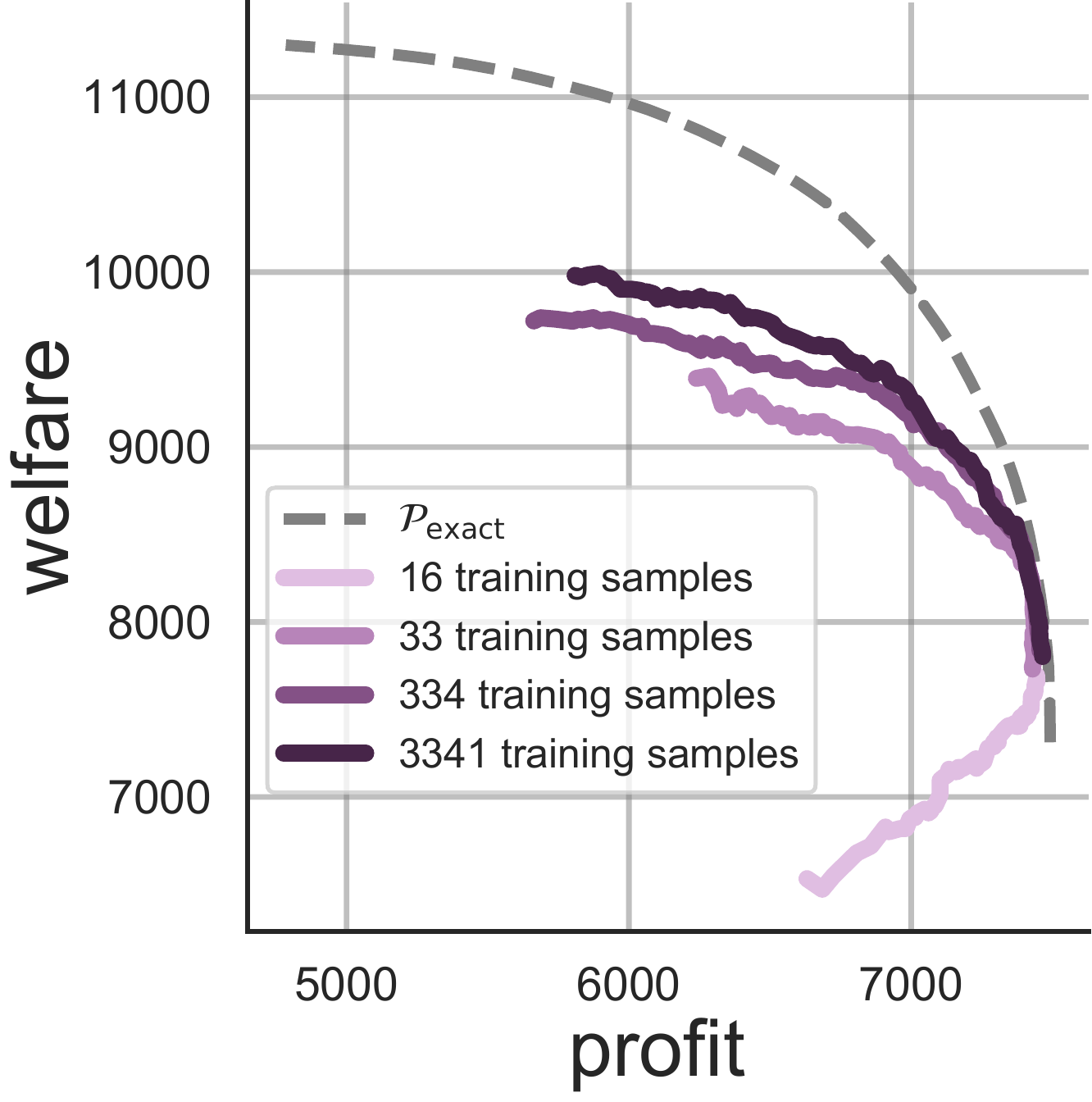}
        \caption{Ridge regression model. \label{fig:abalone_1_ridge}}
    \end{subfigure}%
    ~ 
    \begin{subfigure}[h]{.49\linewidth}
        \centering
        \includegraphics[height=1.6in]{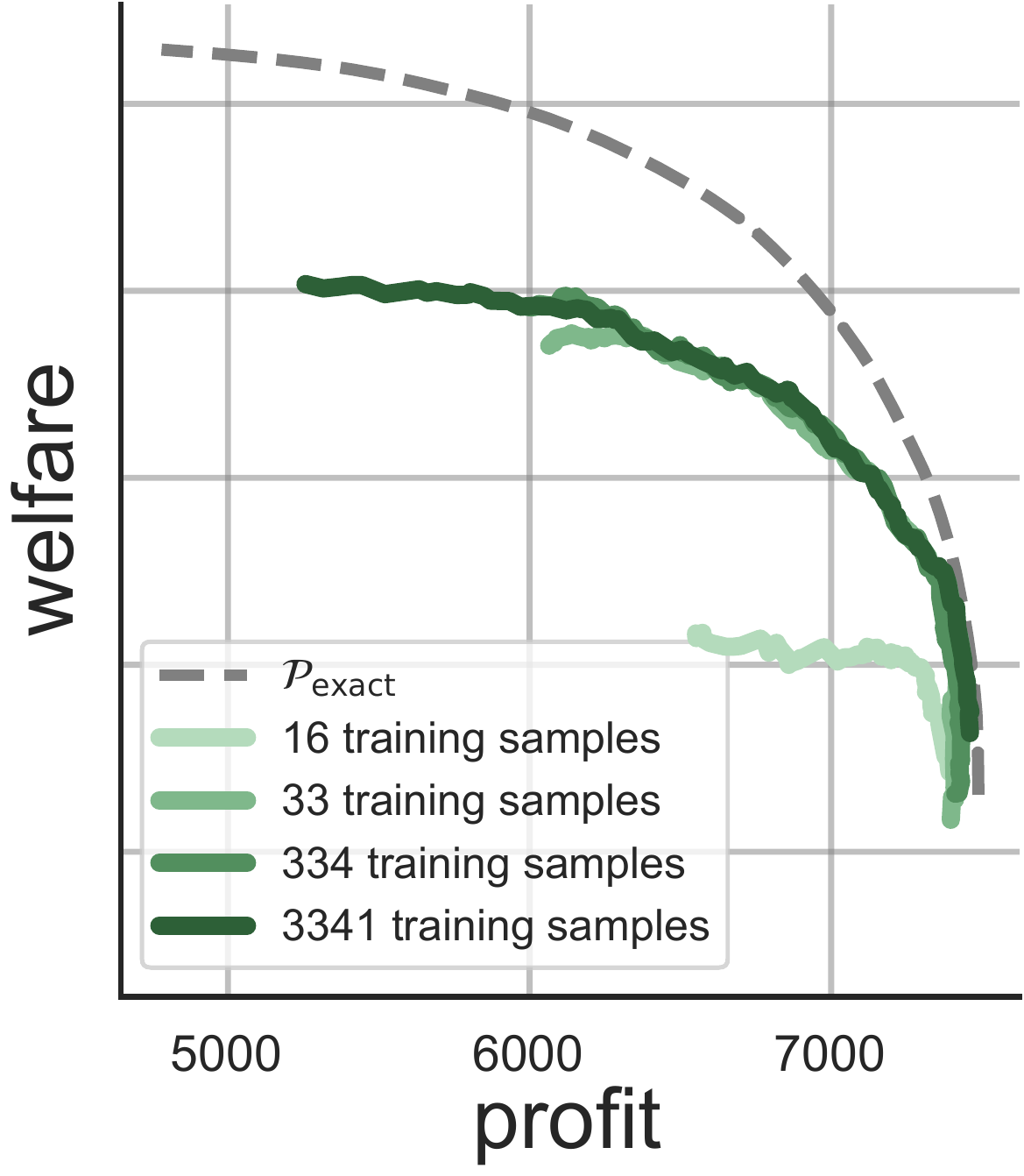}
        \caption{Random forest model. \label{fig:abalone_n_forest}}
    \end{subfigure}
        \caption{Abalone empirical frontiers as training set size increases. \label{fig:abalone_n}}
\end{figure}
\begin{figure}[ht!]
    \centering
    \begin{subfigure}[h]{.49\linewidth}
        \centering
        \includegraphics[height=1.5in]{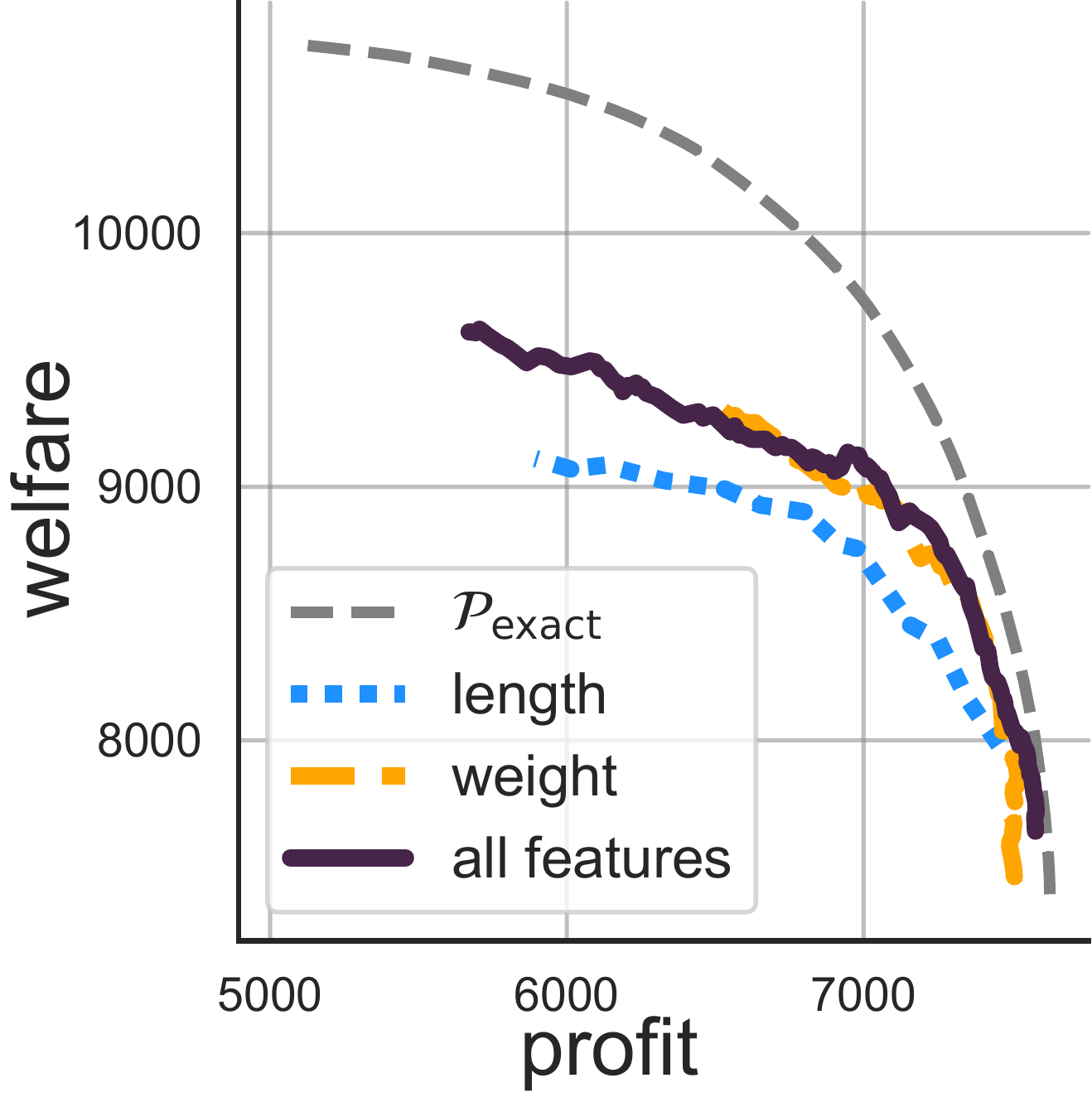}
        \caption{Ridge regression model. \label{fig:abalone_d_ridge}}
    \end{subfigure}%
    ~ 
    \begin{subfigure}[h]{.49\linewidth}
        \centering
        \includegraphics[height=1.5in]{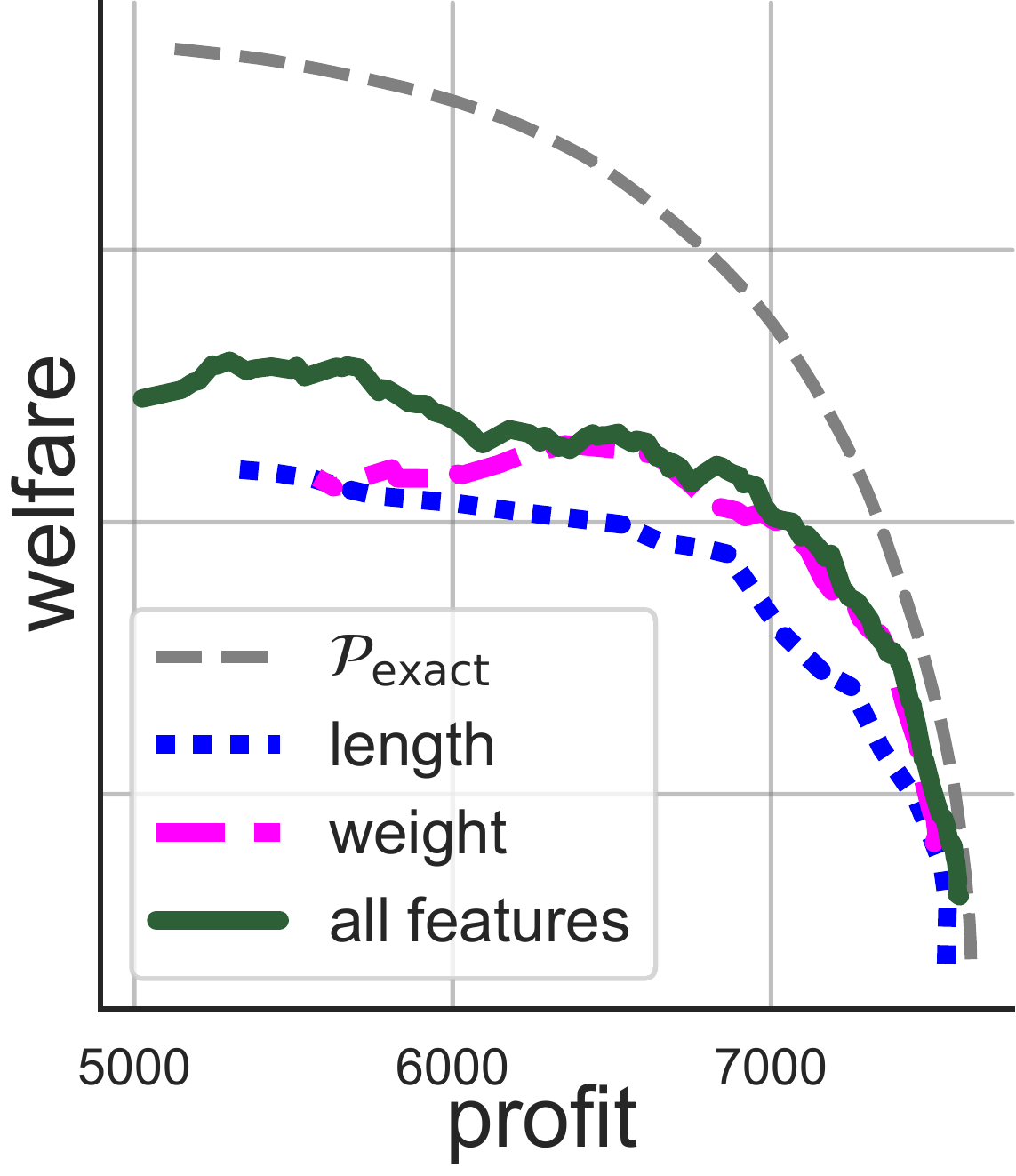}
        \caption{Random forest model. \label{fig:abalone_d_forest}}
    \end{subfigure}
        \caption{Abalone empirical frontiers for different feature sets.  \label{fig:abalone_d}}
\end{figure}
}

\iftoggle{arxiv}{
\begin{figure}[ht!]
\centering
        \includegraphics[height=1.6in]{\figpath 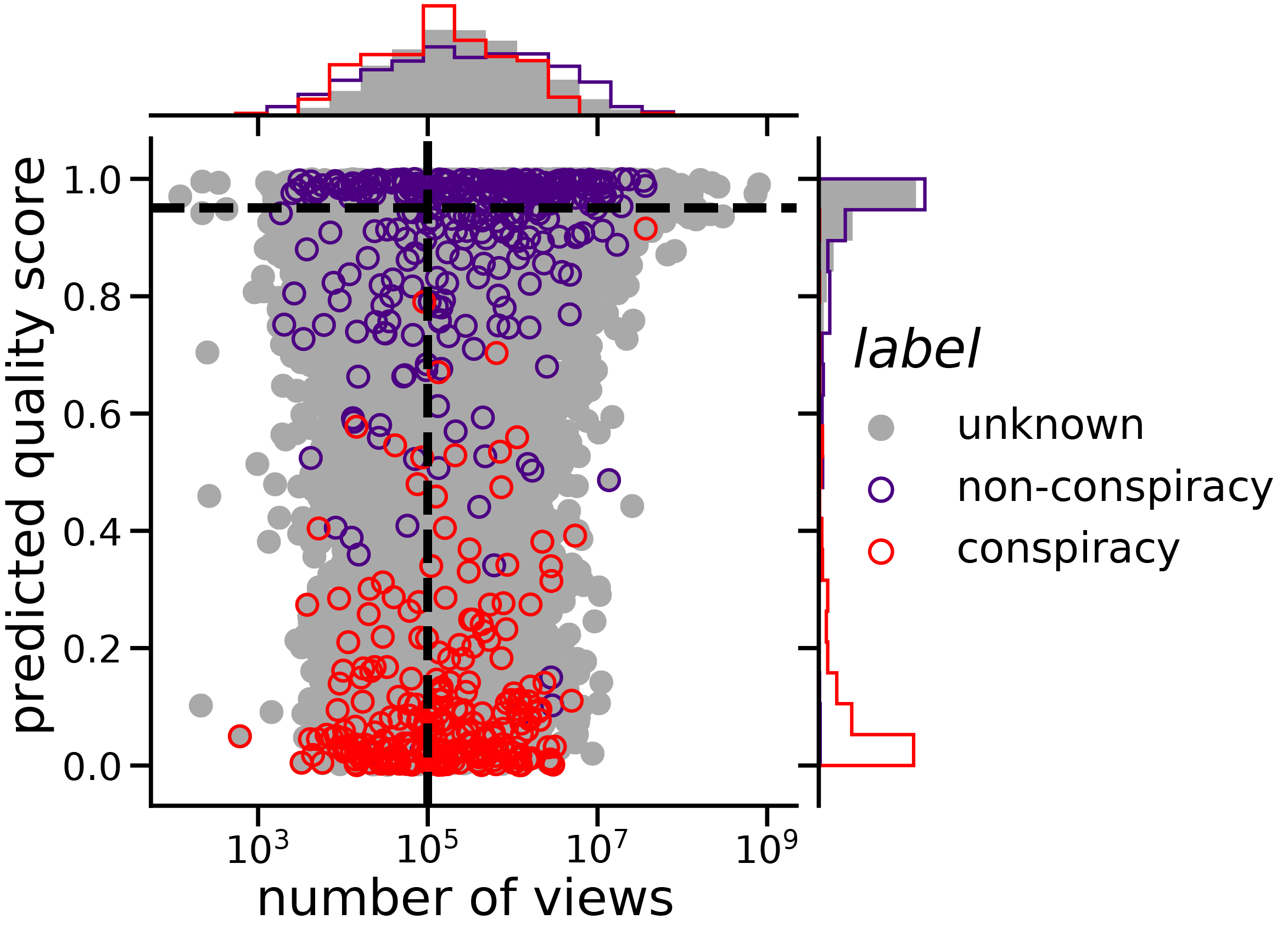} 
        \caption{Distribution of YouTube data predicted quality scores unlabeled videos (gray), and hand labeled conspiracy (red) and non-conspiracy (purple) videos. \label{fig:yt_training_dist}}
\end{figure}

\begin{figure*}[t]
\centering
    \begin{subfigure}[h]{.42\textwidth}
        \centering
        \includegraphics[height=2.in]{\figpath 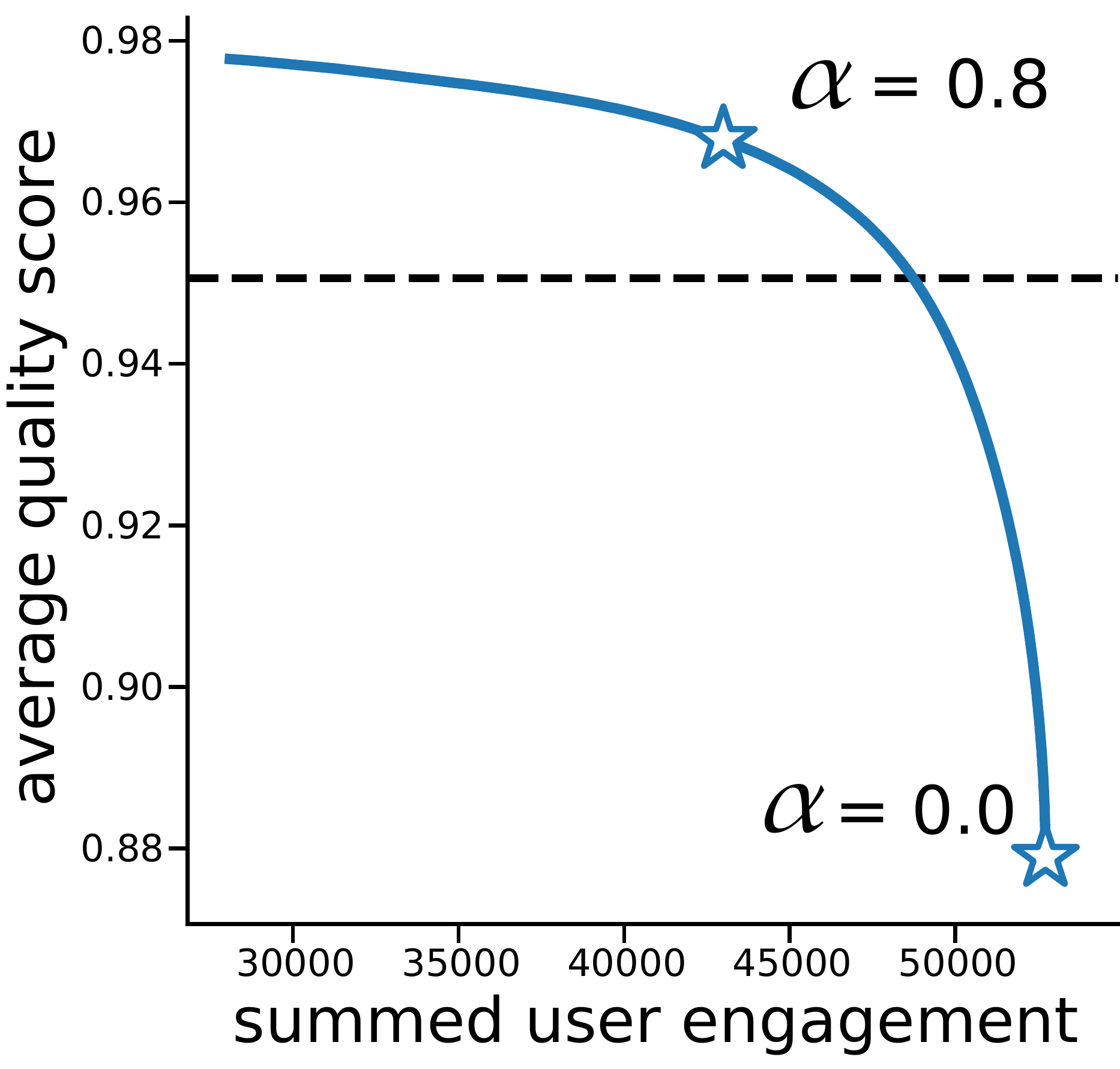}
        \caption{ Estimated Pareto curve using pre-computed predictions. Stars indicate specific $\alpha$ trade-offs.  \label{fig:youtube_pareto_no_labels}}
    \end{subfigure}%
    ~~~
    \begin{subfigure}[h]{.48\textwidth}
        \centering
        \includegraphics[height=2in]{\figpath 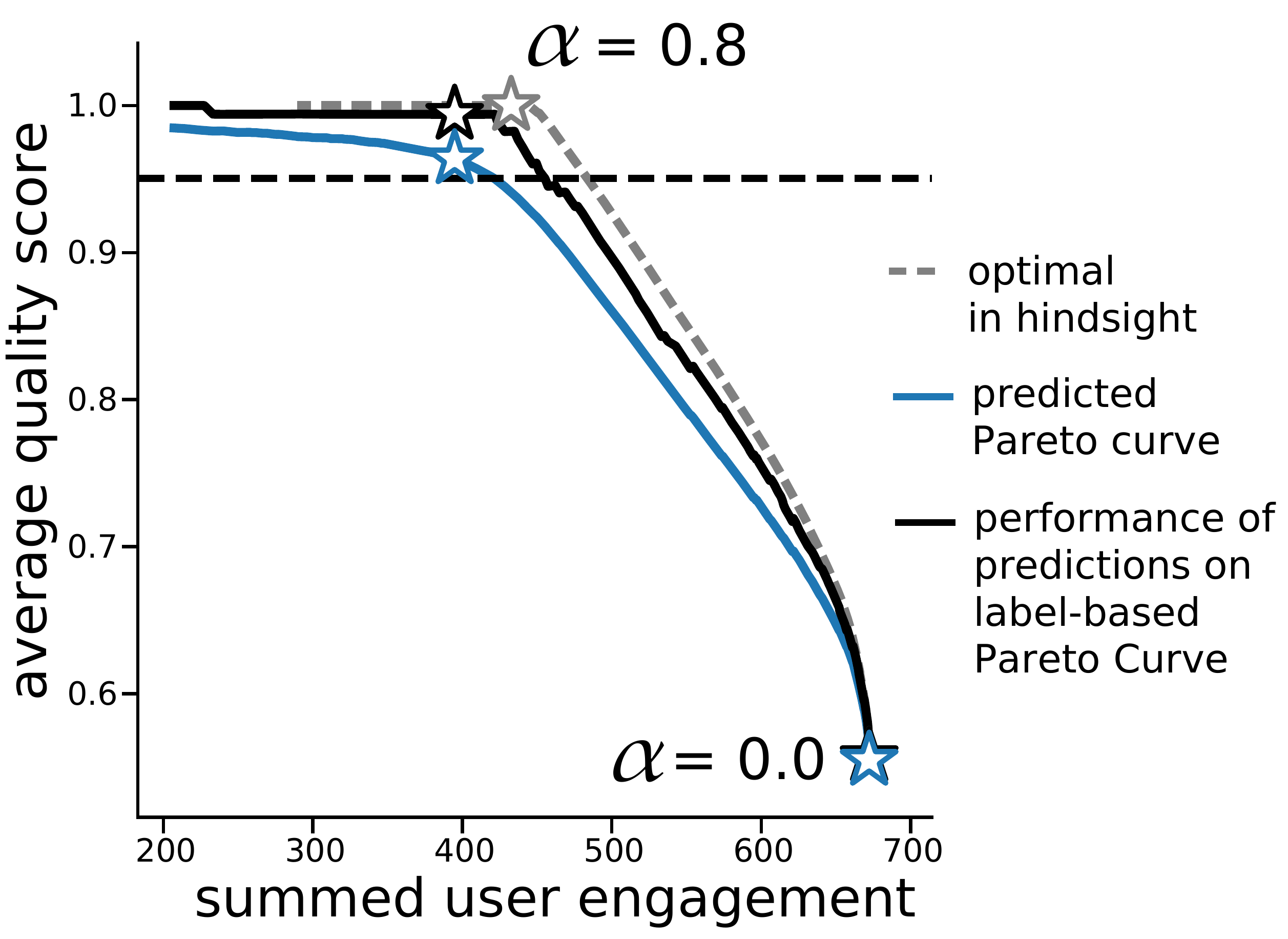}
        \caption{ Estimated Pareto curve on labeled data subset (blue). Optimal-in-hindsight curve (dashed gray) and performance of predictions on label set (black). \label{fig:youtube_pareto_with_labels}}
    \end{subfigure}
        \caption{Balancing user engagement and health of hosted YouTube videos.  }
\end{figure*}
}
{   
\begin{figure*}[t]
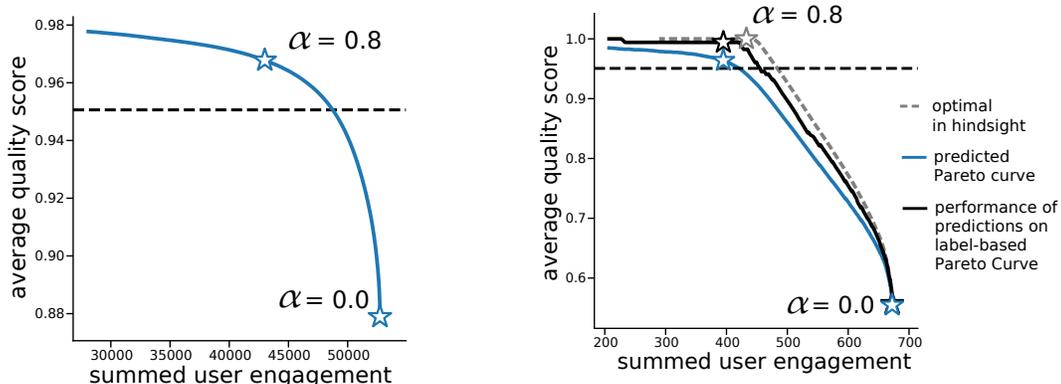

    \centering
   \begin{subfigure}[h]{.30\textwidth}
        \includegraphics[height=1.6in]{\figpath youtube_data_dist_labels.png} 
        \caption{Distribution of YouTube data predicted quality scores unlabeled videos (gray), and hand labeled conspiracy (red) and non-conspiracy (purple) videos. \label{fig:yt_training_dist}}
    \end{subfigure}
    ~~~
    \begin{subfigure}[h]{.27\textwidth}
        \centering
        \includegraphics[height=1.7in]{\figpath youtube_pareto.pdf}
        \caption{ Estimated Pareto curve using pre-computed predictions. Stars indicate specific $\alpha$ trade-offs.  \label{fig:youtube_pareto_no_labels}}
    \end{subfigure}%
    ~~~
    \begin{subfigure}[h]{.38\textwidth}
        \centering
        \includegraphics[height=1.7in]{\figpath youtube_pareto_with_labels.pdf}
        \caption{ Estimated Pareto curve on labeled data subset (blue). Optimal-in-hindsight curve (dashed gray) and performance of predictions on label set (black). \label{fig:youtube_pareto_with_labels}}
    \end{subfigure}
        \caption{Balancing user engagement and health of hosted YouTube videos.  }
\end{figure*}
}

In this setting, we study the effectiveness of two models --- ridge regression and random forests --- to learn scores with which to instantiate the plug-in policy. To assess how the empirical Pareto frontiers degrade under realistic notions of imperfect data, we subsample training instances to reflect a hypothetical regime were data is sparse and we subsample features to reflect a hypothetical regime where entire measurements were not recorded in the original dataset.

Figure~\ref{fig:abalone_n} shows the empirical Pareto frontiers reached as we change the size of the training data set from which learn the profit and welfare scores. Even with $33$ training samples ($1\%$ of the original training set), the set of plug-in policies traces a meaningful trade-off over $\alpha$. For severely degraded scores (16 training samples - just $0.5\%$ of the original training sets), the error on the welfare score predictions is so high that instantiating a plug-in policy with $\alpha >0$ actually decreases welfare overall.

Figure~\ref{fig:abalone_d} shows the empirical Pareto frontiers reached as we change the features learned to train the model, using just length, just weight, or all seven features as in Fig.~\ref{fig:abalone_n}.

The trends to increasing the data set size and feature set are consistent with four replications done on separate  training and evaluation splits; we find that Pareto frontiers dominate each other roughly in accordance with the mean average error of the score predictions (Figs.~\ref{fig:abalone_all_n} and ~\ref{fig:abalone_all_d} in Appendix~\ref{sec:abalone_details}). 
The mean average error of welfare scores is substantially greater than the average error of profit scores for most prediction instances (Figures~\ref{fig:abalone_all_n} and~\ref{fig:abalone_all_d} in Appendix~\ref{sec:abalone_details}), thus the empirical frontiers are farther from $\Pareto_\exact$ in the welfare dimension than the profit dimension.

Altogether, the empirical Pareto frontiers are relatively robust to small data regimes, as well as to missing predictors. However, when predictions have very high error (diagnosable by cross-validation or holdout set error), empirical Pareto frontiers degrade quickly.

\subsection{Balancing User Engagement and Health}
\label{sec:youtube}

We now illustrate how the multi-objective framework can be used to balance the desire to promote high quality content with the need for profit. We work with a dataset that contains measures of content quality and content engagement for 39,817 YouTube videos, which was constructed as part of an independent effort to automatically ascertain the quality and truthfulness of YouTube videos \citep{faddoul_longitudinal_2019}.

The measure of quality $\fwhat$ we use is a function of the `conspiracy score' developed by \citet{faddoul_longitudinal_2019}, which estimates the probability that the video promotes a debunked conspiracy theory. 
From this score $s_{\textrm{conspiracy}} \in [0,1]$ we derive a predicted `quality score' as $(0.95 - s_{\textrm{conspiracy}})$ (see Appendix~\ref{sec:youtube_details} for details).

We instantiate the profit score $\fp[i]$ for video $i$ as $\log ( (1 + \textrm{\# views}[i])/100,000)$. Dividing by a large constant represents that videos with low view counts may not be profitable due to storage and hosting costs.
The resulting distribution over $\fp$ and $\fwhat$ is shown in Figure~\ref{fig:yt_training_dist} (gray dots), where dotted lines denote $0$-utility thresholds in each score. 

Using these scores and predictions, we estimate a Pareto frontier using the optimal policies $\polhat_\alpha$ for learned scores from Eq.~\eqref{eq:empirical_score_threshold}. The resulting estimated Pareto curve is shown in Figure~\ref{fig:youtube_pareto_no_labels}. 
The curve is concave, demonstrating the phenomenon of diminishing returns in the trade-off between total user engagement and average video quality. 
While there is always some quality to gain by sacrificing some total engagement, these relative gains are greatest when the starting point is close to an engagement-maximizing policy.
Specifically, at the  maximum-engagement end of the spectrum (lower right star), 
we can gain a %
$1.1 \%$ increase in average video quality for a 
$0.1 \%$ loss in total engagement. 
However, for a policy with trade-off rate $\alpha=0.8$ (upper left star), to obtain an increase of 
$0.3 \%$ in welfare,
 a larger loss of
 $5.2 \%$ in user engagement is required.

Next, we assess the validity of this estimated Pareto curve using the small set of 541 hand-labeled training set instances from which $s_{\textrm{conspiracy}}$ was learned. This assessment is likely optimistic due to the fact that the score predictor functions were trained on this same data; nonetheless, this is an important check to perform on the estimated Pareto frontier. 

In Figure~\ref{fig:youtube_pareto_with_labels} we plot the optimal-in-hindsight Pareto frontier (dashed gray line) had we known the labels a priori and applied thresholds according to~\eqref{eq:alpha_pareto_score_pol}. We also plot the performance of our estimated policy $\polhat_\alpha$ on the labeled instances (black line). The stars on each curve correspond to decision thresholds with $\alpha = 0$ and $\alpha = 0.8$, and illustrate the alignment of the curves.

Relating back to Theorem~\ref{thm:pareto_inexact}, we see that performance of the learned scores (black line) is dominated by that of the optimal classifier, as is
the predicted Pareto curve (thick blue line).
Here the predicted Pareto curve under-predicts the actual performance; in general it is possible for the opposite to be true. 
Encouragingly, we observe that the curves representing the predicted and actual performance show similar qualitative trade-offs.

%% file: connections_to_fairness_shorter.tex

Having shown our main results on learning Pareto-optimal policies with limited data, we now illustrate connections between our framework and approaches based on fair machine learning that constrain classification decisions to satisfy certain criteria.
For example, in the setting of hiring or admissions, one might require that the same proportion of male and female candidates are admitted, i.e. \emph{demographic parity}.
We demonstrate that profit maximization with group fairness constraints corresponds to multi-objective optimization over profit and welfare for an induced definition of welfare.
This connection illustrates that even though we consider a welfare function defined from individual welfare scores, our framework can encode more collective conceptions of welfare, like those arising from group fairness constraints.

Consider the setting of requiring demographic parity between two subgroups $\popa$ and $\popb$ of a larger population (more general results are presented in Appendix~\ref{app:fairness}). 
In this case, we decompose policies over groups such that $\pol = (\pola, \polb)$.
Policies are chosen to maximize the following $\epsilon$-demographic parity constrained problem:
\begin{align}
\begin{split}\label{eq:fair_prob}
\max_{\pol,\beta} ~~\utilp(\pol)~~
	\mathrm{s.t.}~& \Exp[\polj(x) ~| x~\text{in group } \popj] = \beta_\popj, \\
	& |\beta_\popa - \beta_\popb| \leq \epsilon
\end{split}
\end{align}
We can
restrict our attention to threshold policies $\polj(p) = \I(p\geq t_\popj)$ where $t_\popj$ are group-dependent thresholds~\citep{liu18delayed}. 
Notice that the unconstrained solution would simply be $\pol^\maxprof(p) = \I(p\geq 0)$ for all groups.
For this reason, we consider groups with $t_\popj < 0$ as comparatively \emph{disadvantaged} (since their threshold increases in the absence of fairness constraints) and $t_\popj > 0$ as \emph{advantaged}.
Then, the  multi-objective framework provides an additional perspective on the trade-offs between $\epsilon$-fairness and profit.
\begin{cor}
\label{corr:epsilon_decreases_alpha}
It is possible to define fixed welfare scores such that
the family of inexact fair policies parametrized by any $\epsilon \geq 0$ in~\eqref{eq:fair_prob} corresponds to a family of Pareto-optimal policies parametrized by $\alpha(\epsilon)$. The group-dependent welfare scores are such that $w\geq0$ for all individuals in the disadvantaged group and $w\leq0$ in the advantaged group. Furthermore, the 
induced trade-off parameter $\alpha(\epsilon)$ increases as $\epsilon$ decreases. 
 \end{cor}
\Cref{corr:epsilon_decreases_alpha} follows from Theorem~\ref{claim:epsilon_decreases_alpha}.
Fairness constraints can be seen as encoding implicit group-dependent welfare scores for individuals, where members of disadvantaged groups are assigned positive welfare weights and members of advantaged groups assigned negative weights. 
Figure~\ref{fig:fairness} illustrates this result applied to data from a credit lending scenario from~\citet{barocas-hardt-narayanan}, where welfare scores are induced for individuals depending on their race and likelihood of repayment.
Further details on the generation of these weights are presented in Appendix~\ref{app:fairness}.
This correspondence is related to the analysis of welfare weights in~\citet{hu_welfare_2018}, however, our perspective focuses on trade-offs between welfare and profit objectives, in contrast to pure welfare maximization.  

In the case that group membership is believed to correspond to the welfare impact of selection,~\Cref{corr:epsilon_decreases_alpha} connects our results in Section~\ref{sec:methods} with a body of work on achieving fairness when group labels are approximate or estimated~\citep{kallus2020assessing}.
While some applications may directly call for statistical parity as a criterion, \Cref{corr:epsilon_decreases_alpha} emphasizes the inevitability of fairness constraints as trade-offs between multiple objectives, and frames these trade-offs explicitly in terms of welfare measures.

\iftoggle{arxiv}{
\begin{figure}[ht!]
    \centering
    \begin{subfigure}[t]{.4\linewidth}
        \centering
        \includegraphics[width=\linewidth]{\figpath 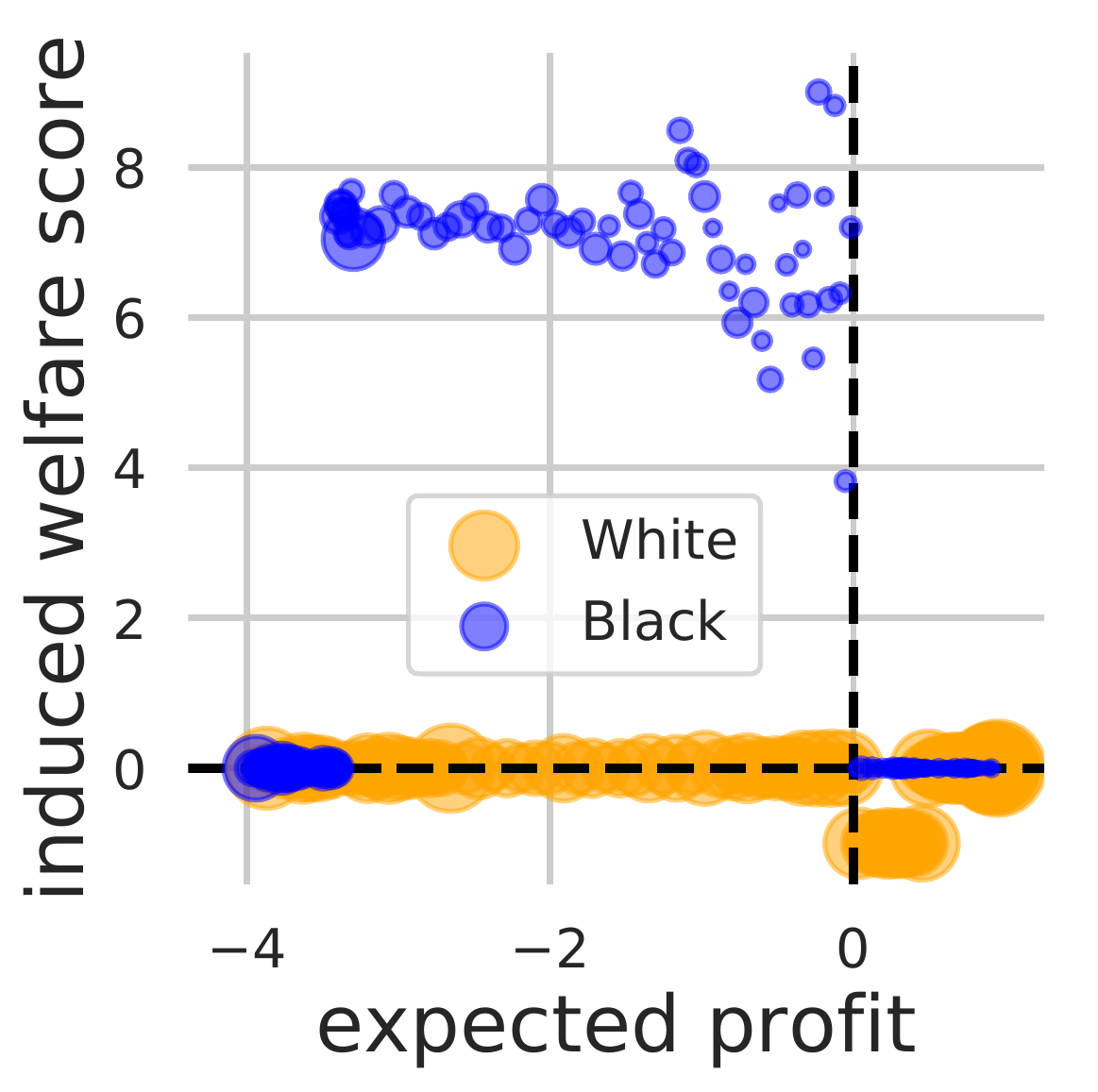}
        \caption{Distribution of profit and welfare scores. Marker size indicates population sizes.\label{fig:fairness_induced_weights}}
    \end{subfigure}%
    ~~
    \begin{subfigure}[t]{.48\linewidth}
        \centering
        \includegraphics[width=\linewidth]{\figpath 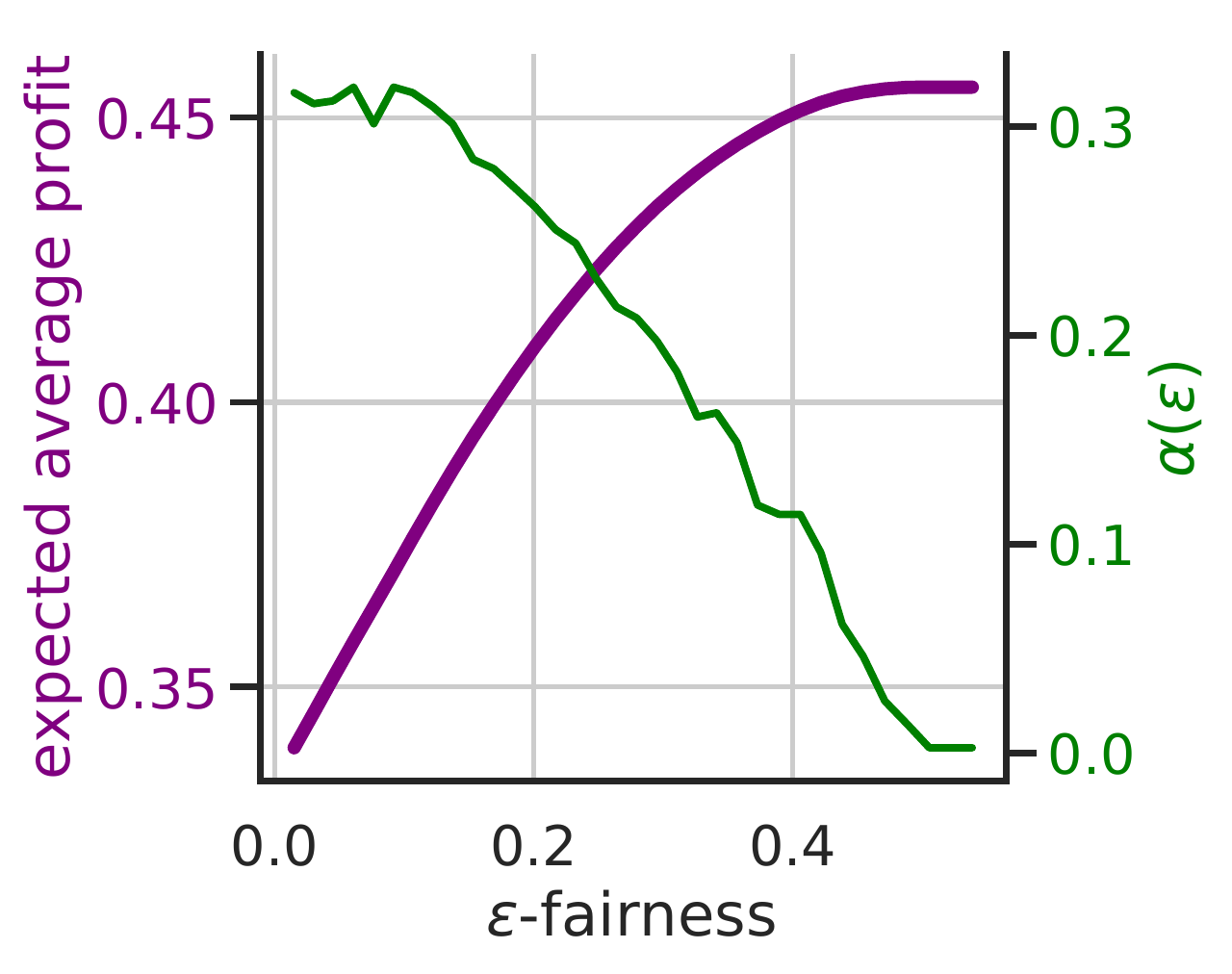}
        \caption{The fairness parameter $\epsilon$ determines the profit trade-off and corresponds to the welfare weight $\alpha$. \label{fig:fairness_alpha_epsilon}}
    \end{subfigure}
        \caption{Trade-offs between profit and fairness in lending can be equivalently encoded by a multi-objective framework. \label{fig:fairness}}
\end{figure}
}
{   
\begin{figure}[ht!]
    \centering
    \begin{subfigure}[t]{.44\linewidth}
        \centering
        \includegraphics[height=1.4in]{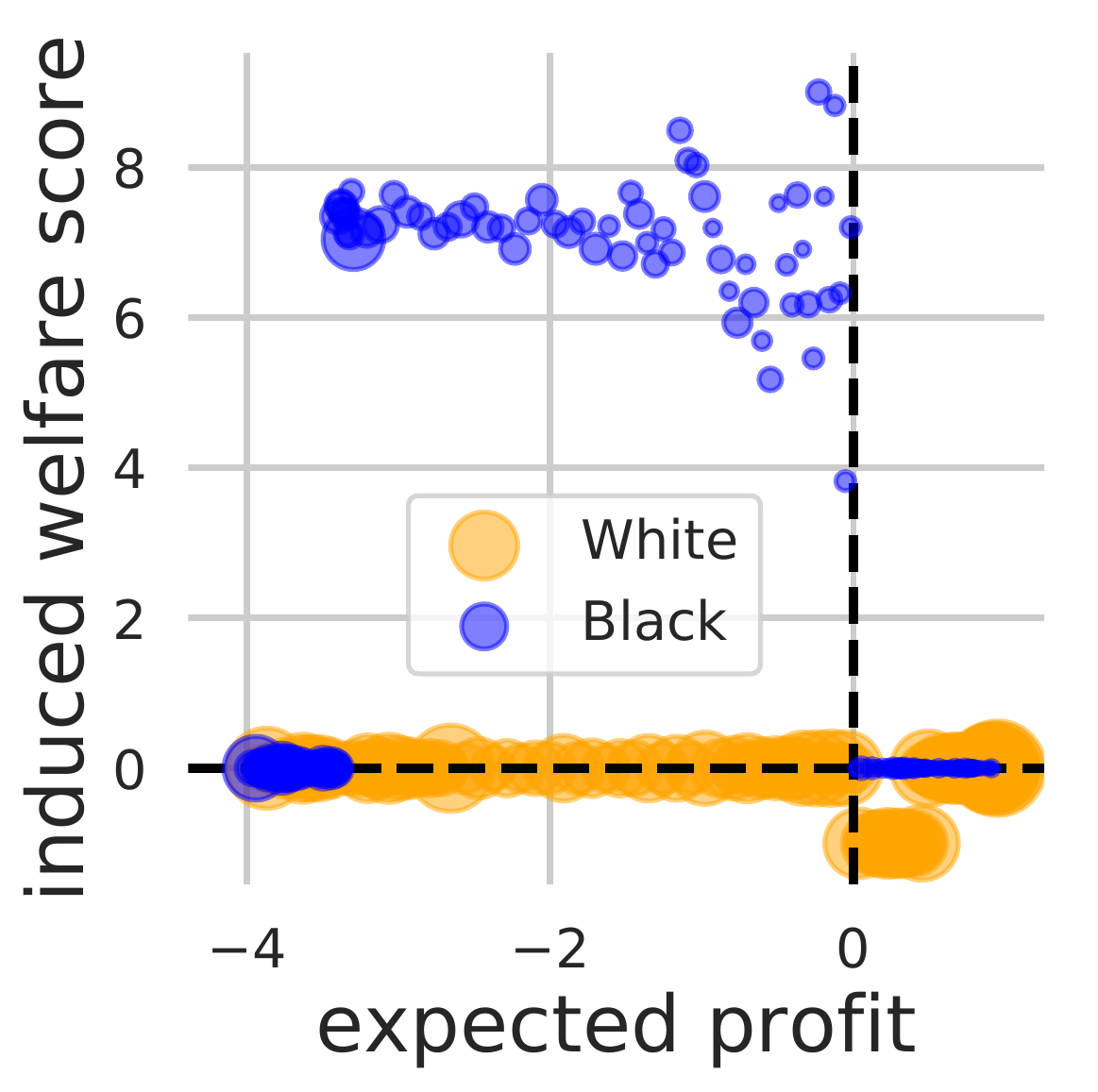}
        \caption{Distribution of profit and welfare scores. Marker size indicates population sizes.\label{fig:fairness_induced_weights}}
    \end{subfigure}%
    ~~
    \begin{subfigure}[t]{.54\linewidth}
        \centering
        \includegraphics[height=1.4in]{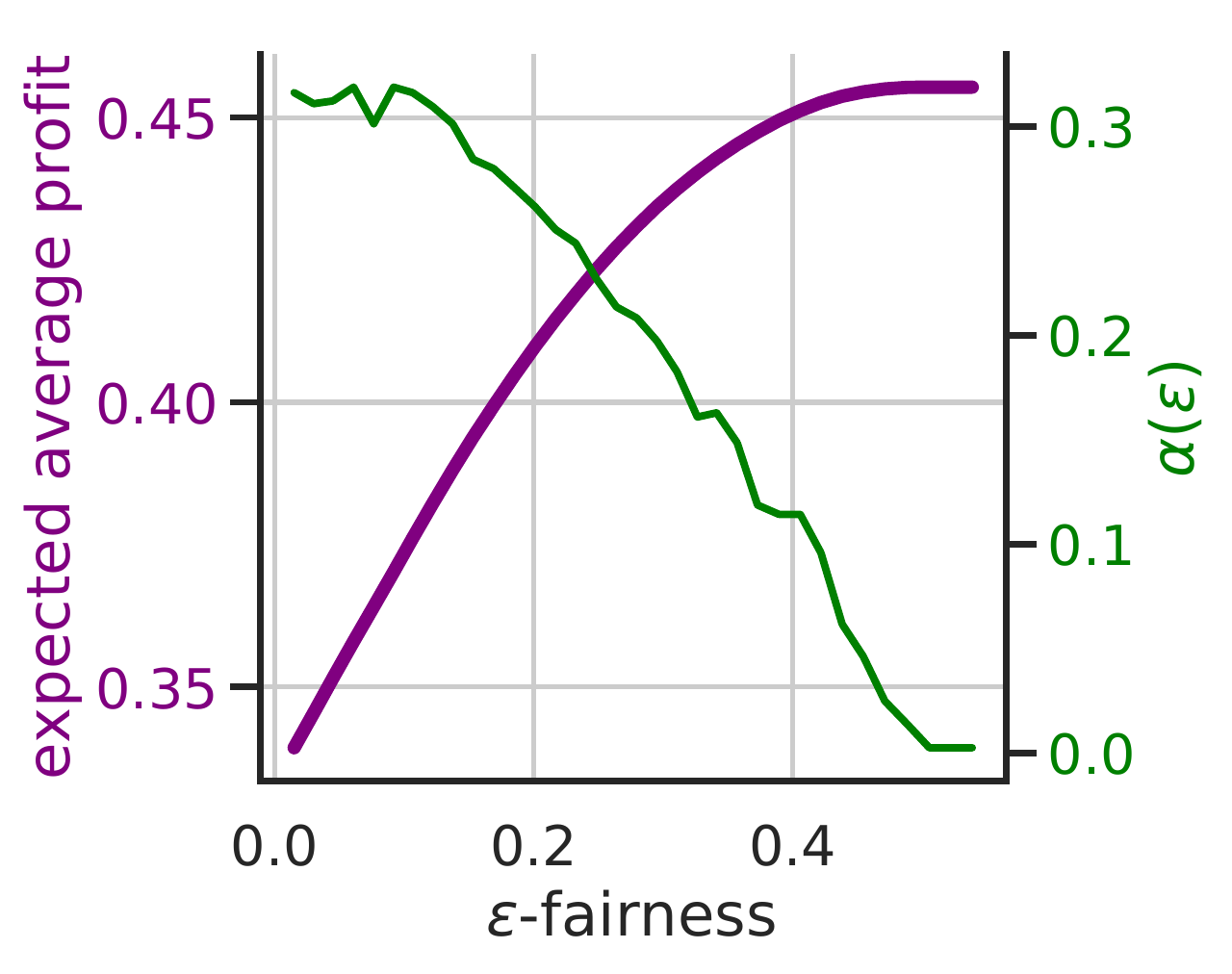}
        \caption{The fairness parameter $\epsilon$ determines the profit trade-off and corresponds to the welfare weight $\alpha$. \label{fig:fairness_alpha_epsilon}}
    \end{subfigure}
        \caption{Trade-offs between profit and fairness in lending can be equivalently encoded by a multi-objective framework. \label{fig:fairness}}
\end{figure}
}

%% file: conclusions.tex
We present a methodology for developing welfare-aware policies that jointly optimize a private return (such as profit) with a public objective (such as social welfare).
Taking care to consider data-limited regimes, we develop theory around the optimality of using learned predictors to make decisions.
Experiments corroborate our theoretical results, showing that thresholding on predicted scores can approach a Pareto-optimal policy.

This score-based approach to balancing competing objectives with noisy data is attractive for several reasons:
\begin{itemize}
\item{
Score-based policies can trade off multiple objectives with scalar predictions, with error bounded by a weighted sum of the errors in the learned scores.}
\item{The plug-in policy is a learned decision rule that is easily explained and diagnosed --- in line with the desire for transparent classification rules in practice.}
\item{It provides a crisp and interpretable connection to fair-constrained profit maximization, but reframes the problem as one of multi-objective optimization (see Sec.~\ref{sec:connections_to_fairness}).}
\end{itemize}

While separating the problem of instantiating learned policies from the problem of learning scores has desirable benefits, we note the limitations of this approach as well. First, the plug-in policy is not guaranteed to be the optimal policy learned from data. Thus, when further assumptions on the problem structure are appropriate, it may be worthwhile to consider more general policy classes learned from data. Second, the score-based approach shifts much of the difficulty of welfare-aware machine learning toward defining and predicting welfare, which is an area of active academic and policy debate~\citep{griffin1986well,kahneman2006developments}.

When welfare utilities are estimable, the ability to trade off context-sensitive measures with general policies can improve upon the status quo of applying machine learning policies in welfare-sensitive domains.  Further, a multi-objective framework could allow communities to understand the trade-offs between competing definitions of welfare or fairness in data constrained situations.

Taken together, these results help illustrate how machine learning can be used to design policies that prioritize the social impact of an algorithmic decision from the outset, rather than as an afterthought. 
By elucidating 
the possible trade-offs between competing objectives, and by illustrating the importance of measurement and prediction error in multi-objective optimization, we hope this work encourages new ways of thinking about welfare-aware machine learning.

%% file: acks.tex

This work was supported by NSF grant DGE1752814, the Bill and Melinda Gates Foundation, the Center for Effective Global Action, and DARPA and NIWC under contract N66001-15-C-4066. Any opinions,
findings, and conclusions or recommendations expressed in this material are those of the
authors and do not necessarily reflect the views of the National Science Foundation. The U.S. Government is authorized to reproduce and distribute reprints for Governmental purposes not withstanding any copyright notation thereon. The views, opinions, and/or findings expressed are those of the authors and should not be interpreted as representing the official views or policies of the Department of Defense or the U.S. Government.
MS is supported by the Open Philanthropy AI Fellowship. LTL is supported by the Open Philanthropy AI Fellowship and the Microsoft Ada Lovelace Fellowship. DB was partly supported by Microsoft Research.

%% file: technical_appendix.tex

\subsection{Pareto Policies Optimize Weighted Combination of Utilities}
\label{sec:pareto_char}
\begin{prop}[Pareto optimal policies optimize a composite objective]\label{prop:composite} Let $\polclass$ be a set of policies which is convex, and compact in a topology in which $\pol \mapsto \utilp(\pol)$ and $\utilw(\pol)$ are continuous. \footnote{The convexity of $\polclass$ means that $\polclass$ is closed under the randomized combination of policies. In the simplest case, compactness is achieved when the space of features is finite (e.g. features $x$ can only take a values in a discrete, finite subset of $\R^d$).} Then, a policy $\polst \in \polclass$ is Pareto optimal if and only if there exists an $\alpha \in [0,1]$ for which
\begin{align*}
\polst &\in \argmax_{\pol \in \polclass} \ \util_{\alpha}(\pol) \\
 \util_{\alpha}(\pol) &:= (1-\alpha)\utilp(\pol) + \alpha \utilw(\pol).
\end{align*}
\end{prop}
\begin{proof}
First, we prove that if $\polst \in \argmax_{\pol} \ \util_{\alpha}(\pol) := (1-\alpha)\utilp(\pol) + \alpha \utilw(\pol)$, then $\polst$ is Pareto optimal. Suppose that there exists an $\alpha$ for which $\polst \in \argmax_{\pol} \ \util_{\alpha}(\pol)$. If $\alpha \in \{0,1\}$, then $\polst$ maximizes either $\utilw(\cdot)$ or $\utilp(\cdot)$, and is therefore Pareto optimal by definition. Otherwise, if $\alpha \in (0,1)$, suppose for the sake of contradiction that $\polst$ is not Pareto optimal. Then there exists a policy $\pol$ for which $\utilw(\polst) \le \utilw(\pol)$ and $\utilp(\polst) \le \utilp(\pol)$, where one of these inequalities is strict. We can then check that $\util_{\alpha}(\polst) < \util_{\alpha}(\pol)$, contradiction that $\polst \in \argmax_{\pol} \ \util_{\alpha}(\pol)$. 

To show the other direction, suppose that $\polst$ is Pareto optimal. If $\polst$ maximizes either profit or welfare, then $\polst \in \argmax_{\pol} \ \util_{\alpha}(\pi)$ for either $\alpha = 1$ or $\alpha = 0$. Otherwise, let $W = \utilw(\polst)$. Then, by Pareto optimality,
\begin{align*}
\polst &\in \argmax \{\utilp(\pol) : \utilw(\pol) \ge \utilw(\polst), \pol \in \Pi \} \\
&= \argmax_{\pol \in \Pi} \left(\utilp(\pol) + \min_{t \ge 0} t(\utilw(\pol)- \utilw(\polst)\right)\\
&= \argmax_{\pol \in \Pi} \min_{t \ge 0}\left(\utilp(\pol) + t\left(\utilw(\pol)- \utilw(\polst)\right)\right).
\end{align*}
The map $\utilp(\pol)$ and $\utilw(\pol)$ are both linear functions in $\pol$. Hence, if $\polclass$ is a a convex, and compact in a topology in which $\pol \mapsto \utilp(\pol)$ and $\utilw(\pol)$ are continuous, Sion's minimax theorem~\citep{komiya1988elementary} ensures that strong duality holds, which means that we can switch order of the minimization over $t$ and maximization over $\pol$. Thus, for some $t \ge 0$,
\begin{align*}
\polst  &\in \argmax_{\pol} \left( \utilp(\pol) + t \cdot \utilw(\pol) - t \cdot\utilw(\polst) \right)~=~ \argmax_{\pol} \left( \utilp(\pol) + t \cdot \utilw(\pol) \right)\\
&=  \argmax_{\pol} \left(\frac{1}{1+t} \util(\pol) + \frac{t}{1+t}\utilw(\pol) \right) ~=~  \argmax_{\pol} \left(\util_{1/(1+t)}(\pol) \right),
\end{align*}
as needed. 
\end{proof}

\subsection{Optimal Policies under Exact Information\label{app:optimal_exact}}

Here we verify the optimality of threshold policies under exact information:
\begin{prop}[Pareto optimal policies on exact scores are threshold policies]\label{prop:pareto-thresholds}
	For the weighted utility $\util_{\alpha}$, the optimal policy $\polst_{\alpha}$ over the unrestricted class $\polclasst$ is a \emph{threshold} on a weighted combination of $\welfscore$ and $\profscore$, namely\footnote{When the distribution over $(\welfscore,\profscore)$ is sufficiently smooth, we can ignore the case where $\alpha \welfscore + (1-\alpha)\profscore = 0$.},
	\begin{align} 
	\polst_{\alpha}(\profscore,\welfscore) = \I(\alpha \welfscore + (1-\alpha) \profscore \ge 0).
	\end{align}
\end{prop}
\begin{proof}[Proof of~\Cref{prop:pareto-thresholds}]

Consider the reward of an arbitrary policy $\pi$. Recall that $\pi(x) \in [0,1]$ denotes the probability that $\pi$ classifies an individual with features $x$ as a $1$. Since $\pi(x) \cdot z \le z \cdot \I(z\ge 0)$ for $\pi(x) \in [0,1]$ and $z \in \R$, we can bound $\util_{\alpha}[\pi] = \Exp_{x}[ \pi(x) \cdot (\alpha \welfscore(x) + (1-\alpha)\profscore(x))] \le \Exp_x[\max\{0,\alpha \welfscore(x) + (1-\alpha)\profscore(x)\}]$. We can directly check that the threshold policy $\polst_{\alpha}(\profscore,\welfscore) = \I(\alpha \welfscore + (1-\alpha) \profscore \ge 0)$ saturates this bound.
\end{proof}

\subsection{Well-Behaved Pareto Curves~\label{app:Well_Behaved_Appendix}}
In this section, we establish that under mild regularity conditions, the Pareto frontier takes of the form of a continuous curve. The following assumption stipulates these conditions:
\begin{asm}[Well-Behaved Policy Class]\label{asm:well_behaved} Let $\polclass$ be a policy class, and let $\prof_{\max} := \sup_{\pi \in \polclass} \utilp(\pol)$, $\welf_{\max} := \sup_{\pol}\utilw(\pol)$, let $\prof_{\min} := \sup_{\pi \in \polclass} \{\utilp(\pol) : \utilw(\pol) = \welf_{\max}) $. A policy class $\polclass$ is said to be well-behaved if for any $\pol \in \polclass$:
\begin{enumerate}
\item[(a)]$\{p: (p,w) \in \Pareto \text{ for some } w \in \R\} =  [\prof_{\min},\prof_{\max}]$
\item[(b)] For any $p \in [\prof_{\min},\prof_{\max}]$, $\argmax_{\pol \in \polclass} \{\utilw(\pol): \utilp(\pol) = p\} $ is achieved. 
\end{enumerate}
\end{asm}
The following lemma shows that the above assumptions are reasonable, in that we shouldn't expect Pareto optimal policies with $p \notin [\prof_{\max},\prof_{\min}]$
\begin{lem}\label{lem:p_interval} Suppose that $\Pi$ is \emph{any} policy class such that there exists $\pi_{\welf}$ attaining $\utilw(\pi_{\welf}) = \welf_{\max}$. Then, for any  $(p,w) \in \Pareto(\Pi)$ which is a Pareto-optimal pair, $p \in [\prof_{\min},\prof_{\max}]$. 
\end{lem}
\begin{proof} Clearly $p \le \prof_{\max}$, since $\prof_{\max}$ is the maximal attainable profit. Now, suppose that $(p,w) \in \Pareto(\Pi)$ is a Pareto optimal pair, and assume for the sake of contradiction that $p < \prof_{\min}$. Since there exists a $\utilw(\pi_{\welf}) = \welf_{\max}$, there exists, for any $\epsilon > 0$, some policy $\pi_{\epsilon}$ such that $\utilw(\pi_{\epsilon}) = \utilw(\pi_{\welf})$, and $\utilp(\pi_{\epsilon}) \ge \prof_{\min} - \epsilon$. My making $\epsilon$ sufficiently small, we can ensure that $\utilp(\pi_{\epsilon}) > p$. On the other hand, $\utilw(\pi_{\epsilon}) = \utilw(\pi_{\welf}) = \max_{\pi \in \Pi} \util(\pi)$, so that in particular $\utilw(\pi_{\epsilon}) \ge w$. Hence, $\pi_{\epsilon}$ dominates the policy with utilities $(p,w)$ in a pareto sense, so that $(p,w) \notin \Pareto(\Pi)$.
\end{proof}

We now establish the existence of Pareto curves for for well-behaved policy classes:
\begin{lem}[Properties of the Pareto Curve]\label{lem:pareto_facts} 
For well-behaved function classes (Assumption~\ref{asm:well_behaved}), there exists a unique, increasing function $\parfunc_{\Pi}$ such that 
\begin{align}
\Pareto(\Pi) = \left\{(p,\parfunc_{\Pi}(p)): p \in [\prof_{\min},\prof_{\max}]\right\}\label{eq:pareto_function},
\end{align}
where $\parfunc_{\Pi}(p):= \sup_{\pol \in \polclass}\{\utilw(\pol): \utilp(\pol) = p\}$, and where we recall $\prof_{\min},\prof_{\min}$ from Assumption~\ref{asm:well_behaved}. If in addition $\Pi$ is \emph{convex}, then $\parfunc_{\Pi}(p)$ is concave. 
\end{lem}

\begin{proof}[Proof of Lemma~\ref{lem:pareto_facts}]
Suppose Assumption~\ref{asm:well_behaved} holds. 

First, we show $\Pareto(\Pi) \subseteq \left \{(p,\parfunc_{\Pi}(p)): \\ p \in [\prof_{\min},\prof_{\max}]\right\}$. Given $(p,w) \in \Pareto(\Pi)$ corresponding to a policy $\pol$, we must have that $p \in [\prof_{\min},\prof_{\max}]$  by \Cref{lem:p_interval}. Moreover, by Pareto optimality, $w = \parfunc_{\Pi}(p):= \sup_{\pol \in \polclass}\{\utilw(\pol): \utilp(\pol) = p\}$ since this optimal is attained by Assumption~\ref{asm:well_behaved}. 

For the reverse inclusion, we know that if $p \in [\prof_{\min},\prof_{\max}]$, then by Assumption~\ref{asm:well_behaved}(a), there exists a policy $\pol$ such that $\utilp(\pol) = p$. Then, by Assumption~\ref{asm:well_behaved}(b), there exists a policy $\pol$ which maximizes $\utilw(\pol): \utilp(\pol) = p$. By definition, $\utilw(\pol) = \parfunc_{\Pi}(p)$, and $\pol$ is Pareto optimal by definition. Hence $(p,w) \in \Pareto(\Pi)$. 

We now show that that convexity of $\polclass$ implies concavity of $\parfunc_{\polclass}$. It suffices to show that, for any  points $(p_1,w_1),\,(p_2,w_2) \in \Pareto(\Pi)$, and any $\lambda \in [0,1]$, $\lambda w_1 + (1-\lambda)w_2 \le \parfunc_{\polclass}(\lambda p_1 + (1-\lambda)p_2)$. Indeed, by definition of the Pareto curve, there exist policies $\pol_1$ and $\pol_2$ such that $(\utilp(\pol_i),\utilw(\pol_i)) = (p_i,w_i)$ for $i \in \{1,2\}$. By convexity of $\Pi$, the policy $\pol := \lambda \pol_1 + (1-\lambda) \pol_2 \in \Pi$. Moreover, $\utilp(\pol) = \lambda p_1 + (1-\lambda)p_2$ and $\utilw(\pol) = \lambda w_1 + (1-\lambda)w_2$. Finally, by definition of $\parfunc_{\polclass}$, $\utilw(\pol) \le \parfunc_{\polclass}(\utilp(\pol))$, which concludes the proof.
\end{proof}

\Cref{lem:pareto_facts} confirms that the Pareto curve, as we might intuitively imagine it, actually exists.
With this in hand,
we now show that given the Pareto-optimal policies with \emph{exact} scores, the parameterizing function $\parfunc_{\Pi}$ is concave. Pictorially, $\parfunc_{\Pi}(p)$ is the Pareto frontier interpreted as a function of allowable profit $p$ which returns the maximum amount of welfare $w$ that can be achieved at this profit level (e.g. the black curve in Figure~\ref{fig:pareto_sim_first}, left).
\begin{thm}[Pareto Frontier under exact knowledge]\label{thm:exact_pareto} Suppose that the unconstrained policies are a well-behaved class. Consider the setting where the welfare and profit are specified exactly by scores $w$ and $p$. Then, given any population distribution over $\profscore, \welfscore$, the Pareto optimal policies $\polst_{\alpha}$ are given by Eq.~\eqref{eq:alpha_pareto_score_pol}  
and the Pareto frontier $\Pareto(\polclasst)$ is given by
	\begin{align*}
	 \Pareto_{\exact} := \{(\util_{\prof}(\polst_{\alpha}), \util_\welf(\polst_\alpha)): \alpha \in [0,1]\}
	 \end{align*}
	Moreover, the associated function $\parfunc_{\exact}(p)$ is non-increasing and concave. 
\end{thm}
\begin{proof}[Proof of \Cref{thm:exact_pareto}]
The first statement follows from Proposition \ref{prop:pareto-thresholds}, and that $\parfunc_{ \Pareto_{\mathrm{exact}}}(p)$ is non-decreasing follows from  Lemma~\ref{lem:pareto_facts}. 
	
Convexity of $\parfunc_{ \Pareto_{\mathrm{exact}}}(p)$ follows from Lemma~\ref{lem:pareto_facts}, and the fact that $\polclasst$ is convex. 
\end{proof}

\subsection{Proof of Theorem~\ref{thm:pareto_inexact}}\label{sec:prop_suff_stat} 
For the theorem, we assume that both policies based on empirical scores and those based on exact scores are \emph{well-behaved} in the sense of \Cref{asm:well_behaved}.

We first establish part (a), namely that 
\begin{align*}
\piopt_{\alpha} \in\argmax_{\pol \in \Piemp} \Exp \left[\util_{\alpha}(\pol(\fwhat, \fphat))\right].
\end{align*} 
Where recall that $
\piopt_{\alpha} := (1-\alpha) \cdot \mubar_{\prof}+\alpha \cdot \mubar_{\welf}
$. We have that
\begin{align*}
\Util_{\alpha}(\pi) &= \Exp[\left((1-\alpha)\profscore + \alpha \welfscore)\right)\pi(\fphat,\fwhat)]\\
&= \Exp[\left((1-\alpha)\Exp[\profscore \mid \fphat, \fwhat] + \alpha \Exp[\welfscore \mid \fphat,\fwhat]\right) \cdot \pi(\fphat,\fwhat)]\\
&:= \Exp[\left((1-\alpha)\mubar_{\prof}(\fphat,\fwhat) + \alpha \mubar_{\welf}( \fphat,\fwhat) \right)\cdot \pi(\fphat,\fwhat)]\\
&\le \Exp\left[\max\left\{(1-\alpha)\mubar_{\prof}((  \fphat,\fwhat) + \alpha \mubar_{\welf}(  \fphat,\fwhat) , 0 \right\}\right]\,~= \util_{\alpha}(\piopt_{\alpha}).
\end{align*}
Hence, we obtain the Pareto optimality of $\piopt_{\alpha}$ by Proposition~\ref{prop:composite}.

Part (b) is a direct consequence of \Cref{lem:pareto_facts}  and the assumption that our policy class is well behaved.

For part (c), empirical policies are dominated by those induced by the true score functions because, as established, the Pareto optimal policies based on the true score functions are in fact Pareto optimal over all policies that are induced by a function of the features $x$.
\qed

\subsection{Utilitity Loss Induced by Score Function Suboptimality\label{sec:subopt}}

\begin{proof}[Proof of Proposition~\ref{prop:suboptimal}]
We compute 
	\begin{align*}
	\util_\alpha(\polhat_{\alpha}) - \util_\alpha(\pol_{\alpha}) = \Exp[ \left((1-\alpha) \profscore +\alpha  \welfscore\right) \left(\polhat_{\alpha} - \pol_{\alpha}\right)]~.
	\end{align*}
	Define the functions $Y(x) =  (1-\alpha)\fp(x)+\alpha \fw(x) $, and let $E(x) = (1-\alpha)(\fphat(x) - \fp(x))+\alpha(\fwhat(x) - \fw(x))$. Then, $\polhat_{\alpha}(x) - \pol_{\alpha} = \I(Y(x)+E(x) \ge 0) -  \I(Y(x) \ge 0)$. We see that this difference is at most $1$ in magnitude, and is $0$ unless possibly if $|Y(x)| \le |E(x)|$. Hence,
	\begin{align*}
	|Y(x)| \cdot |\polhat_{\alpha}(x) - \pol_{\alpha}(x)| \le |E(x)|~.
	\end{align*}
	Therefore
	\begin{align*}
	|\util_\alpha(\polhat_{\alpha}) - \util_\alpha(\pol_{\alpha})| &= |\Exp[Y(x)(\polhat_{\alpha}(x) - \pol_{\alpha}(x)]|\\
	&\le \Exp [|Y(x)|\cdot |\polhat_{\alpha}(x) - \pol_{\alpha}(x)|]\\
	&\le \Exp [|E(x)|] = \Exp[|(1-\alpha)(\fphat(x) - \fp(x))+\alpha(\fwhat(x) - \fw(x))|]\\
	&\le (1-\alpha) \Exp[|\fphat(x) - \fp(x)|] +\alpha\Exp[|\fwhat(x) - \fw(x))|].
	\end{align*}
	\end{proof}
We remark that in general, optimizing arbitrary loss functions for function value states (e.g. estimating $\alpha$-utilities for all $\alpha$ directly from features) requires a prohibitively large sample \citep{balkanski2017sample}. The structures of the combined learning problems and $\alpha$-utility in our setting allow us to circumvent this lower bound. 

\begin{proof}[Proof of~\Cref{claim:gaussian_score_bound}]
By definition of $\polst_{\alpha}$,
\begin{align}
\Util_\alpha(\polst_{\alpha}) - \Util_\alpha(\polhat_{\alpha}) = \Exp[|\alpha w + (1-\alpha) p| \cdot \I(\polst_{\alpha} \neq \polhat_{\alpha})]
\end{align}
Now consider the event $\polst_{\alpha} \neq \polhat_{\alpha}$. This happens only when the predicted scores incur an opposite classification by the $\alpha$ threshold policy, that is, $(\alpha w_i + (1-\alpha) p_i) \cdot (\alpha \hat{w}_i + (1-\alpha) \hat{p}_i) < 0$. Define the quantities 
\begin{align*}
y_i &:= \alpha w_i + (1-\alpha) p_i \\
z_i &:= \alpha (\hat{w}_i - w_i) + (1-\alpha)(\hat{p}_i - p_i)~,
\end{align*}
so that 
\begin{align*}
\Util_\alpha(\polst_{\alpha}) - \Util_\alpha(\polhat_{\alpha}) = \Exp[|y| \cdot \I(y (y + z ) < 0)]~,
\end{align*}
where  $y \sim \calN(0, \alpha^2\sigma^2_w + (1-\alpha)^2 \sigma^2_p + 2 \rho \alpha (1-\alpha) \sigma_w \sigma_p)$ and $z$ is sub-Gaussian with squared parameter $\tilde{\sigma}^2 = 4(\alpha^2 \sigma^2_{\varepsilon_{w}} + (1-\alpha)^2 \sigma^2_{\varepsilon_{p}})$~\citep{wainwright2019high}.\footnote{When $\epsilon_w$ and $\epsilon_p$ are assumed to be independent, $\tilde{\sigma}^2 = (\alpha^2 \sigma^2_{\varepsilon_{w}} + (1-\alpha)^2 \sigma^2_{\varepsilon_{p}})$ \citep{wainwright2019high}.}
By assumption, the errors are independent of the scores, so that
\begin{align*}
\Exp[|y| \cdot \I(y (y + z ) < 0)] & = 
\Exp\left[|y| \cdot \Exp\left[ \I(y (y + z ) < 0) | y\right] \right]\\
& = \Exp\left[|y| \cdot \P\left( y (y + z ) < 0 \right)\right] ~.
\end{align*}
Now by sub-Gaussianity of $z$, we bound $\P\left( y (y + z ) < 0 \right)$ for any fixed $y$: 
\begin{align*}
\P\left( y (y + z ) < 0 \right)
&\leq \begin{cases}\P\left( z < -y \right) & y > 0\\
\P\left( z > -y \right)  & y < 0
\end{cases}\\
&\leq e^{- \frac{y^2}{2 \tilde{\sigma}^2}}~,
\end{align*}
so that by symmetry of the distribution of $y$, the expectation can be bounded as
\begin{align*}
\Exp[|y| \cdot \I(y (y + z ) < 0)]
&\leq 2 \int_{0}^{\infty} y ( e^{- {y^2}/({2 \tilde{\sigma}^2})}) \cdot \frac{1}{\sigma_y \sqrt{2\pi}} e^{-y^2 / (2\sigma^2_y)} dy \\
&= \frac{ \sqrt{2}}{\sigma_y \sqrt{\pi}} \int_{0}^{\infty} y e^{- \frac{y^2}{2} \left(\frac{1}{\tilde{\sigma}^2} + \frac{1}{\sigma_y^2}\right)} dy \\
& = \frac{1 }{\sigma_y \left(\frac{1}{\tilde{\sigma}^2} + \frac{1}{\sigma_y^2}\right)^{1/2}}  \cdot \frac{\sqrt{2} \left(\frac{1}{\tilde{\sigma}^2} + \frac{1}{\sigma_y^2}\right)^{1/2}}{\sqrt{\pi}}\int_{0}^{\infty} y e^{- \frac{y^2}{2} \left(\frac{1}{\tilde{\sigma}^2} + \frac{1}{\sigma_y^2}\right)} dy ~.
\end{align*}
This is a scaled mean of a half-normal distribution with scale parameter $\left(\frac{1}{\tilde{\sigma}^2} + \frac{1}{\sigma_y^2}\right)^{-1/2}$. Thus, difference in $\alpha$-utility is bounded as
\begin{align*}
\Util_\alpha(\polst_{\alpha}) - \Util_\alpha(\polhat_{\alpha}) 
&\leq \frac{\sqrt{2}}{\sigma_y\sqrt{\pi} } \left(\frac{1}{\tilde{\sigma}^2} + \frac{1}{\sigma_y^2}\right)^{-1} \\
& = \frac{\sigma_y \cdot \sqrt{2}}{\sqrt{\pi}}  \left( \frac{\tilde{\sigma}^2}{\tilde{\sigma}^2 + \sigma_y^2}\right) \\
\end{align*}

Now, note that the expected utility of the optimal classifier is $\Exp[y \I\{\ y \geq 0\}  \geq 0]$ for $y \sim \calN(0,\sigma_y^2)$ as defined above. Then as one half the expectation of a half-normal distribution with scale parameter $\sigma_y$, 
\begin{align*}
\Exp[U_\alpha(\polst)] 
= \frac{\sigma_y}{\sqrt{2\pi}}
= \frac{1}{\sqrt{2\pi}} \sqrt{\alpha^2 \sigma^2_w + (1-\alpha)^2 \sigma^2_p + 2\rho \alpha(1-\alpha) \sigma_w\sigma_p }~,
\end{align*} 
which is nondecreasing in $\rho$ for $\alpha \in [0,1]$, and increasing in $\rho$ for $\alpha \in (0,1)$.
Then, the expected Pareto utility of the plug in policy can be lower bounded as
\begin{align*}
\Exp[\Util_\alpha(\polhat_{\alpha})] &=  \Exp[\Util_\alpha(\polst_{\alpha}) -\left(\Util_\alpha(\polst_{\alpha}) - \Util_\alpha(\polhat_{\alpha})\right)]\\
&\geq  \frac{\sigma_y }{\sqrt{2\pi}} \left( 1 - c \frac{\tilde{\sigma}^2}{\tilde{\sigma}^2 + \sigma_y}\right) \\
&= \Exp[\Util_\alpha(\polst_{\alpha})] \left( 1 - c \frac{\tilde{\sigma}^2}{\tilde{\sigma}^2 + \sigma^2_y}\right)
\end{align*}
Where $c=2$. Since $\sigma_y$ is increasing in $\rho$ for $\alpha \in (0,1)$, the lower bound on $\Exp[\Util_\alpha(\polhat_{\alpha})]$ is increasing as well. 
\end{proof}

%% file: experiment_details.tex

Code for simulation and real data experiments is available at \url{https://github.com/estherrolf/multi-objective-impact}. All experiments were run on a personal laptop running unix with 16GB memory and a 2.5GhZ Intel i7 processor.  

\subsection{Abalone}
\label{sec:abalone_details}

The features included in $x$ for each model are: sex (female/male/infant), length, diameter, height, and whole weight. There are in total $4177$ data points.\footnote{Data is available for download at: \url{https://archive.ics.uci.edu/ml/datasets/Abalone}~.}  We instantiate scores as:
\begin{align*}
\profscore &:= \texttt{meat\_price\_per\_gram} \cdot (200 \cdot \texttt{shucked\_weight})
+ \texttt{shell\_price\_by\_cm}^2 \cdot (20 \cdot \texttt{length}) \cdot (20 \cdot \texttt{diameter}) \\
\welfscore &: = c \cdot \log((\texttt{rings}+1.5)/ 10)
\end{align*}
where $\texttt{meat\_price\_per\_gram} = 0.25$ 
and $\texttt{shell\_price\_per\_cm}^2 = 0.32$, and the constant factors of 20 and 200 match units of the original data with units of these prices.  We add $1.5$ to the ring count to get age, and divide by $10$ before taking the logarithm to encode that harvesting abalone less than 10 years of age has negative welfare. We scale the welfare weights by constant $c$ so that the distribution of welfare and profit have the same standard deviation.

\begin{figure}
\centering
\includegraphics[width=.7\textwidth]{\figpath 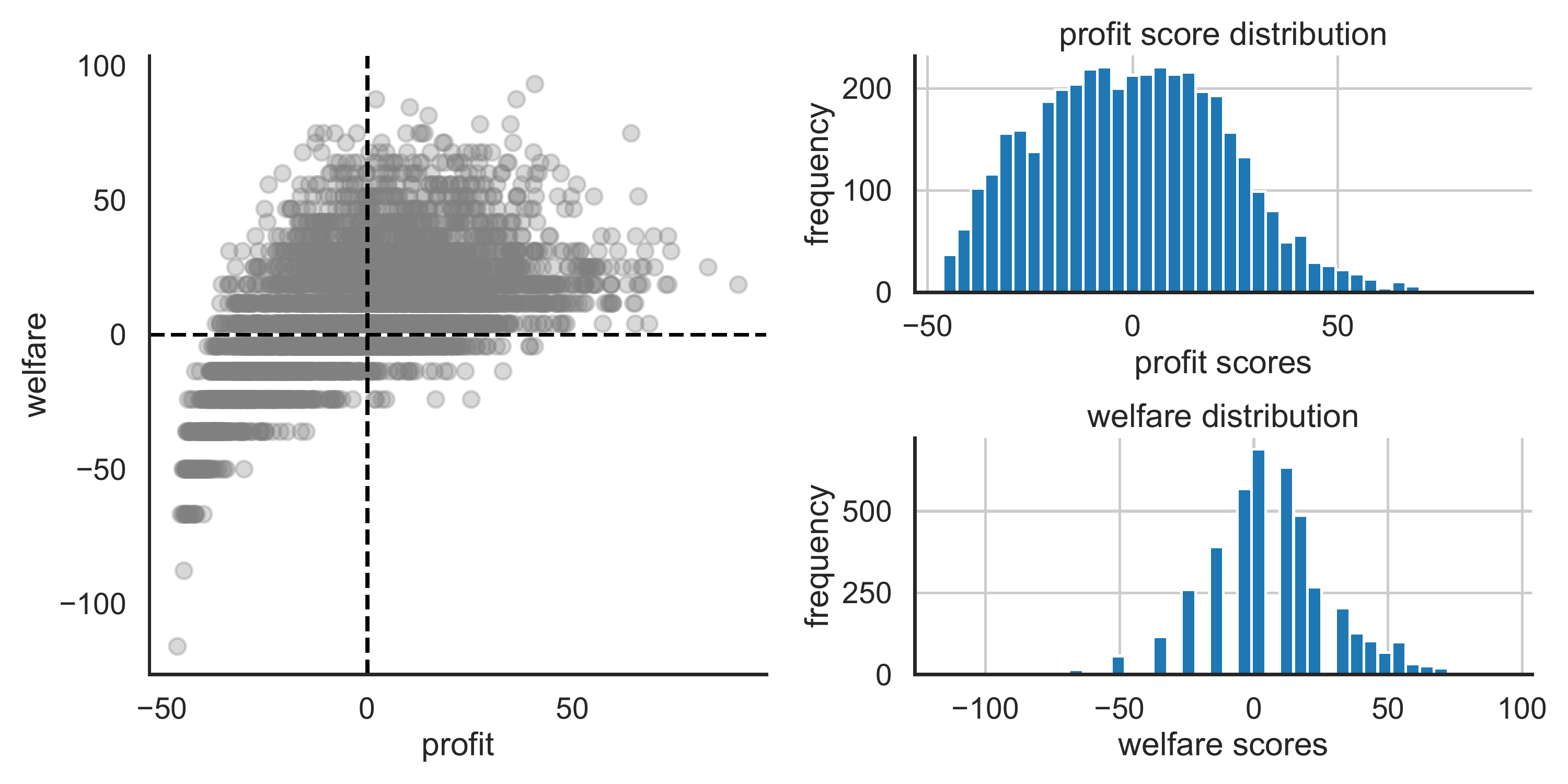}
\caption{Summary of abalone score distributions. \label{fig:abalone_data_summary}}
\end{figure}

Figure~\ref{fig:abalone_data_summary} shows the distributions of the scores. The average of the welfare scores is $9.13$, the average of the profit scores is $0.00$, and the correlation of welfare and profit scores is $0.56$.

\begin{figure}
\centering
\begin{subfigure}[h]{.3\textwidth}
\includegraphics[width=.9\textwidth]{\figpath 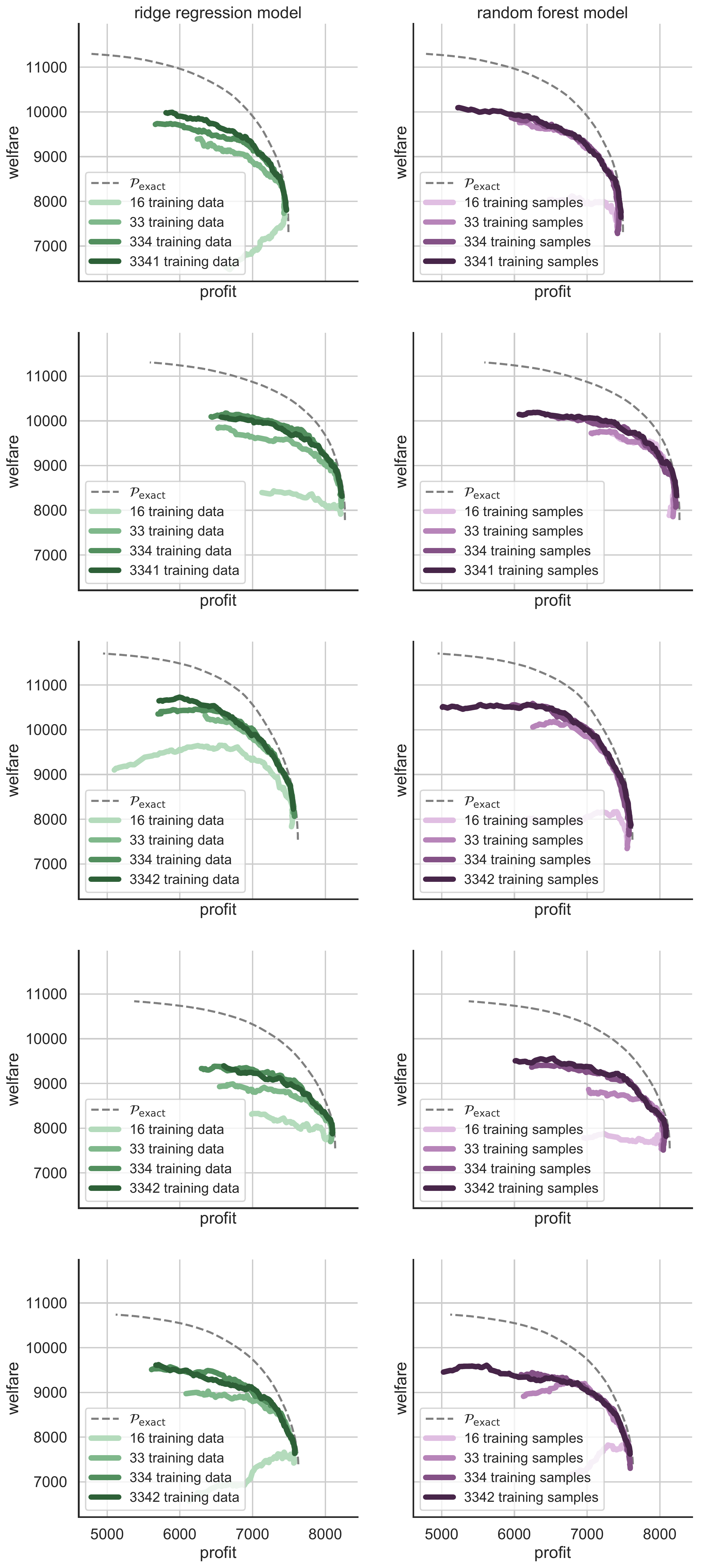}
\caption{Per fold predictions (fold 0 is shown in Fig.~\ref{fig:abalone_n} in main text).}
 \end{subfigure}
 ~
\begin{subfigure}[h]{.3\textwidth}
\includegraphics[width=.9\textwidth]{\figpath 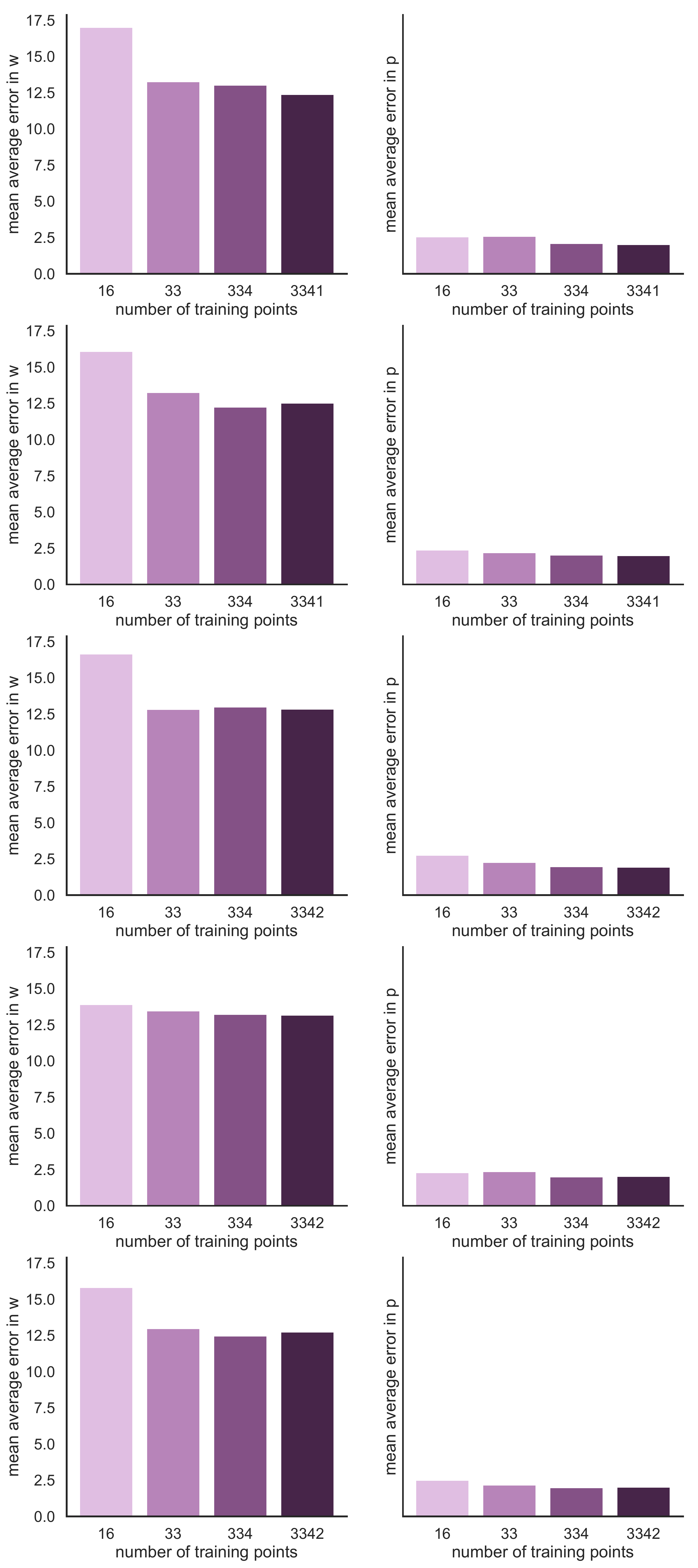}
\caption{Errors in scores from ridge regression model.}
\end{subfigure}
~
\begin{subfigure}[h]{.3\textwidth}
\includegraphics[width=.9\textwidth]{\figpath 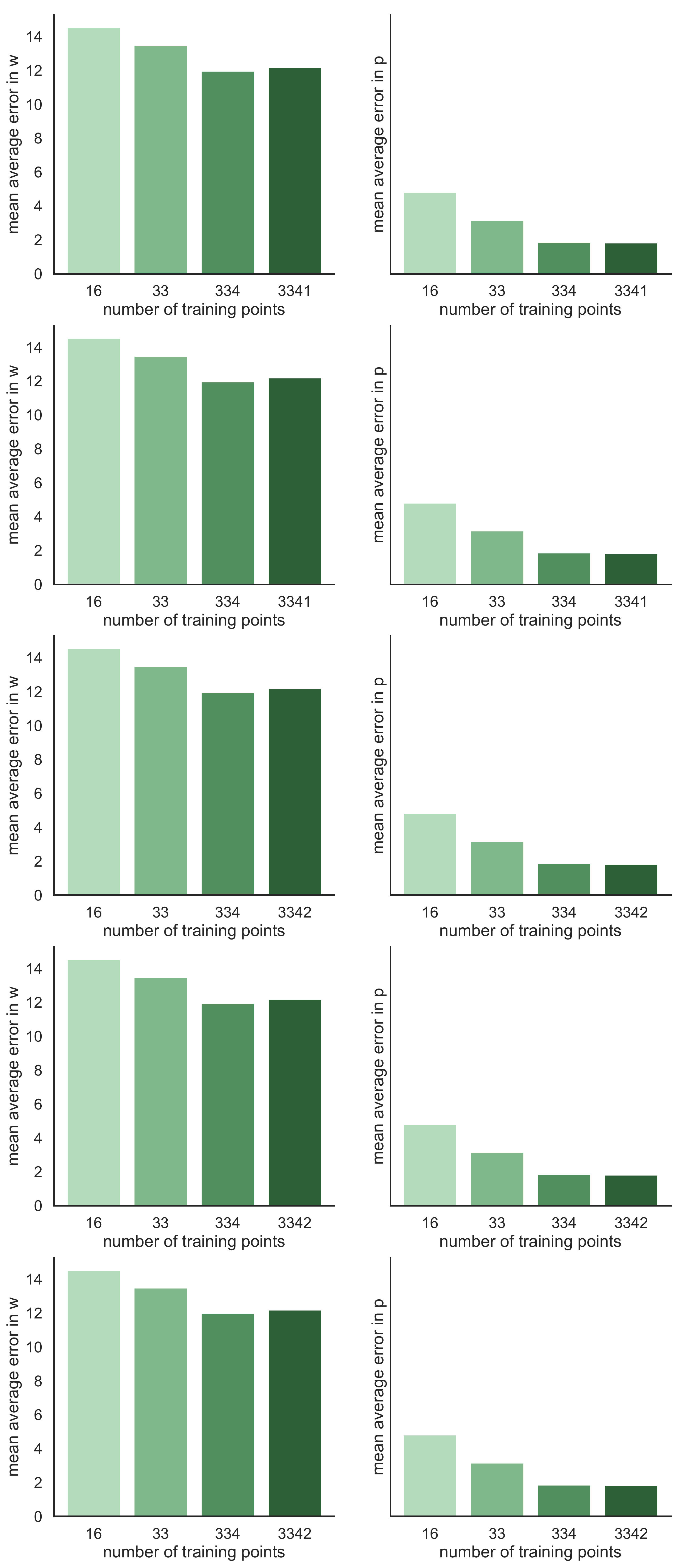}
\caption{Errors in scores from random forest model.}
\end{subfigure}
\caption{Performance with different training set sizes. Accompanies Fig.~\ref{fig:abalone_n} in main text. \label{fig:abalone_all_n}}
\end{figure}

\begin{figure}
\centering
\begin{subfigure}[h]{.3\textwidth}
\includegraphics[width=.9\textwidth]{\figpath 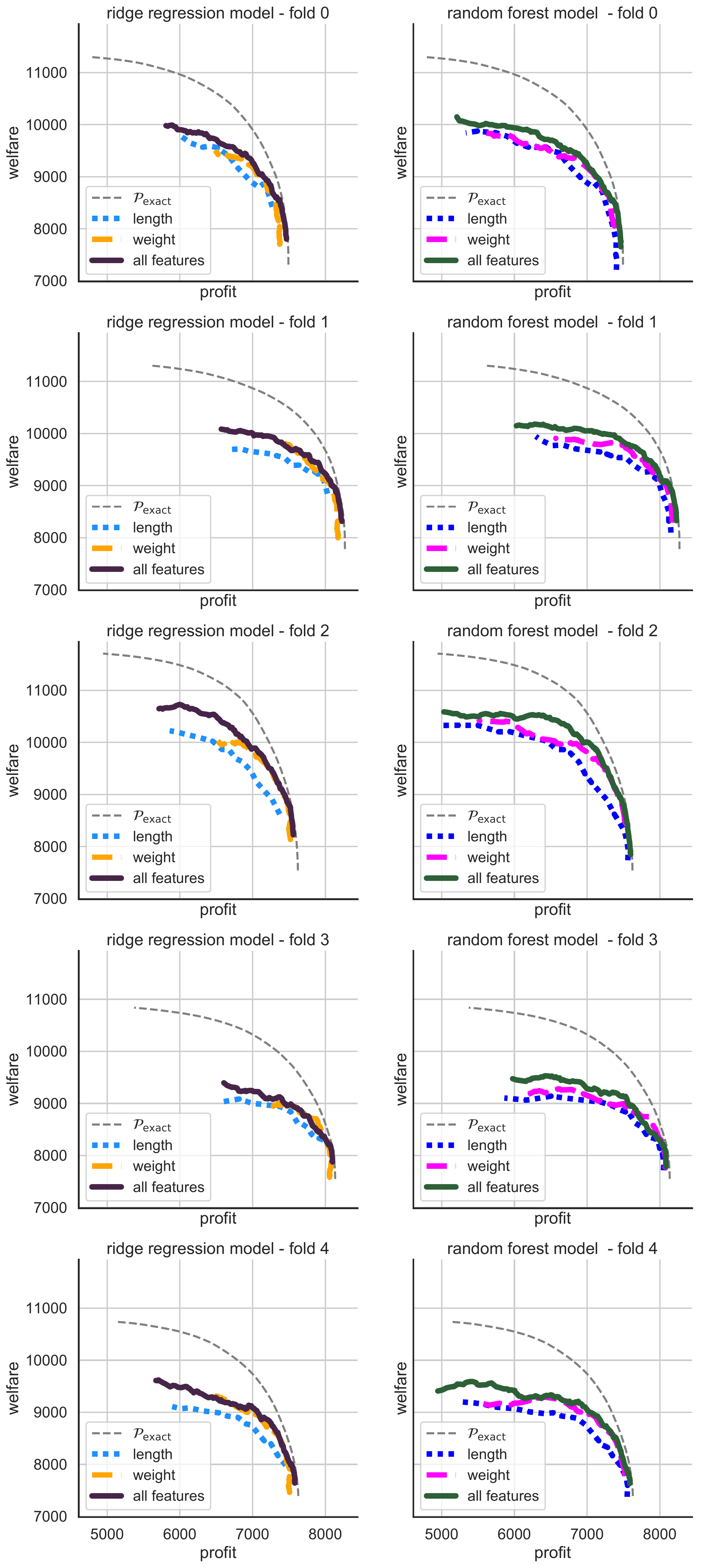}
\caption{Per fold predictions (fold 0 is shown in Fig.~\ref{fig:abalone_d} in main text).}
 \end{subfigure}
 ~
\begin{subfigure}[h]{.3\textwidth}
\includegraphics[width=.9\textwidth]{\figpath 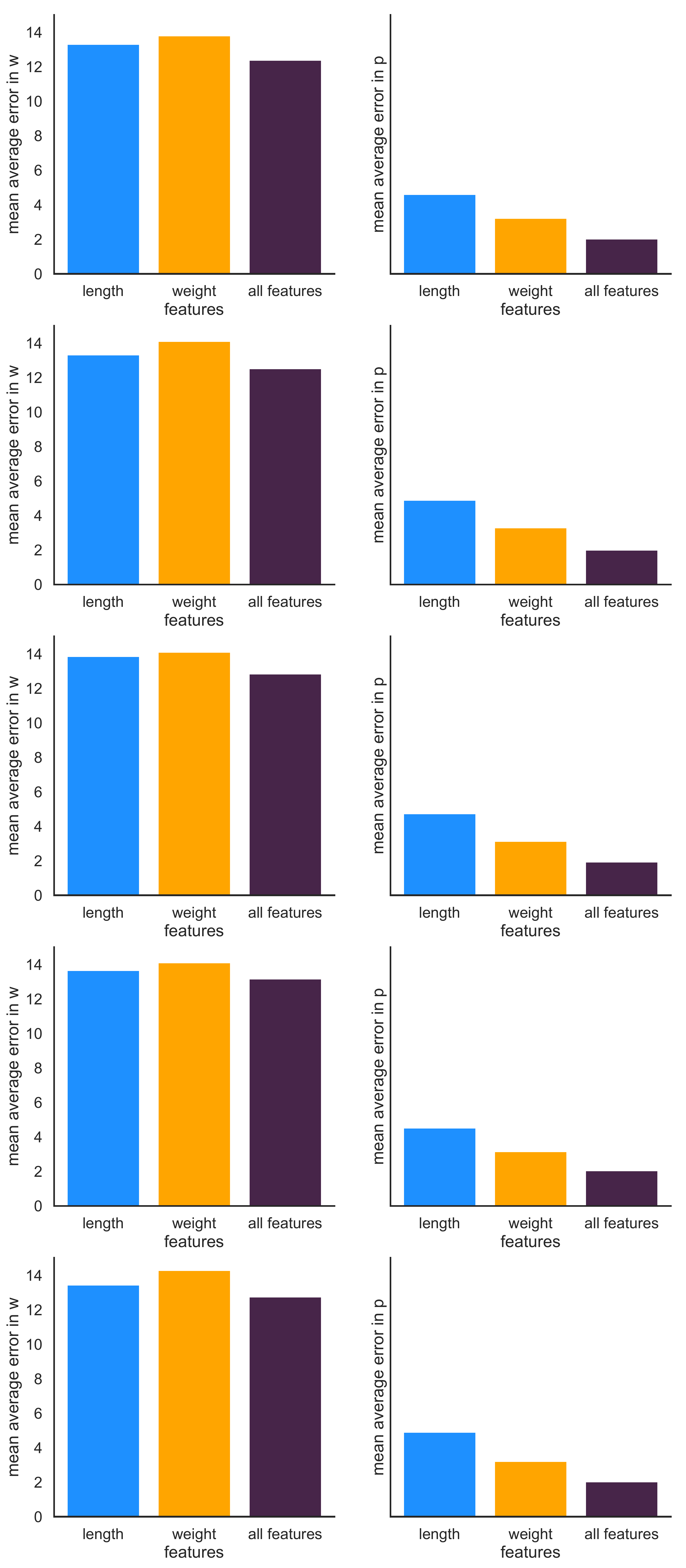}
\caption{Errors in scores from ridge regression model.}
\end{subfigure}
~
\begin{subfigure}[h]{.3\textwidth}
\includegraphics[width=.9\textwidth]{\figpath 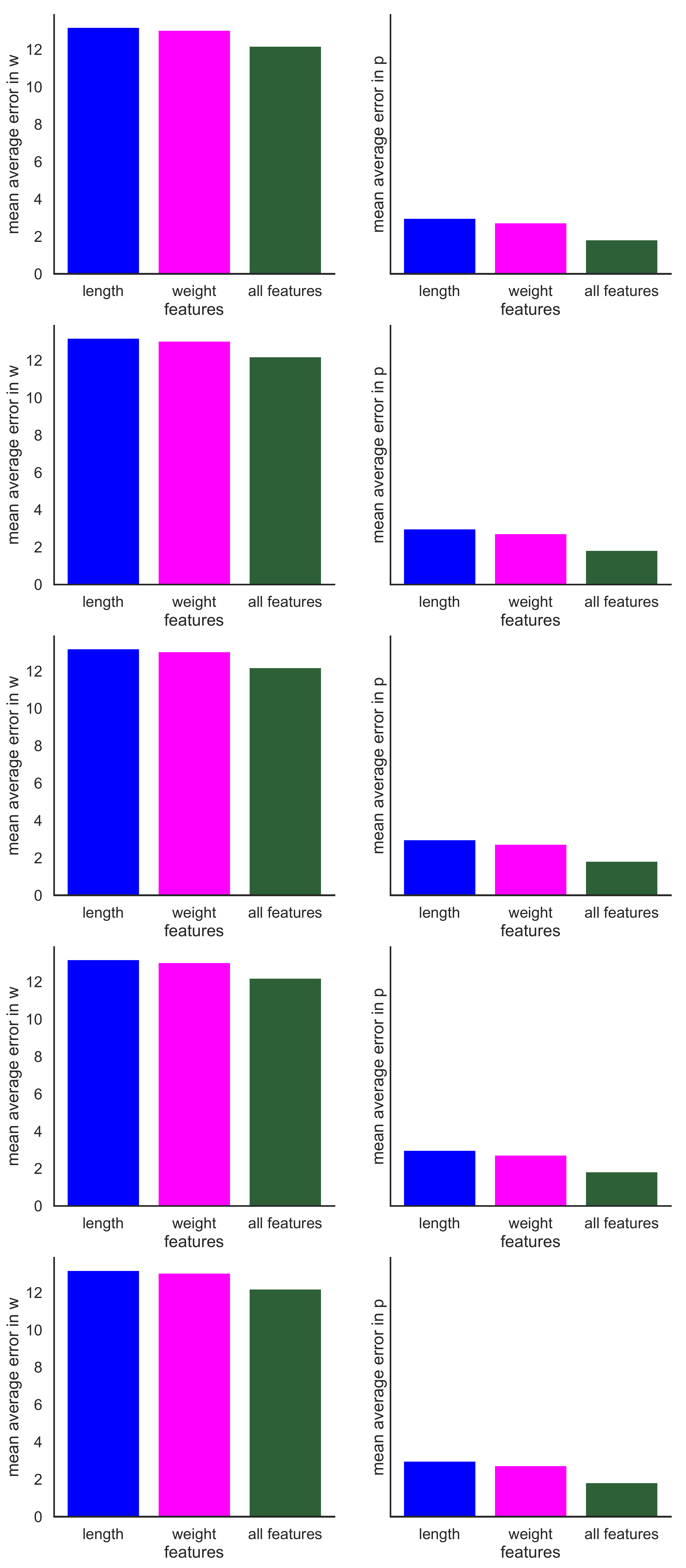}
\caption{Errors in scores from random forest model.}
\end{subfigure}
\caption{Performance with different features sets. Accompanies Fig.~\ref{fig:abalone_d} in main text.\label{fig:abalone_all_d}}
\end{figure}

Figures~\ref{fig:abalone_all_n} and~\ref{fig:abalone_all_d} show the performance of the scores and plug-in policies for different number of training set sizes and different feature sets, to augment the results shown in Fig.~\ref{fig:abalone_n} and Fig.~\ref{fig:abalone_d} of the main text. Main text figures show performance on the first fold of five randomly chosen cross-validation folds, in these figures we show all five folds along with the mean average errors of the predictors.

Within each of the five evaluation folds, we train the score models through cross-validation with 4 folds within the training set. Since we leave $20\%$ of data for evaluation in each outer fold, each model in each row of Figure~\ref{fig:abalone_all_d} is trained via 4-fold cross-validation on $80\%$ of the data. 
For figure~\ref{fig:abalone_all_d}, we subset from the 5 training sets of $80\%$ of the total data, and run 4-fold cross-validation on these subsets to select model hyperparameters. 
The hyperparameters we consider are $\lambda = \texttt{np.logspace(-8,2,base=10,num=11)}$ for ridge regression and $\texttt{num\_estimators} = [200,400]$ and $\texttt{depths} = [4,8]$ for the random forest model. 
We choose hyperparameters to minimize the mean average error over the $4$ folds. 
 For ridge regression, we use the implementation from $\texttt{sklearn.linear\_model.Ridge()}$ and for random forest, $\texttt{sklearn.ensemble.RandomForestRegressor()}$.

The hyperparameters chosen for generating Fig.~\ref{fig:abalone_n} in the main text are presented in \Cref{table:hps_abalone_n}. The hyperparameters chosen for generating Fig.~\ref{fig:abalone_d} in the main text are presented in \Cref{table:hps_abalone_d}.

\begin{table}[h]
  \caption{Hyperparameter configurations to generate Fig.~\ref{fig:abalone_n}.}
   \label{table:hps_abalone_n}
  \centering
  \begin{tabular}{lllllll}
    \toprule
 \multicolumn{1}{c}{}  & \multicolumn{2}{c}{ridge regression} & \multicolumn{4}{c}{random forest}\\
    \cmidrule(r){2-3} \cmidrule(r){4-7}
  \multicolumn{1}{c}{}   & \multicolumn{2}{c}{regularization $\lambda$  } & \multicolumn{2}{c}{num. estimators}     & \multicolumn{2}{c}{maximum depth} \\
   \cmidrule(r){2-3}  \cmidrule(r){4-5} \cmidrule(r){6-7}
   \multicolumn{1}{l}{\# training points}   & \multicolumn{1}{l}{w }  & \multicolumn{1}{l}{p}   & \multicolumn{1}{l}{w }  & \multicolumn{1}{l}{p}   & \multicolumn{1}{l}{w }  & \multicolumn{1}{l}{p}   \\
    \cmidrule(r){0-0} \cmidrule(r){2-7}  
    16   	&  1e1  &  1e-3   &  400 & 400 & 8 & 8  \\
    33      & 1e-3  &  1e-8   &  200 & 200 & 8 & 8  \\
    334     & 1e-3  &  1e-3   &  200 & 400 & 4 & 8 \\
    3341    & 1e-1  &  1e-1   &  200 & 400 & 4 & 8  \\
    \bottomrule
  \end{tabular}
\end{table}

\begin{table}[h]
  \caption{Hyperparameter configurations to generate Fig.~\ref{fig:abalone_d}.}
   \label{table:hps_abalone_d}
  \centering
  \begin{tabular}{lllllll}
    \toprule
 \multicolumn{1}{c}{}  & \multicolumn{2}{c}{ridge regression} & \multicolumn{4}{c}{random forest}\\
    \cmidrule(r){2-3} \cmidrule(r){4-7}
  \multicolumn{1}{c}{}   & \multicolumn{2}{c}{regularization $\lambda$  } & \multicolumn{2}{c}{num. estimators}     & \multicolumn{2}{c}{maximum depth} \\
   \cmidrule(r){2-3}  \cmidrule(r){4-5} \cmidrule(r){6-7}
   \multicolumn{1}{l}{\# features}   & \multicolumn{1}{l}{w }  & \multicolumn{1}{l}{p}   & \multicolumn{1}{l}{w }  & \multicolumn{1}{l}{p}   & \multicolumn{1}{l}{w }  & \multicolumn{1}{l}{p}   \\
    \cmidrule(r){0-0} \cmidrule(r){2-7}  
    length      & 1e-8  &  1e-2   &  200 & 200 & 4 & 8 \\
    weight   	&  1e0  &  1e-2   &  400 & 200 & 4 & 4  \\
    all     &  1e-1  &  1e-1   &  400 & 400 & 4 & 8  \\
    \bottomrule
  \end{tabular}
\end{table}

\subsection{YouTube}
\label{sec:youtube_details}
Data for the YouTube experiment is used with permission of the authors of~\citet{faddoul_longitudinal_2019} (in progress; data to be released publicly). 

Defining an allowable quality threshold as the median score of all videos ($=0.95$), we instantiate $\fwhat = (1 - s_{\textrm{conspiracy}}) - 0.95$. Note that no notions of engagement (e.g. view count, comment count) were included as training data to learn $s_{\textrm{conspiracy}}$. Instantiating profit scores as $\log ( (1 + \textrm{\# views}[i])/100,000)$ models that videos with view counts below $100,000$ (roughly $32\%$ of the videos in the validation set) do not break a profit margin.

The ``hand annotated'' subset of the data consists of 541 video instances which are hand-labeled as either conspiracy (251) or non-conspiracy (290), as well as their view counts and predictions. These validation points are drawn from a different distribution, and thus tend to lie toward the extremes of the quality measure (Figure~\ref{fig:yt_training_dist}).

%% file: fairness_appendix.tex

Much recent work on designing the outcomes of decisions has considered adding fairness criteria to the maximum profit objective. 
In the setting of group fairness, the population is partitioned into subgroups, and fairness criteria generally seek to ensure that classifications satisfy a notion of equality (or near-equality) between these groups. 
In this section, we demonstrate how fairness constrained profit maximization corresponds to multiobjective optimization over profit and welfare for a particular definition of welfare.

Consider a population partitioned into subgroups $\popj\in\Omega$ and a classifier which has access to the profit score $p$ of each individual. In this case, we decompose policies over groups such that $\pol = (\polj)_{\popj\in\Omega}$ and
the fairness-constrained profit maximization is given as 
\begin{align}\label{eq:fairness_prob_app}
\pol_\mathrm{fair}^\epsilon \in \argmax_{\pol,\beta}& ~~\utilp(\pol)~~
\mathrm{s.t.}~~ \Exp[\polj(p) ~|~\text{in group }\popj\cap \calC] = \beta_\popj,~~ |\beta_\popi - \beta_\popj| \leq \epsilon ~\text{for all}~\popi,\popj\in\Omega
\end{align}
where the choice of $\calC$ encodes particular fairness criteria. 
For a large class of fairness criteria including \emph{demographic parity} and \emph{equal opportunity},
we can restrict our attention to threshold policies $\polj(p) = \I(p\geq t_\popj)$ where $t_\popj$ are group-dependent thresholds~\citep{liu18delayed}. 
Notice that due to the definition of profit score~\eqref{eq:utility}, the unconstrained solution would simply be $\pol^\maxprof(p) = \I(p\geq 0)$ for all groups.
For this reason, we consider groups with $t_\popj < 0$ as comparatively disadvantaged (since their threshold increases in the absence of fairness constraints) and $t_\popj > 0$ as advantaged.

In this setting, there exist fixed welfare scores $\welfscore$ which achieve the same solution policy for any population. 

\begin{prop}\label{prop:fixed_threshold}
Any fairness-constrained threshold policy giving rise to thresholds $\{t^\star_\popj\}_{j=1}^N$ is equivalent to a set of $\alpha$-Pareto policies for $\alpha \in (0,1)$ in~\eqref{eq:alpha_pareto_score_pol} with welfare scores fixed within each group and defined as 
\begin{align*}
\welfscore_\popj=-\frac{1-\alpha}{\alpha} t^\star_\popj\:.
\end{align*}
In particular, $w_\popj$ and $t^\star_\popj $ have opposite signs for all settings of $\alpha \in (0,1)$, and any relative scale between them achieved by some choice of $\alpha$.
\end{prop}
\begin{proof}
The equivalence follows by comparing the policies
\begin{align*}
 \pol_{\alpha}(\welfscore,\profscore) = \I(\alpha \welfscore + (1-\alpha)\profscore\geq 0)\quad\text{and}\quad
 \pol_{\mathrm{fair},\popj}(\profscore) = \I(\profscore\geq t^\star_\popj)\:.
\end{align*}
Restricting the choice to a fixed score within each group yields the expression
\begin{align*}
\welfscore_j=-\frac{1-\alpha}{\alpha} t^\star_j =: -ct^\star_j\:.
\end{align*}
Thus we have that $w_j \propto - t_j^*$ for all $j \in \{1,...,N\}$. Further, notice that for any $c>0$ there exists some $\alpha\in(0,1)$ achieving that $c$ with $\alpha=\frac{1}{1+c}$.
\end{proof}


\subsection*{Trade-offs between profit and fairness.}

While the result presented above is valid for even inexact fairness constraints, it does not shed light on the trade-off between profit and fairness as the parameter $\epsilon$ varies.
We now show how trading off in the fairness setting can be modeled equivalently by the multi-objective framework. For simplicity, we restrict our attention to the setting of two groups and criteria of demographic parity. 
We note that with additional mild assumptions, our arguments extend naturally to other criteria, including equal opportunity (analogously to section 6.2 of~\citet{liu18delayed}).

Define the two groups as $\popa$ and $\popb$. In this section, we assume that the distribution of the profit score $p$ has continuous support within these populations.
The following proposition shows that the that solution to the constrained profit maximization problem~\eqref{eq:fairness_prob_app} changes monotonically with the fairness parameter $\epsilon$.
\begin{prop}
\label{claim:tau_shrinking}
Suppose that 
the unconstrained selection rate in group $\popa$ is less than or equal to the unconstrained selection rate in group $\popb$.
Then the policies $\pi_\popa^{\epsilon}, \pi_\popb^{\epsilon}$ that optimize eq.~\eqref{eq:fairness_prob_app} with the demographic parity constraint are equivalent to randomized group-dependent threshold policies with thresholds $t_\popa^{\epsilon}$ and  $t_\popb^{\epsilon}$ satisfying the following:
\begin{itemize}
\item{$t_\popa^{\epsilon} \leq 0$ for all $\epsilon \geq 0$ and $t_\popa^{\epsilon}$ is increasing in $\epsilon$}~, 
\item{$t_\popb^{\epsilon} \geq 0$ for all $\epsilon \geq 0$ and $t_\popb^{\epsilon}$ is decreasing in $\epsilon$}~.
\end{itemize}
\end{prop}
Notice that the unconstrained selection rate in group $\popa$ being less than the unconstrained selection rate in group $\popb$ is equivalent {to} $\popa$ being disadvantaged compared with $\popb$.
Thus we see that as $\epsilon$ increases, the group-dependent optimal thresholds shrink toward the unconstrained profit maximizing solution, where $t_\popa = t_\popb = 0$.
We present the proof of this result in the next section.

We define the map $\epsilon_\popa(p) := \epsilon~\text{s.t.}~t^\epsilon_\popa=p$ for $p \in[t_\popa^{0}, 0]$. By \Cref{claim:tau_shrinking}, $\epsilon_\popa(p)$ in increasing in $p$.

Using the previous ingredients, we define a policy based on welfare scores which is equivalent to a fair policy.

\begin{thm}
\label{claim:epsilon_decreases_alpha}
Under the conditions of Proposition~\ref{claim:tau_shrinking}, the family of policies $\pi^\epsilon_\mathrm{fair}$ parametrized by $\epsilon$ corresponds to a family of $\alpha$-Pareto policies solutions for a fixed choice of group-dependent welfare weightings. In particular, denoting the associated thresholds as $t_\popa^{\epsilon}$ and  $t_\popb^{\epsilon}$ and  defining for each individual in $\popa$ with profit score $p$,
\[\welfscore_\popa = \begin{cases}- \frac{p}{t_\popb^{\epsilon_\popa(p)}}&t_\popa^{0} \leq p \leq 0\\ 0 &\text{otherwise} \end{cases}\]
and for all individuals in $\popb$,
\[\welfscore_\popb = \begin{cases}-1 & 0\leq p \leq t_\popb^{0}\\ 0 &\text{otherwise} \end{cases}\:,\]
then for each $\pi^\epsilon_\mathrm{fair}$ there exists an equivalent {$\alpha^{\epsilon}$-Pareto} policy $\pi_{\alpha^\epsilon}$ where the trade-off parameter $\alpha^{\epsilon}$ decreases in $\epsilon$. 
 \end{thm}

\begin{proof}
By Proposition~\ref{claim:tau_shrinking},
the policy $\pi^{\epsilon}$ is equivalent to a threshold policy with group dependent thresholds denoted $t_\popa^\epsilon$ and $t_\popb^\epsilon$. The group dependent threshold policy 
$\I\{ p \geq t_\popj^\epsilon\}$ is equivalent to an $\alpha$-Pareto optimal policy {(for some definition of welfare score $w$)} if and only if {for all values of $p$}:
\begin{align*}
\I\{ p \geq t_\popj^\epsilon\} = \I\{ \alpha^{\epsilon} w + (1-\alpha^{\epsilon}) p \geq 0\}\:.
\end{align*}
It is sufficient to restrict our attention to welfare scores $w$ that depend on profit score and group membership, which we denote as $w_\popj^p$. Starting with group $\popb$, we have that for $0\leq p \leq t_\popb^{0}$, {$w_\popb^p = -1$, so }
\begin{align*}
\pi_{\alpha^{\epsilon}}=\I\{ -\alpha^{\epsilon} + (1-\alpha^{\epsilon}) p \geq 0\}=\I\{   p \geq \frac{\alpha^{\epsilon}}{1-\alpha^{\epsilon}}\}\:.
\end{align*}
Thus, equivalence is achieved for this case if
$\frac{\alpha^{\epsilon}}{1-\alpha^{\epsilon}} = t_\popb^\epsilon$, or equivalently,  
\begin{align}\label{eq:alpha_def}
\alpha^{\epsilon} = \frac{t_\popb^\epsilon}{1+t_\popb^\epsilon}\:.
\end{align}
We will use this definition for $\alpha^\epsilon$ moving forward, and verify that the proposed welfare score definitions work.

We now turn to group $\popa$ in the case that $t_\popa^{0}\leq p \leq 0$. {We have $w_\popa^p = \frac{p}{t_\popb^{\epsilon_\popa(p)}}$, so}
\begin{align*}
\pi_{\alpha^{\epsilon}}=\I\{ - \frac{t_\popb^\epsilon}{t_\popb^{\epsilon_\popa(p)}} \frac{ p }{1+t_\popb^\epsilon} + \frac{p}{1+t_\popb^\epsilon} \geq 0\}\:.
\end{align*}
Because $1+t_\popb^\epsilon\geq 0$ and $p\leq 0$, the indicator will be one if and only if {
$t_\popb^\epsilon   \geq t_\popb^{\epsilon_\popa(p)}$.} By Proposition~\ref{claim:tau_shrinking}, this is true if and only if {$\epsilon\leq \epsilon_\popa(p)$}, which is true if any only if {$t^{\epsilon}_\popa \leq t_\popa^{\epsilon_\popa(p)} = p$}. This is exactly the condition for $\pi_{\mathrm{fair},\popa}^\epsilon$, as desired.

Then finally we consider the remaining cases. 
In the case that $p\leq t_\popa^{0}$ in $\popa$ or $p \leq 0$ in $\popb$, we have that $\pi_{\mathrm{fair},\popj}^\epsilon = 0$ for all $\epsilon$ by Proposition~\ref{claim:tau_shrinking}. Then as desired, $0+(1-\alpha^{\epsilon}) p \leq 0$ in this case.
 In the case that $p \geq 0$ in $\popa$ or $p \geq t_\popb^{0}$ in $\popb$, we have that $\pi_{\mathrm{fair},\popj}^\epsilon = 1$ for all $\epsilon$. Then as desired,  $0+(1-\alpha^{\epsilon}) p \geq 0$ in this case.

Finally, we remark on the form of $\alpha^\epsilon$.
By~\Cref{claim:tau_shrinking}, $t^{\epsilon}_\popb \geq 0$ and is decreasing in $\epsilon$, so $\alpha^{\epsilon}$ is decreasing in $\epsilon$.
\end{proof}

Note that the presented construction of induced welfare scores is not unique. In fact, simply switching the roles of $\popa$ and $\popb$ in the proof verifies the alternate definitions,
\begin{align}\label{eq:flipped_alternative}
\welfscore_\popa = \begin{cases} 1 & t_{\popa}^0 \leq p \leq 0\\ 0 &\text{otherwise} \end{cases}\:,\qquad \welfscore_\popb = \begin{cases}-\frac{p}{t_\popa^{\epsilon_\popb(p)}} & 0\leq p \leq t_\popb^{0}\\ 0 &\text{otherwise} \end{cases}\:,
\end{align}
in which case we define {$\epsilon_\popb(p)$ to be the value of $\epsilon$ such that $p = t^{\epsilon}_\popb$. }
We further remark that this construction generalizes in a straightforward manner to multiple groups, where functions similar to $\epsilon_\popb(p)$ would be defined for each group.

\subsection*{Numerical demonstration.}

We demonstrate the induced welfare scores in the context of a credit lending scenario. In this context, we define the profit score as the expected gain from lending to an individual,
\[p = u_+ \cdot \rho + u_-\cdot (1-\rho)\:,\]
where $\rho$ is the individual's probability of repayment. For this demonstration, we set $u_+ = 1$ and $u_-=-4$, indicating that a default is more costly than a repayment. 

\begin{figure}
    \centering
    \begin{subfigure}[t]{.22\textwidth}
        \centering
        \includegraphics[height=1.4in]{\figpath 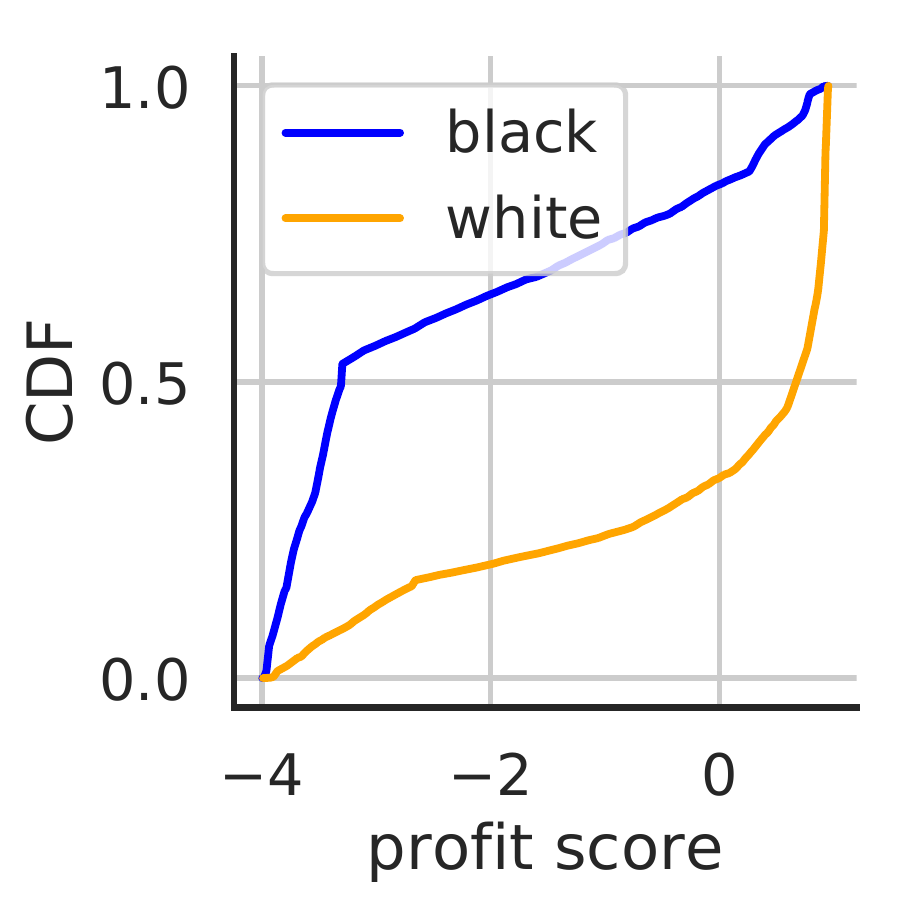}
        \caption{Cumulative density of profit scores group by race, where $88\%$ of the population is white while $12\%$ is black.\label{fig:fairness_cdf}}
    \end{subfigure}%
    ~~
    \begin{subfigure}[t]{.22\textwidth}
        \centering
        \includegraphics[height=1.4in]{\figpath 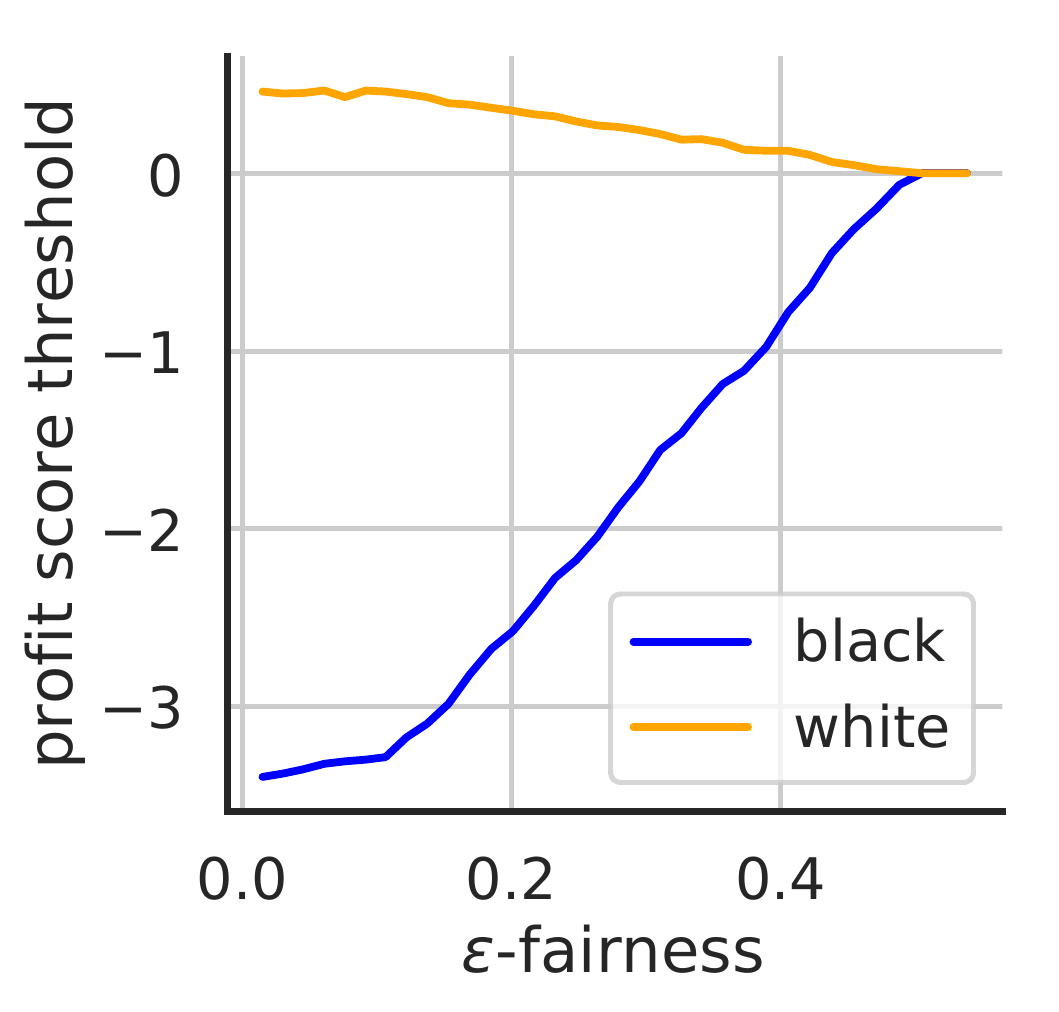}
        \caption{Thresholds for $\epsilon$-fair policies. As $\epsilon$ increases, the magnitude of the thresholds decrease. \label{fig:fairness_threshes}}
    \end{subfigure}%
    ~~
    \begin{subfigure}[t]{.22\textwidth}
        \centering
        \includegraphics[height=1.4in]{\figpath fairness_induced_weights.pdf}
        \caption{Distribution of profit and induced welfare scores. Marker size corresponds to population sizes.\label{fig:fairness_induced_weights_app}}
    \end{subfigure}%
    ~~
    \begin{subfigure}[t]{.22\textwidth}
        \centering
        \includegraphics[height=1.4in]{\figpath fairness_alpha_epsilon.pdf}
        \caption{The fairness parameter $\epsilon$ determines the profit trade-off and corresponds to the welfare weight $\alpha$. \label{fig:fairness_alpha_epsilon_2}}
    \end{subfigure}
        \caption{
        Empirical example of how trade-offs between profit and fairness in lending can be equivalently encoded by a multi-objective framework. \label{fig:fairness_1}}
\end{figure}

We estimate a distribution of profit scores using repayment information from a sample of 301,536 TransUnion TransRisk scores from 2003 published by \citet{fed07}, preprocessed by \citet{hardt_equality_2016}\footnote{
	The data is available at \url{https://github.com/fairmlbook/fairmlbook.github.io/tree/master/code/creditscore/data}.
}.
In this dataset, a default corresponds to failing to pay a debt for at least 90 days on at least one account during a 18-24 month period. 
We consider two race groups: white non-Hispanic (labeled ``white" in figures), and black.
Using the empirical data we estimate the distribution of success probabilities by group and transform this into a distribution over profit scores.
The empirical cumulative density functions are displayed in Figure~\ref{fig:fairness_cdf}.
In this dataset $12\%$ of the population is black while $88\%$ is white.

We solve the optimization problem~\eqref{eq:fairness_prob_app} using a two dimensional grid over thresholds. 
Each pair of thresholds corresponds to an overall profit utility value as well as an acceptance rate difference; these quantities are used to determine the solution to the constrained maximization. 
Figure~\ref{fig:fairness_threshes} shows the thresholds $(t_\popa^\epsilon, t_\popb^\epsilon)$ for various values of $\epsilon$.
As predicted by \Cref{claim:tau_shrinking}, they are generally shrinking in magnitude towards $p=0$, and the threshold is negative for the black group and positive for the white group.
Due to the discrete support of the empirical distributions, the monotonicity of these thresholds is not perfect.

Lastly, we use these threshold values to compute induced welfare scores using the construction given in \Cref{claim:epsilon_decreases_alpha}. 
Figure~\ref{fig:fairness_induced_weights_app} shows how welfare scores are assigned depending on group and on profit score. 
The fact that a restricted range of individuals have nonzero welfare scores highlights the limitations of fairness criteria to affect only individuals within a bounded interval of the max profit solution.
Figure~\ref{fig:fairness_alpha_epsilon_2} shows the welfare weight $\alpha^\epsilon$, which is generally decreasing in $\epsilon$, though not monotonically.
We also see the profit/fairness trade-off.

\subsection*{Technical proofs.}
Before presenting more technical proofs and supporting lemmas, we define important quantities.
Recall that demographic parity constrains the selection rates of policies. Define rate function for each group as
\[r_\popj(\pol) = \Exp[\polj(p) ~|~\text{in group }\popj]\]

Because we focus on threshold policies, we will equivalently write $r_\popj(t) := r_\popj(\I(p\geq t))$.
This function is monotonic in the threshold $t$, and therefore its inverse maps acceptance rates to thresholds which achieve that rate, i.e. $r_\popj^{-1}(\beta) = t_\popj$.
Define $f_\popa(\beta)$ and $f_\popb(\beta)$ to be components of the objective function in~\eqref{eq:fairness_prob_app} due to each group, that is, 
\begin{align*}
f_\popj(\beta) =  \Pr\{\text{in }\popj\}\Exp[\profscore \cdot \I\{ \profscore > r^{-1}_\popj(\beta)\}~|~\text{in }\popj]\:.
\end{align*}
By Proposition 5.3 in \citet{liu18delayed}, the functions $f_\popj$ are concave.
Therefore, the combined objective function is concave in each argument,
\begin{align}\label{eq:obj_fn_beta}
\utilp(\beta_\popa,\beta_\popb) := \sum_{\popj\in\{\popa,\popb\}} \Pr\{\text{in }\popj\}\Exp[\profscore \cdot \I\{ \profscore > r^{-1}_\popj(\beta_\popj)\}~|~\text{in }\popj] = \sum_{\popj\in\{\popa,\popb\}} f_\popj(\beta_\popj)\:.
\end{align}
We restrict our attention to the case that $p$ has continuous support. In this case, the functions $f_\popj$ are differentiable.

\begin{lem}
\label{claim:beta_ordering}
If the distribution of exact profit scores for groups $\popa$ and $\popb$ are such that that maximum profit selection rates $\beta_\popa^\maxprof = r_\popa(0)$ and $\beta_\popb^\maxprof = r_\popb(0)$ and $r_\popa(0) \le r_\popa(0)$, then for any $\epsilon \geq 0$, the selection rates maximizing the optimization problem~\eqref{eq:fairness_prob_app} under demographic parity satisfy the following:
\begin{align*}
\beta_\popa^\maxprof \leq \beta_\popa^{\epsilon} \leq \beta_\popb^{\epsilon} \leq \beta_\popb^\maxprof\:.
\end{align*}
\end{lem}

\begin{proof}[Proof of \Cref{claim:beta_ordering} \\]

First, note we must have that $\beta_\popa^{\epsilon} \leq \beta_\popb^\maxprof$. If it were that $\beta_\popa^{\epsilon} > \beta_\popb^\maxprof$, then the alternate solution $\beta_\popa = \beta_\popb^\maxprof$ and $\beta_\popb=\beta_\popb^\maxprof$ would be feasible for~\eqref{eq:fairness_prob_app}  and achieve a higher objective value by the concavity of~\eqref{eq:obj_fn_beta}.

Then we show that $\beta_\popb^{\epsilon} \leq \beta_\popb^\maxprof$. 
Assume for the sake of contradiction that $\beta_\popb^{\epsilon} > \beta_\popb^\maxprof$. Then since $\beta_\popa^{\epsilon} \leq \beta_\popb^\maxprof$, setting $\beta_\popb = \beta_\popb^\maxprof$ achieves higher objective value without increasing $|\beta_\popb - \beta_\popa^{\epsilon}|$, and thus would be feasible for~\eqref{eq:fairness_prob_app}.
A similar argument shows that $\beta_\popa^{\epsilon} \geq \beta_\popa^\maxprof$.

Then, we show that for any optimal selection rates,  $\beta_\popa^{\epsilon} \leq \beta_\popb^{\epsilon}$ for all $\epsilon\geq 0$. Suppose for the sake of contradiction that $\beta_\popa^{\epsilon} > \beta_\popb^{\epsilon}$. In this case, we can equivalently write that
\[\beta_\popb^\maxprof - \beta_\popb^{\epsilon}  > \beta_\popb^\maxprof - \beta_\popa^{\epsilon} ~~\text{and/or}~~\beta_\popa^{\epsilon} - \beta_\popa^\maxprof > \beta_\popb^{\epsilon} - \beta_\popa^\maxprof\:.\] 
In either case, setting $\beta_\popa^{\epsilon} = \beta_\popb^{\epsilon}$ would be a feasible solution which would achieve a higher objective function value, by the concavity of~\eqref{eq:obj_fn_beta}. This contradicts the assumption that $\beta_\popa^{\epsilon} > \beta_\popb^{\epsilon}$, and thus it must be that $\beta_\popa^{\epsilon} \leq \beta_\popb^{\epsilon}$.
\end{proof}

\begin{lem}\label{lem:single_variable_search}
Under the conditions of \Cref{claim:beta_ordering},
the maximizer $(\beta_\popa^{\epsilon}, \beta_\popb^{\epsilon})$ of the $\epsilon$-demographic parity constrained problem in~\eqref{eq:fairness_prob_app} is either satisfied with the maximum profit selection rates $(\beta_\popa^\maxprof, \beta_\popb^\maxprof)$, or $\beta_\popb^{\epsilon}-\beta_\popa^{\epsilon} = \epsilon$ (or the two conditions coincide).
\end{lem}
\begin{proof}[Proof of~\Cref{lem:single_variable_search}]
If it were that $|\beta_\popa^{\epsilon} - \beta_\popb^{\epsilon}| = \gamma < \epsilon$ then we could construct an alternative solution using the remaining $\epsilon - \gamma$ slack in the constraint which would achieve a higher objective function value, since the functions $f_\popj$ are concave. Furthermore, by \Cref{claim:beta_ordering}, we have that $|\beta_\popa^{\epsilon} - \beta_\popb^{\epsilon}| = \beta_\popb^{\epsilon}-\beta_\popa^{\epsilon} $~.
\end{proof}

This result implies that the complexity of the maximization~\eqref{eq:fairness_prob_app} can be reduced to a single variable search:
\begin{align}\label{eq:obj_fn_beta_1}
\beta^\star = \argmax_\beta f_\popa(\beta) + f_\popb(\beta+\epsilon),~~~\pol^\epsilon_\mathrm{fair} = (\I\{p\geq r_\popj^{-1}(\beta^\star)\},\I(p\geq r_\popj^{-1}(\beta^\star+\epsilon)\})
\end{align}
This expression holds when  $|\beta_\popa^\maxprof- \beta_\popb^\maxprof| > \epsilon$, and otherwise the solution is given by $(\beta_\popa^\maxprof, \beta_\popb^\maxprof)$ .

\begin{lem}
\label{claim:beta_contraction}
Under the conditions of \Cref{claim:beta_ordering},
as $\epsilon \geq 0$ decreases, the group-dependent selection rates $\beta_\popa^\epsilon$ and $\beta_\popb^\epsilon$ become closer to the profit maximizing selection rates for each group. That is, the functions $|\beta_\popa^\epsilon - \beta_\popa^\maxprof|$ and $|\beta_\popb^\epsilon - \beta_\popb^\maxprof|$ are both increasing in $\epsilon$.
\end{lem}
\begin{proof}[Proof of \Cref{claim:beta_contraction}]
We show that for any $\epsilon' \geq \epsilon \geq 0$, it must be that 
$|\beta_\popj^{\epsilon} - \beta_\popj^\maxprof| \leq |\beta_\popj^{\epsilon'} - \beta_\popj^\maxprof|$.
First, we remark that if $|\beta_\popa^\maxprof - \beta_\popb^\maxprof| \leq \epsilon$
or if $\epsilon \leq |\beta_\popa^\maxprof - \beta_\popb^\maxprof| \leq \epsilon'$, the claim holds by application of~\Cref{claim:beta_ordering}.

Otherwise, let the $\epsilon$-demographic parity constrained solution be optimized by 
$ (\beta, \beta +\epsilon)$ and the $\epsilon'$-demographic parity constrained solution be optimized by $(\beta', \beta' + \epsilon')$. 
This is valid by Lemma~\ref{lem:single_variable_search}.
Equivalently, $\beta \in \argmax\{ f_\popa(\beta) + f_\popb(\beta + \epsilon)\}$ and $\beta' \in \argmax\{ f_\popa(\beta') + f_\popb(\beta' + \epsilon')\}$. Since $f_\popa$ and $f_\popb$ are concave and differentiable, 
\begin{align*}
f'_\popa(\beta) + f'_\popb(\beta + \epsilon) &= 0 \quad \textrm{ and } \quad
f'_\popa(\beta') + f'_\popb(\beta' + \epsilon') = 0~.
\end{align*}

Assume for sake of contradiction that $\beta < \beta'$
and recall that by \Cref{claim:beta_ordering} we further have $\beta' > \beta \geq \beta_\popa^{\maxprof}$, so by the concavity of $f_\popa$,
\begin{align*}
f_\popa(\beta) \geq f_\popa(\beta') \quad \textrm{ and } \quad
f_\popa'(\beta') \leq f_\popa'(\beta) ~.
\end{align*}
Analogously, we must have that $\beta_\popb^\maxprof \geq \beta' + \epsilon' > \beta + \epsilon$, so that 
\begin{align*}
f_\popb(\beta' + \epsilon') \geq f_\popb(\beta + \epsilon) \quad \textrm{ and } \quad
f_\popb'(\beta' + \epsilon') \geq f_\popb'(\beta + \epsilon) ~.
\end{align*}
Using the equations above, we have that
\begin{align*}
f'_\popb(\beta + \epsilon) &= - f'_\popa(\beta)\\
&\leq - f'_\popa(\beta')\\
&= f'_\popb(\beta'+\epsilon')
\end{align*}
Since $f_\popb$ is concave and thus its derivative is decreasing, this statement implies that $\beta + \epsilon \geq \beta'+\epsilon'$, which is a contradiction.
Thus, it must be that $\beta \geq \beta'$, i.e. $\beta_\popa^{\epsilon} \geq \beta_\popa^{\epsilon'}$.
With an analogous proof by contradiction, one can show that $\beta_\popb^{\epsilon'} \geq \beta_\popb^{\epsilon}$.

Combining these two  inequalities in \Cref{claim:beta_ordering} completes the proof of \Cref{claim:beta_contraction}.
\end{proof}
\begin{proof}[Proof of \Cref{claim:tau_shrinking} \\]
The proof makes use of \Cref{claim:beta_ordering} and \Cref{claim:beta_contraction}.

First, we show that $t_\popa^{\epsilon} \leq 0$ for all $\epsilon \geq 0$. This is a consequence of \Cref{claim:beta_ordering}, which shows that $\beta_\popa^{\epsilon} \geq \beta_\popa^\maxprof$.  Since $r_\popa$ is a decreasing function (and thus, $r_\popa^{-1}$ is also a decreasing function), this implies that 
\begin{align*}
t_\popa^{\epsilon} = r_\popa^{-1}(\beta_\popa^{\epsilon}) \leq r_\popa^{-1}(\beta_\popa^\maxprof) = 0
\end{align*}
A similar argument holds to show that $t_\popb^\epsilon \geq 0$ for all $\epsilon \geq 0$.

Now we show that $t_\popa^{\epsilon}$ is increasing in $\epsilon$ and $t_\popb^{\epsilon}$ is decreasing in $\epsilon$ to show that both are shrinking toward $0$ as $\epsilon$ increases.
Since $t_\popj = r_\popj^{-1}(\beta)$ is decreasing in $\beta$, \Cref{claim:beta_contraction} implies that the functions $|t_\popa^{\epsilon}| = |t_\popa^{\epsilon} - t_\popa^\maxprof|$ and $|t_\popb^{\epsilon}| = |t_B^{\epsilon} - t_\popb^\maxprof|$ are also decreasing in $\epsilon$ toward the max profit thresholds of $t_\popa^\maxprof = t_\popb^\maxprof = 0$. Since $t_\popa^\epsilon \leq 0$ and $t_\popb^{\epsilon} \geq 0$ for all $\epsilon \geq 0$, this concludes the proof of \Cref{claim:tau_shrinking}.
\end{proof}